%% file: arxiv.tex
\documentclass[11pt,english]{article}
\usepackage[left=1in,right=1in,top=1.in,bottom=1.in]{geometry}
\usepackage[round]{natbib}
\usepackage{microtype}
\usepackage{graphicx}
\usepackage{subfigure}
\usepackage{booktabs} 
\usepackage{graphicx,graphics}
\usepackage{amssymb}
\usepackage{amsbsy}
\usepackage{stmaryrd} 
\usepackage{amsmath}
\usepackage{multirow}
\usepackage{graphicx}
\usepackage{url}  
\usepackage[toc,page]{appendix}
\newcommand{\M}{\B}
\usepackage{ragged2e}
\usepackage{dblfloatfix}
\usepackage{booktabs}       

\DeclareMathOperator*{\tr}{tr}
\DeclareMathOperator*{\argmin}{argmin}
\DeclareMathOperator*{\argmax}{argmax}
\DeclareMathOperator*{\Supp}{Supp}
\DeclareMathOperator*{\subG}{subG}

\newtheorem{theorem}{Theorem}
\newtheorem{defn}{Definition}
\newtheorem{lemma}{Lemma}
\newtheorem{cor}{Corollary}
\newtheorem{asu}{Assumption}
\newenvironment{Proof}{\paragraph{Proof:}}{\hfill$\square$}

\newenvironment{myarray}[2][1]
{\def\arraystretch{#1}\array{#2}}
{\endarray}

\newcommand{\sbt}{\mathrm{s.t. }}
\newcommand{\sign}{\text{sign}}
\newcommand{\B}{\boldsymbol}
\usepackage{amsmath,color}

\title{Improved error rates for sparse (group) learning with Lipschitz loss functions}
\author{Antoine Dedieu}

\begin{document}
\maketitle

\input{main_text_copy0921}

\end{document}

%% file: main_text_copy0921.tex
\begin{abstract}
We study a family of sparse estimators defined as minimizers of some empirical Lipschitz loss function---which include the hinge loss, the logistic loss and the quantile regression loss---with a convex, sparse or group-sparse regularization. In particular, we consider the L1 norm on the coefficients, its sorted Slope version, and the Group L1-L2 extension. 
We propose a new theoretical framework that uses common assumptions in the literature to simultaneously derive new high-dimensional L2 estimation upper bounds for all three regularization schemes. 
For L1 and Slope regularizations, our bounds scale as $(k^*/n) \log(p/k^*)$---$n\times p$ is the size of the design matrix and $k^*$ the dimension of the theoretical loss minimizer $\B{\beta}^*$---and match the optimal minimax rate achieved for the least-squares case. 
For Group L1-L2 regularization, our bounds scale as $(s^*/n) \log\left( G / s^* \right) + m^* / n$---$G$ is the total number of groups and $m^*$ the number of coefficients in the $s^*$ groups which contain $\B{\beta}^*$---and improve over the least-squares case. 
We show that, when the signal is strongly group-sparse, Group L1-L2 is superior to L1 and Slope.
In addition, we adapt our approach to the sub-Gaussian linear regression framework and reach the optimal minimax rate for Lasso, and an improved rate for Group-Lasso.
Finally, we release an accelerated proximal algorithm that computes the nine main convex estimators of interest  when the number of variables is of the order of $100,000s$.
\end{abstract}

\section{Introduction}

We consider a training data of independent samples $\left\{ (\B{x_i},y_i) \right\}_{i=1}^n$, $( \B{x_i},y_i) \in \mathbb{R}^p\times \mathcal{Y}$ from a distribution $\mathbb{P}(\B{X}, \B{y} )$. We fix a loss $f$ and consider a theoretical minimizer $ \B{\beta}^*$ of the theoretical loss $\mathcal{L}(\B{\beta}) = \mathbb{E} \left( f \left( \langle \B{x},  \B{\beta} \rangle ;  y \right)  \right)$:
\begin{equation} \label{def-beta0}
\B{\beta}^* \in \argmin \limits_{ \B{\beta} \in \mathbb{R}^{p}} \mathbb{E} \left( f \left( \langle \B{x},  \B{\beta} \rangle ;  y \right) \right).
\end{equation}
In the rest of this paper, $f(.,y)$ will be assumed to be Lipschitz and to admit a subgradient. We denote $k^* = \| \B{\beta}^*\|_0$ the number of non-zeros coefficients of the theoretical minimizer and $R=\| \B{\beta}^*\|_1$ its L1 norm. We consider the L1-constrained learning problem
\begin{equation} \label{general}
\min \limits_{ \B{\beta} \in \mathbb{R}^{p}:\  \| \B{\beta}  \|_1 \le 2R } \;\; \frac{1}{n}  \sum_{i=1}^n f \left( \langle \B{x_i},  \B{\beta} \rangle ;  y_i \right) + \Omega( \B{\beta} ),
\end{equation}
where $\Omega( \B{\beta} )$ is a regularization function. The L1 constraint in Problems \eqref{general} guarantees that the estimator lies in a bounded set, which is useful for our statistical analysis. We study sparse estimators, i.e. with a small number of non-zeros. To this end, we restrict $\Omega(.)$  to  a class of sparsity-inducing regularizations. We first consider the L1 regularization, which is well-known to encourage sparsity in the coefficients \citep{tibshirani1996regression}. Problem  \eqref{general} becomes:
\begin{equation} \label{l1-problem}
\min \limits_{ \B{\beta} \in \mathbb{R}^{p}:\  \| \B{\beta}  \|_1 \le 2R } \;\; \frac{1}{n}  \sum_{i=1}^n f \left( \langle \B{x_i},  \B{\beta} \rangle ;  y_i \right) + \lambda \| \B{\beta} \|_1.
\end{equation}
The second  problem we study is inspired by the sorted L1-penalty aka the Slope norm~\citep{slope-proximal,bellec2018slope}, used in the context of least-squares for its statistical properties. We note $\mathcal{S}_p$ the set of permutations of $\left\{1,\ldots ,p \right\}$ and consider a sequence $\lambda \in \mathbb{R}^p:$ $\lambda_1 \ge \ldots \ge \lambda_p >0$. For $\eta>0$, we define the L1-constrained Slope estimator as a solution of the convex minimization problem:
\begin{equation} \label{slope-problem}
\min \limits_{ \B{\beta} \in \mathbb{R}^{p}: \ \| \B{\beta} \|_1 \le 2R }  \frac{1}{n} \sum_{i=1}^n  f \left( \langle \B{x_i},  \B{\beta} \rangle ;  y_i \right)  + \eta \| \B{\beta} \|_S
\end{equation}
where $\| \B{\beta} \|_S = \max \limits_{\phi \in \mathcal{S}_p } \sum_{j=1}^p | \lambda_j |  | \beta_{\phi(j)} | = \sum_{j=1}^p \lambda_j | \beta_{(j)} |$ is the Slope regularization and $|\beta_{(1)}| \ge \ldots \ge |\beta_{(p)}| $ is a non-increasing rearrangement of $\B{\beta}$. 

\smallskip
\noindent
Finally, in several applications, sparsity is structured---the coefficient indices occur in groups a-priori  known and it is desirable to select a whole group. In this context, group variants of the L1 norm are often used to improve the performance and interpretability~\citep{yuan2006model, huang2010benefit}. We consider the use of a Group L1-L2 regularization \citep{bach2011convex} and define the L1-constrained Group L1-L2 problem:
\begin{equation} \label{group-problem}
\min \limits_{ \B{\beta} \in \mathbb{R}^{p}: \ \| \B{\beta} \|_1 \le 2R }  \frac{1}{n} \sum_{i=1}^n  f \left( \langle \B{x_i},  \B{\beta} \rangle ;  y_i \right)  + \lambda \sum_{g=1}^G \| \B{\beta}_g \|_{2}.
\end{equation}
where $g=1, \ldots, G$ denotes a group index (the groups are disjoint), $\B\beta_{g}$ denotes the vector of coefficients belonging to group $g$, $\mathcal{I}_g$ the corresponding set of indexes, $n_g = | \mathcal{I}_g|$ and $\B{\beta} = \left(\B{\beta}_1, \ldots, \B{\beta}_G \right)$. In addition, we denote $g_* := \max_{g=1,\ldots,G} n_g$, $\mathcal{J}^* \subset \{1, \ldots, G \}$ the smallest subset of group indexes such that the support of $\B{\beta}^*$ is included in the union of these groups, $s^*:= | \mathcal{J}^* |$ the cardinality of $\mathcal{J}^*$,  and $m^*$ the sum of the sizes of these $s^*$ groups. 



\medskip
\noindent{\bf{What this paper is about:} } In this paper, we propose a statistical analysis of a large class of estimators $\hat{\B{\beta}}(\lambda,R)$\footnote{When no confusion can be made, we drop the dependence upon the parameters $\lambda,R$.}, defined as solutions of Problems \eqref{l1-problem},  \eqref{slope-problem} and \eqref{group-problem} when $f(.; y)$ is a convex Lipschitz loss which admits a subgradient (cf. Assumption \ref{asu1}). 
In particular, we derive new error bounds for the L2 norm of the difference between the empirical and theoretical minimizers $\| \hat{\B{\beta}} - \B{\beta}^*\|_2$. Our bounds are reached under standard assumptions in the literature, and hold with high probability and in expectation. As a critical step, we derive stronger versions of existing cone conditions and restricted strong convexity conditions in the following Theorems \ref{cone-condition} and \ref{restricted-strong-convexity}. Our approach draws inspiration from least-squares analysis \citep{lasso-dantzig, bellec2018slope, huang2010benefit, negahban2012unified}, while discussing the main differences with these works. Finally, our framework is flexible enough to apply to (i) coefficient-based and group-based regularizations (ii)  (quantile) regression and classification problems (logistic regression, SVM) which are notoriously hard and have received little attention (iii) the sub-Gaussian least-squares framework, for which we derive new results.

For the L1 Problem \eqref{l1-problem} and the Slope Problem \eqref{slope-problem}, our bounds scale as $(k^*/n)\log(p/k^*)$. They improve over existing results---which study the specific cases of the L1-regularized hinge, logistic and quantile regression losses \citep{L1-SVM, Wainwright-logreg, quantile-reg}---and reach the optimal minimax rate achieved for the least-squares loss \citep{raskutti_wainwright}. 

For the Group L1-L2 Problem  \eqref{group-problem}, our bounds scale as $(s^*/n)\log(G/s^*)+ m^*/ n$ and appear to be the first existing result. This rate is also better than the existing one for least-squares \citep{huang2010benefit} due to a stronger cone condition (cf. Theorem \ref{cone-condition}). In addition, similar to \citet{huang2010benefit}, we show that when the signal is strongly group-sparse, Group L1-L2 regularization is superior to L1 and Slope regularizations. 

Finally, all the estimators studied herein are tractable, but do not all have available implementations. We therefore release a proximal gradient algorithm to compute them for settings where the number of variables is of the order of $100,000s$, compatible with modern large-scale machine learning applications.

\medskip
\noindent{\bf{Organization of the paper:} }
The rest of this paper is organized as follows. Section \ref{sec:litreview} discusses related work and influential high-dimensional studies for regression and classification problems. Section \ref{sec: framework} builds our theoretical framework, using common assumptions in the literature. Section \ref{sec: error-bound} derives our statistical results. In particular, our main bounds are presented in Theorem \ref{main-results}.  Section \ref{sec:regression} extends our framework to Lasso and Group Lasso. Finally, Section \ref{sec: First-Order} introduces an efficient algorithm to solve Problems \eqref{l1-problem},  \eqref{slope-problem} and \eqref{group-problem}.

\section{Related work}\label{sec:litreview}
Statistical performance and L2 consistency for high-dimensional linear regression have been widely studied \citep{candes2007dantzig, lasso-dantzig, candes-sparse-estimation, bellec2018slope, lounici2011oracle, negahban2012unified}. One important statistical performance measure is the L2 estimation error defined as $\| \hat{\B{\beta}} - \B{\beta}^* \|_2$ where $\B{\beta}^*$ is the $k^*$-sparse vector used in generating the true model and $\hat{\B{\beta}}$ is an estimator. For regression problems with least-squares loss, \citet{candes-sparse-estimation,raskutti_wainwright} established a  $(k^*/n)\log(p/k^*)$ lower bound for estimating the L2 norm of a sparse vector, regardless of the input matrix and estimation procedure. This optimal minimax rate is known to be achieved by a global minimizer of a L0 regularized estimator \citep{bic-tsybakov}. This minimizer is sparse and adapts to unknown sparsity---the degree $k^*$ does not have to be specified---however, it is intractable in practice. Recently, \citet{bellec2018slope} reached this optimal minimax bound for a Lasso estimator with knowledge of the sparsity $k^*$, and proved that a recently introduced and polynomial-time Slope estimator \citep{slope-introduction} achieves the optimal rate while adapting to unknown sparsity. In a related work, \citet{wang2013l1} reached a near-optimal $(k^*/n)\log(p)$ rate for L1 regularized least-angle deviation loss. \citet{quantile-reg} reached this same bound for L1 regularized quantile regression. Finally, in the regime where sparsity is structured, \citet{huang2010benefit} proved a $(s^*/n) \log(G)+ m^*/n~~$ L2 estimation upper bound for a Group L1-L2 estimator---where, similar to our notations, $G$ is the number of groups, $s^*$ the number of relevant groups and $m^*$ their aggregated size---and showed that this group estimator is superior to standard Lasso when the signal is strongly group-sparse---i.e. $m^* / k^*$ is low and the signal is efficiently covered by the groups. \citet{lounici2011oracle} similarly showed that, in the multitask setting, a Group L1-L2 estimator is superior to Lasso.

Little work has been done on deriving estimation error bounds on high-dimensional classification problems. Existing work has focused on the analysis of generalization error and risk bounds  \citep{tarigan, vssvm}. The influential study of \citet{vdg_linear_models} focuses on the analysis of the excess risk for a rich class of Lasso problems. However, the bounds proposed do not explicit the influence of the sparsity degree $k^*$. More importantly, the author does not propose an error bound for L2 coefficients estimation scaling with the problem sizes $(n, p, k^*)$, which is the main result of our work. Note that, unlike the least-squares case, for the problems studied $k^*$ is the sparsity of the theoretical minimizer to estimate (cf. Equation  \eqref{def-beta0}).

In a recent work, \citet{L1-SVM} proved a $(k^*/n) \log(p)$ upper-bound for L2 coefficients estimation of a L1 regularized Support Vector Machines (SVM). \citet{Wainwright-logreg} obtained a similar bound for a L1-regularized logistic regression estimator in a binary Ising graph. Note that both works do not discuss the use of Slope or group regularizations, which we do. This rate of $(k^*/n)\log(p)$ is not the best known for a classification estimator: \citet{one-bit} proved a $k^*\log(p/k^*)$ error bound for estimating a single vector through sparse models---including 1-bit compressed sensing and logistic regression---over a bounded set of vectors. Contrary to this work, our approach does not assume a generative vector and applies to a larger class of losses (hinge loss, quantile regression loss) and regularizations (Slope, Group L1-L2).  
Finally, we are not aware of any existing results for Slope or group regularizations for classification problems.

\section{Framework of study}\label{sec: framework}  
We design herein our theoretical framework of study. Our first assumption requires $f(.,y)$ to be $L$-Lipschitz and to admit a subgradient $\partial f(.,y)$. We list three main examples that fall into this framework. 

\smallskip

\begin{asu} \label{asu1}
	\textbf{Lipschitz loss and existence of a subgradient: } 
	The loss $f(. , y)$ is non-negative, convex and Lipschitz continuous with constant $L$, that is, $| f(t_1, y) - f(t_2, y) | \le L | t_1 -t_2 |, \ \forall t_1, t_2$.
	In addition, there exists $\partial f(.,y)$ such that $f(t_2, y) - f(t_1, y) \ge \partial f(t_1,y) (t_2 -t_1), \ \forall t_1, t_2$. 
\end{asu}

\smallskip
\noindent{\bf{Support vectors machines (SVM)}}
For $\mathcal{Y} = \left\{ -1, 1\right\} $, the SVM problem learns a classification rule of the form $\sign( \langle \B{x},  \B{\beta} \rangle )$ by solving Problem \eqref{general} with the hinge loss $f \left( t;  y \right)= \max(0, 1 - y t)$ with subgradient $\partial f(t,y)=\B{1}(1-yt \ge 0)yt$. The loss satisfies Assumption \ref{asu1} for $L=1$. 

\smallskip
\noindent{\bf{Logistic regression} }
We assume $\log\left( \mathbb{P}(y =1 | \B{x}) \right)$ $- \log\left( \mathbb{P}(y = -1 | \B{x}) \right)= \langle \B{x},  \B{\beta} \rangle$. The maximum likelihood estimator solves Problem \eqref{general} for the logistic loss $f \left( t ;  y \right) = \log(1 + \exp ( - y t))$. The loss satisfies Assumption \ref{asu1} for $L=1$. 

\smallskip
\noindent{\bf{Quantile regression} }
We consider $\mathcal{Y}=\mathbb{R}$ and fix $\theta \in (0,1)$. Following \citet{advance-quantile-reg}, we assume the $\theta$th conditional quantile of $y$ given $\B{X}$ to be $Q_{\theta} (y | \B{X} =  \B{x}) = \langle \B{x},  \B{\beta}_{\theta} \rangle$. We define the quantile loss\footnote{Note that the hinge loss is a translation of the quantile loss for $\theta = 0$.} $\rho_{\theta} (t) = (\theta - \B{1}(t\le0 ) )t$. $\rho_{\theta}$ satisfies Assumption \ref{asu1} for $L = \max(1 - \theta, \theta).$ In addition, it is known \citep{koenker1978regression} that  $\B{\beta}_{\theta} \in \argmin_{ \B{\beta} \in \mathbb{R}^{p}} \mathbb{E} \left[\rho_{\theta} \left( y - \langle \B{x},  \B{\beta} \rangle  \right)\right]$. 





\smallskip
\noindent
We additionally assume the unicity of $ \B{\beta}^*$ and the twice differentiability of the theoretical loss $\mathcal{L}$. 
\begin{asu} \label{asu2}
	\textbf{Differentiability of the theoretical loss: }
	The theoretical minimizer is unique. In addition, the theoretical loss is twice-differentiable: we note its gradient $\nabla \mathcal{L}(.) $ and its Hessian matrix $\nabla^2 \mathcal{L}(.).$ It finally holds: $\nabla \mathcal{L}(.) = \mathbb{E}\left( \partial f \left( \langle \B{x},  .  \rangle ;  y \right) \B{x} \right).$
\end{asu}
Assumption \ref{asu2} is guaranteed for the logistic loss. \citet{lemma2} proved that Assumption \ref{asu2} holds for the hinge loss (the result extends to the quantile regression loss) when the following Assumption \ref{asu2-svm}  is satisfied. \citet{L1-SVM} used this same Assumption \ref{asu2-svm} in their high-dimensional study of L1-SVM.
\begin{asu}  \label{asu2-svm}
The conditional density functions of $\B{x}$ given $y = 1$ and $y=-1$ are continuous with common support and have finite second moments.
\end{asu}

\smallskip
\noindent
Our next assumption controls the entries of the design matrix. Let us first recall the definition of a sub-Gaussian random variable~\citep{lecture-notes}:
\begin{defn} \label{def-asu3}
	A random variable $Z$ is said to be sub-Gaussian with variance $\sigma^2>0$ if $\mathbb{E}(Z) = 0$ and $\mathbb{E}( \exp(t Z)) \le \exp \left( \frac{\sigma^2 t^2}{2} \right), \  \forall t>0$. 
\end{defn}
This variable will be noted $Z \sim \subG(\sigma^2)$. Under Assumptions \ref{asu1}  and \ref{asu2}, it holds $\mathbb{E} \left[  \partial f\left( \langle \B{x}_i,  \B{\beta}^*  \rangle; y_i  \right)  x_{ij}\ \right] =0, \forall i,j$ since $\B{\beta}^*$ minimizes the theoretical loss. In particular, if $|x_{ij} | \le M, \ \forall i,j$, then Hoeffding's lemma guarantees that  $\forall i,j$, $\partial f\left( \langle \B{x}_i,  \B{\beta}^*  \rangle; y_i  \right)  x_{ij}$, is sub-Gaussian with variance $L^2 M^2$. We therefore define Assumption \ref{asuSG} as follows:
\begin{asu} \label{asuSG}
	\textbf{Sub-Gaussian entries: }
	\newline
	$\bullet$ There exists $M>0$: $\partial f\left( \langle \B{x}_i,  \B{\beta}^*  \rangle, y_i  \right)  x_{ij} \sim \subG(L^2 M^2), \ \forall i,j$.
	
	\noindent
	$\bullet$ For Group L1-L2 regularization, we additionally assume that \newline:
	$\partial f\left( \langle \B{x}_i,  \B{\beta}^*  \rangle, y_i  \right)  (\B{x}_i)_{g} \sim \subG(L^2 M^2). \ \forall i,g$.
\end{asu}

\smallskip
\noindent
The next assumption draws inspiration from the restricted eigenvalue conditions defined for all three L1, Slope and Group L1-L2 regularizations in the least-squares settings \citep{lasso-dantzig, bellec2018slope, lounici2011oracle}. 
\begin{asu} \label{asu4}
	\textbf{Restricted eigenvalue conditions:}
		\newline
		$\bullet$ Let $k \in \left\{1, \ldots ,p \right\}$. Assumption \ref{asu4}$.1(k)$ is satisfied if there exists a non-negative constant $\mu(k) $ such that almost surely:
		$$ \mu(k) \ge \sup \limits_{  \B{z} \in \mathbb{R}^{p} : \ \| \B{z}  \|_0 \le k  } \frac{ \sqrt{k} \| \B{X} \B{z}  \|_2  }{ \sqrt{n} \|\B{z}\|_1  }.
		$$
		
		\noindent
		$\bullet$   Let $\gamma_1, \gamma_2>0$. Assumption \ref{asu4}$.2(k, \gamma)$ holds if there exists $\kappa(k, \gamma_1, \gamma_2)$ which almost surely satisfies:
		$$0 < \kappa(k, \gamma_1, \gamma_2) \le \inf \limits_{| S | \le k } \ \inf \limits_{ \substack{\B{z} \in \Lambda(S, \gamma_1, \gamma_2) } } \frac{ \B{z}^T \nabla^2 \mathcal{L}(\B{\beta}^*)  \B{z}    }{ \|\B{z}\|_2^2  },$$
		where $\gamma=(\gamma_1, \gamma_2)$ and for every subset $S \subset \left\{1, \ldots ,p\right\}$, the cone $\Lambda(S, \gamma_1, \gamma_2) \subset \mathbb{R}^{p}$ is defined as:
		$$\Lambda(S, \gamma_1, \gamma_2) = \left\{ \B{z}: \ \| \B{z}_{S^c}  \|_1 \le \gamma_1 \| \B{z}_{S}  \|_1 + \gamma_2 \| \B{z}_{S}  \|_2 \right\}.$$
		
		\noindent
		$\bullet$   Let $\omega > 0$. Assumption \ref{asu4}$.3(k, \omega)$ holds if there exists a constant $\kappa(k, \omega) > 0$ such that a.s.:
		$$ 0 < \kappa(k, \omega) \le  \inf \limits_{ \substack{\B{z} \in \Gamma(k, \omega) } } \frac{ \B{z}^T \nabla^2 \mathcal{L}(\B{\beta}^*)  \B{z}  }{ \|\B{z}\|_2^2  },$$
		where the cone $\Gamma(k, \omega) \subset \mathbb{R}^{p}$ is defined as:
		$$\Gamma(k, \omega) = \left\{ \B{z}: \ \sum_{j=k+1}^p \lambda_j |z_{(j)}| \le   \omega \sum_{j=1}^{k} \lambda_j |z_{(j)}| \right\},$$
		with $| z_{(1)}| \ge \ldots \ge |z_{(p)}|, \ \forall \B{z}$.
		
		\noindent
		$\bullet$  Let $\epsilon_1, \epsilon_2 > 0$. Assumption \ref{asu4}$.4(s, \epsilon)$ holds if there exists a constant $\kappa(s, \epsilon_1, \epsilon_2)>0$ such that a.s.:
		$$0 < \kappa(s, \epsilon_1, \epsilon_2) \le \inf \limits_{| \mathcal{J} | \le s } \ \inf \limits_{ \substack{\B{z} \in \Omega(\mathcal{J}, \epsilon_1, \epsilon_2) } } \frac{ \B{z}^T \nabla^2 \mathcal{L}(\B{\beta}^*)  \B{z}   }{ \|\B{z}\|_2^2  },$$
		where $\epsilon=(\epsilon_1, \epsilon_2)$ and for every subset $\mathcal{J} \subset \left\{1, \ldots ,G \right\}$, we define $\mathcal{T}(\mathcal{J}) = \cup_{g \in \mathcal{J}} \mathcal{I}_g$ the subset of all indexes accross all the groups in $\mathcal{J}$. $\Omega(\mathcal{J}, \epsilon_1, \epsilon_2) $ is defined as:
		$$\left\{ \B{z}: \ \sum_{g \notin \mathcal{J}} \| \B{z}_g \|_{2} \le  \epsilon_1 \sum_{g \in \mathcal{J}} \| \B{z}_g \|_{2} + \epsilon_2 \| \B{z}_{\mathcal{T}(\mathcal{J})  } \|_{2} \right\}.$$
\end{asu}
In the SVM framework \citep{L1-SVM}, Assumptions A3 and A4 are similar to our Assumptions $4.1$ and $4.2$. For logistic regression \citep{Wainwright-logreg}, Assumptions A1 and A2 similarly define a dependency and incoherence conditions. For quantile regression \citep{quantile-reg}, Assumption D.4 is equivalent to a uniform restricted eigenvalue condition. 



\medskip
\noindent
Since $\B{\beta}^*$ minimizes the theoretical loss, it holds  $\nabla \mathcal{L}(\B{\beta}^*) = 0$. In particular, under Assumption \ref{asu4}, the theoretical loss is lower-bounded by a quadratic function on a certain subset surrounding $\B{\beta}^*$. By continuity, we define the maximal radius on which the following lower bound holds:
\begin{align*}
\begin{split}
r^* = \max \left\{  
r \; \bigg \lvert \; \mathcal{F}(\B{\beta}^*, \B{z}, \kappa^*) \ge 0, \;
\forall \B{z} \in \mathcal{C}^*, \;\; \| \B{z} \|_2 \le r
\right\}
\end{split}
\end{align*}
where we have defined:

$\bullet$ $\mathcal{F}(\B{\beta}^*, \B{z}, \kappa^*) = \mathcal{L}(\B{\beta}^* + \B{z} ) - \mathcal{L}(\B{\beta}^*) - \frac{1}{4} \kappa^* \| \B{z} \|_2^2.$

$\bullet$ $ \mathcal{C}^* = \bigcup \limits_{S \subset \left\{1, \ldots ,p\right\}: \; | S | \le k^*} \Lambda(S, \gamma_1, \gamma_2)$ and $\kappa^* = \kappa \left(k^*, \gamma_1, \gamma_2 \right)$ for L1 regularization.

$\bullet$ $ \mathcal{C}^* =  \Gamma(k^*, \omega) $ and $\kappa^* = \kappa \left(k^*, \omega \right)$ for Slope.

$\bullet$ $ \mathcal{C}^* = \bigcup \limits_{\mathcal{J} \subset \left\{1, \ldots ,G\right\}: \; | \mathcal{J} | \le s^*} \Omega(\mathcal{J}, \epsilon_1, \epsilon_2)$ and $\kappa^* = \kappa \left(s^*, \epsilon_1, \epsilon_2  \right)$ for Group L1-L2 regularization.

\smallskip
\noindent
$r^*$ depends upon the same parameters than $\kappa^*$. The following growth conditions give relations between the number of samples $n$, the dimension space $p$, the sparsity levels $k^*$ and $s^*$ , the maximal radius $r^*$, and a parameter $\delta$.

\begin{asu} \label{asu5}
	\textbf{Growth conditions:}
	Let $\delta \in (0,1)$. 
	\newline
	We first assume that $n\le p$ and $\log(7Rp) \le k^*$.
	Assumptions \ref{asu5}$.1(p, k^*, \alpha, \delta)$ and \ref{asu5}$.2(p, k^*, \alpha, \delta)$---defined for L1 and Slope regularizations---are said to hold if:  
	$$\kappa^* r^* \ge  4\sqrt{k^*} (\tau^*+ \lambda)$$ 
	where $\lambda$ and $\tau^*= \tau^*(k^*)$ are respectively defined in the following Theorems \ref{cone-condition} and  \ref{restricted-strong-convexity}. 
	\smallskip
	\newline
	In addition, for Group L1-L2 regularization, Assumption \ref{asu5}$.3(G, s^*, m^*, \alpha, \delta)$ is said to hold if:
	$$\kappa^* r^* \ge  4 (\tau^*\sqrt{m^*} + \lambda_G\sqrt{s^*}  ), ~ \text{ and } m_0 \le \gamma m^*$$
	where $\gamma \ge 1$ and $m_0, \lambda_G$, are also defined in the following Theorems \ref{cone-condition} and  \ref{restricted-strong-convexity}.
\end{asu}
Note that Assumption \ref{asu5} is similar to Equation (17) for logistic regression \citep{Wainwright-logreg}. A similar definition is proposed in the proof of Equation $(3.9)$ for quantile regression \citep{quantile-reg}: the corresponding quantity $q$ is introduced in Equation $(3.7)$. 

\medskip
\noindent
Our framework can now be used to derive upper bounds for L2 coefficients estimation, scaling with the problem size parameters and the constants introduced.


\section{Statistical analysis}\label{sec: error-bound}

In this section, we study the statistical properties of the estimators defined as solutions of Problems \eqref{l1-problem},  \eqref{slope-problem} and \eqref{group-problem} and derive new upper bounds for L2 coefficients estimation.

\subsection{Cone conditions}\label{sec: cone condition}

Similar to the regression cases for L1, Slope and Group L1-L2 regularizations~\citep{lasso-dantzig, bellec2018slope, lounici2011oracle}, Theorem \ref{cone-condition} first derives cone conditions satisfied by respective solutions of Problem \eqref{l1-problem},  \eqref{slope-problem} or  \eqref{group-problem}. Theorem \ref{cone-condition} says that, for each problem, the difference between the theoretical and empirical minimizers belongs to one of the families of cones defined in Assumption \ref{asu4}. 
The cone conditions are derived by selecting a regularization parameter large enough so that it dominates the sub-gradient of the loss $f$ evaluated at the theoretical minimizer $\B{\beta}^*$.

\begin{theorem} \label{cone-condition}
	Let $\delta \in \left(0, 1 \right)$, $\alpha\ge 2$, and assume that Assumptions \ref{asu1} and \ref{asuSG} are satisfied. 
	\newline We denote $\lambda_j^{(r)} =  \sqrt{  \log(2re/j) }, \forall j, \forall r$ and fix the parameters $\eta = 12 \alpha L M \sqrt{\frac{\log(2/ \delta)}{n}}$, $\lambda = \eta \lambda_{k^*}^{(p)}$ for L1 regularization and $\lambda_G = \eta \lambda_{s^*}^{(G)} + \alpha LM \sqrt{ \frac{\gamma m^*}{s^* n}}$ for Group L1-L2 regularization. 
	\newline The following results hold with probability at least $1 - \frac{\delta}{2}$.
	
	    \noindent
		$\bullet$  Let $\hat{\B{\beta}}_1$ be a solution of the L1 regularized Problem  \eqref{l1-problem} with parameter $\lambda$, and $S_0\subset \{1,\ldots,p\}$ be the subset of indexes of the $k^*$ highest coefficients of $\B{h}_1:= \hat{\B{\beta}}_1 - \B{\beta}^*$. It holds:
		$$\B{h}_1 \in \Lambda \left( S_0, \ \gamma_1^*:= \frac{\alpha}{\alpha -1},  \ \gamma_2^*:= \frac{\sqrt{k^*}}{\alpha -1} \right).$$
		
		\noindent
		$\bullet$   Let $\hat{\B{\beta}}_{\mathcal{S}}$ be a solution of the Slope regularized  Problem \eqref{slope-problem} with parameter $\eta$ and for the sequence of coefficients $\lambda_j^{(p)} =  \sqrt{  \log(2pe/j) }, \forall j$. It holds:
		$$\B{h}_{\mathcal{S}} := \hat{\B{\beta}}_{\mathcal{S}}  - \B{\beta}^*\in \Gamma \left( k^*, \; \omega^*:= \frac{\alpha +1}{\alpha -1} \right).$$
		
		\noindent
		$\bullet$ Let $\hat{\B{\beta}}_{L1-L2}$ be a solution of the Group L1-L2 Problem \eqref{group-problem} with parameter $\lambda_G$. Let $\mathcal{J}_0 \subset \{1,\ldots,G\}$ be the subset of indexes of the $s^*$  highest groups of $\B{h}_{L1-L2} :=\hat{\B{\beta}}_{L1-L2}  - \B{\beta}^*$ for the L2 norm, and $m_0$ be the total size of the $s^*$ largest groups. Finally let $\mathcal{T}_0 = \cup_{g \in \mathcal{J}_0} \mathcal{I}_g$ define the subset of size $m_0$ of all indexes across all the $s^*$ groups in $\mathcal{J}_0$. It holds:
		$$\B{h}_{L1-L2} \in \Omega \left( \mathcal{J}_0, \; \epsilon^*_1:=\frac{\alpha}{\alpha - 1}, \; \epsilon^*_2:=\frac{\sqrt{s^*}}{\alpha - 1} \right).$$
		
\end{theorem}
The proof is presented in Appendix \ref{sec: appendix_cone-condition}. To derive it, we introduce a new result in Lemma \ref{upper-bound-sup} (cf. Appendix \ref{sec: appendix_lemma_bound})  which controls the maximum of sub-Gaussian random variables. 

\medskip
\noindent
\textbf{Connection with prior work: }
For the L1 regularized Problem \eqref{l1-problem}, the parameter $\lambda^2$ is of the order of $\log(p/k^*) /n $. In particular, our conditions are stronger than  \citet{L1-SVM}, \citet{Wainwright-logreg} and \citet{wang2013l1}, which all propose a $\log(p) / n$ scaling when pairing L1 regularization with the three Lipschitz losses considered herein. 

In addition, for Group L1-L2 regularization, the parameter $\lambda_G^2$ is of the order of $\log(G/s^*) / n + m^* /  (s^* n)$: our condition is also stronger than \citet{huang2010benefit}, which derive a $\log(G) / n + m^*/n$ scaling for least-squares. 

\subsection{Restricted strong convexity conditions}\label{sec: restricted}

The next Theorem \ref{restricted-strong-convexity} says that the loss $f$ satisfies a restricted strong convexity (RSC) \citep{negahban2012unified}  with curvature $ \kappa^* /4$ and L1 tolerance function. It is derived by combining (i) a supremum result presented in Theorem \ref{hoeffding-sup} (ii) the minimality of $\B{\beta}^*$ and (iii) restricted eigenvalue conditions from Assumption \ref{asu4}.

\begin{theorem} \label{restricted-strong-convexity}
	Let $\delta \in (0,1)$ and assume that Assumptions \ref{asu1},  \ref{asu2} and \ref{asuSG} hold. In addition, assume that Assumptions \ref{asu4}$.1(k^*)$ and \ref{asu4}$.2(k^*, \gamma^*)$ hold for L1 regularization, Assumptions \ref{asu4}$.1(k^*)$ and \ref{asu4}$.3(k^*, \omega^*)$ for Slope, Assumptions \ref{asu4}$.1(g_*s^*)$ and  \ref{asu4}$.4(s^*, \epsilon^*)$ for Group L1-L2---where $ \gamma^*$, $\omega^*$ and $\epsilon^*$ are defined in Theorem \ref{cone-condition}. 
	\newline
	Finally, let $\tau(k) = 9 L  \mu(k) \sqrt{\frac{1 }{n} + \frac{ \log\left( 2/\delta\right)  }{nk}  } $ for all integers $k,m,q$ and let $\B{h}=\hat{\B{\beta}} - \B{\beta}^*$ be a shorthand for $\B{h}_1$, $\B{h}_{\mathcal{S}}$, or $\B{h}_{L1-L2}$.
	
	\smallskip
	\noindent 
	Then, it holds with probability at least $1-\frac{\delta}{2}$: 
	\begin{align}
	&\frac{1}{n} \sum_{i=1}^n f \left( \langle \B{x}_i,  \B{\beta}^*+ \B{h}  \rangle ;  y_i \right)  - \frac{1}{n} \sum_{i=1}^n  f \left( \langle \B{x}_i,  \B{\beta}^* \rangle ;  y_i \right)
	\ge \frac{1}{4}  \kappa^* \left\{  \|\B{h}\|_2^2 \wedge  r^* \|\B{h}\|_2  \right\}  - \tau^* \| \B{h} \|_1 - \frac{4}{n} L \mu(k),
	\end{align}
	where $\tau^* = \tau(k^*) $.  $\kappa^*$, $r^*$ are shorthands for the restricted eigenvalue constant and maximum radius introduced in Assumptions \ref{asu4} and \ref{asu5}: they depend upon the regularization used.
\end{theorem}

\medskip
\noindent
\textbf{Connection with prior work: }
The above conditions can be extended to the use of an L2 tolerance function: our parameter $(\tau^*)^2$ would scale as $k^*/n$. In contrast, \citet{L1-SVM, Wainwright-logreg, negahban2012unified} propose a parameter scaling as $k^* \log(p)/n$ with an L2 tolerance function: our conditions are stronger.

\medskip
\noindent
\textbf{Deriving RSC conditions : }
The following Theorem \ref{hoeffding-sup} is a critical step to prove Theorem \ref{restricted-strong-convexity}, and is one of the main novelty of our analysis. To motivate it, it helps considering the difference between the linear regression framework and the one studied herein. The former assumes the generative model $\B{y} = \B{X} \B{\beta}^* + \B{\epsilon}$. Therefore, when $f$ is the least-squares loss, using the notations of Theorem \ref{hoeffding-sup} it holds:  $\Delta(\B{\beta}^*, \B{z}) = \frac{1}{n} \| \B{X}\B{z} \|_2^2 - \frac{2}{n} \B{\epsilon}^T\B{X}\B{z} $. By combining a cone condition (similar to Theorem 1) with an upper-bound of the term $\epsilon^T \B{X} \B{z} $, we directly obtain a RSC condition similar to Theorem \ref{restricted-strong-convexity} (see Section \label{sec:regression}). However, in our study, $\B{\beta}^*$ is simply defined as the minimizer of the theoretical risk. Two majors differences appear: (i) we cannot simplify $\Delta( \B{\beta}^*, \B{z} )$ with linear algebra, (ii) we need to introduce the expectation $\mathbb{E} (\Delta(\B{\beta}^*, \B{z} ))$ and to control the quantity $ | \mathbb{E}(\Delta(\B{\beta}^*,  \B{z} )) - \Delta(\B{\beta}^*, \B{z}) |$. Theorem \ref{hoeffding-sup} explicits the cost for controlling this quantity over a bounded set of sparse vectors with disjoint supports. Its proof is presented in Appendix \ref{sec: hoeffding-sup}.

\begin{theorem} \label{hoeffding-sup} We define  $\forall  \B{w}, \B{z} \in \mathbb{R}^p$:
	$$\Delta(\B{w},  \B{z}) = \frac{1}{n} \sum_{i=1}^n f \left( \langle \B{x}_i,  \B{w}+ \B{z}  \rangle ;  y_i \right)  - \frac{1}{n} \sum_{i=1}^n  f \left( \langle \B{x}_i,  \B{w} \rangle ;  y_i \right)$$ 
	Let  $k, m, q\in \left\{1, \ldots, p\right\}$ be such that $m \le k$, $n \le q$ and $m \log(7Rq) \le k$. Let $S_1, \ldots S_q$ be partition of $\left\{1, \ldots, p\right\}$ of size $q$ such that $|S_{\ell}| \le m, \ \forall \ell$. Let $\phi = \frac{2}{nq}  L \mu(k)$. We assume that Assumptions \ref{asu1} and \ref{asu4}$.1(k)$ hold and note $\tau=\tau(k)$ as defined in Theorem \ref{restricted-strong-convexity}. 
	
	\smallskip
	\noindent 
	Then, for $\delta \in (0,1)$, it holds with probability at least $1-\frac{\delta}{2}$:
	$$  \sup \limits_{ \substack{  \B{z}_{S_1}, \ldots, \B{z}_{S_q}  \in \mathbb{R}^{p}: \\ \Supp(\B{z}_{S_j})  \subset S_j, \;\; \| \B{z}_{S_j}  \|_1 \le 3R, \ \ \ \forall j } }   \left\{ 
	\mathcal{G}( \B{\beta}^* , \B{z}_{S_1}, \ldots, \B{z}_{S_q}  ) 
	\right\} 
	\le 0,$$
	\newline where $\Supp(.)$ refers to the support of a vector and we define $\B{w}_{\ell} = \B{\beta}^*  + \sum \limits_{j=1}^{\ell} \B{z}_{S_j}, \forall \ell$ and
	\newline
	$$\mathcal{G}( \B{\beta}^* , \B{z}_{S_1}, \ldots, \B{z}_{S_q} ) = \sup \limits_{\ell=1,\ldots,q} 
	\left\{ \left|  \Delta \left( \B{w}_{\ell-1}, \B{z}_{S_\ell} \right)   - \mathbb{E}\left(  \Delta \left( \B{w}_{\ell-1}, \B{z}_{S_\ell} \right) \right) \right| 
	-  \tau \| \B{z}_{S_\ell}\|_1
	\right\}$$
\end{theorem}

\subsection{Upper bounds for coefficients estimation}
We conclude this section by presenting our main bounds in Theorem \ref{main-results}. 
The proof is presented in Appendix \ref{sec: appendix_main-results}. The bounds follow from the cone conditions and the restricted strong convexity conditions derived in Theorems \ref{cone-condition} and \ref{restricted-strong-convexity}. 
\begin{theorem} \label{main-results}
	Let $\delta \in \left(0, 1 \right)$. We consider the same assumptions and notations than in Theorems \ref{cone-condition} and \ref{restricted-strong-convexity}.  In addition, we assume that the growth conditions \ref{asu5}$.1(p, k^*,  \alpha, \delta)$, \ref{asu5}$.2(p, k^*,  \alpha, \delta)$ and \ref{asu5}$.3(G, s^*, m^*,\alpha, \delta)$ respectively hold for L1, Slope and Group L1-L2 regularizations. We select $\alpha\ge 2$ so that $\mu(k^*) \le \alpha M$.
	
	\smallskip
	\noindent 
	Then the L1 and Slope estimators $ \hat{\B{\beta}}_1$ and $ \hat{\B{\beta}}_{\mathcal{S}}$ satisfies with probability at least $1-\delta$:
	\begin{equation*}
	\begin{split}   
	&\| \hat{\B{\beta}}_{1, \mathcal{S}} - \B{\beta}^*\|_2 \lesssim \frac{ L \mu(k^*)  }{ \kappa^*} \sqrt{ \frac{ k^* + \log\left( 2/ \delta \right) }{n}}  + \frac{ \alpha L M }{\kappa^*} \sqrt{ \frac{ k^* \log\left( p/k^* \right) \log\left( 2/\delta \right) }{n} }.
	\end{split}
	\end{equation*}
	In addition, the Group L1-L2 estimator $ \hat{\B{\beta}}_{L1-L2}$ satisfies with probability at least $1-\delta$:
	\begin{equation*}
	\begin{split}
	&\| \hat{\B{\beta}}_{L1-L2} - \B{\beta}^* \|_2 \lesssim 
	 \frac{ L \mu(k^*)  }{ \kappa^*}  \sqrt{ \frac{ m^* + \log\left( 2/ \delta \right)  }{n}} + \frac{ \alpha L M }{\kappa^*} \sqrt{  \frac{ s^* \log\left( G/s^* \right) \log\left( 2/\delta \right) + \gamma m^* }{n} }.
	\end{split}
	\end{equation*}
	where $\kappa^* = \kappa \left(S_0, \gamma_1^*, \gamma_2^* \right)$ for L1 regularization,  $\kappa^* = \kappa \left(k^*, \omega^*\right)$ for Slope regularization and $\kappa^* = \kappa \left(\mathcal{J}_0, \epsilon_1^*, \epsilon_2^*\right)$ for Group L1-L2 regularization. 
\end{theorem}
Theorem \ref{main-results} holds for any $\delta < 1$. Thus, we obtain by integration the following bounds in expectation, which we prove in Appendix \ref{sec: appendix_main-corollary}.
\begin{cor} \label{main-corollary}
	If the assumptions presented in Theorem \ref{main-results} are satisfied for a small enough $\delta$, then:
	\begin{equation*}
	\begin{split}
	&\mathbb{E} \| \B{h}_{1, \mathcal{S}} \|_2 \lesssim \frac{ L (\alpha M + \mu(k^*)) }{\kappa^*} \sqrt{ \frac{k^* \log\left( p /k^* \right)}{n} },
	\end{split}
	\end{equation*}
	\begin{equation*}
	\begin{split}
	&\mathbb{E} \| \B{h}_{L1-L2} \|_2 \lesssim \frac{L (\alpha M + \mu(k^*))}{\kappa^*}  \sqrt{ \frac{s^* \log\left( G / s^* \right) + \gamma m^*}{n} }.
	\end{split}
	\end{equation*}
\end{cor}

\smallskip
\noindent
\textbf{Discussion for L1 and Slope: } For L1 and Slope regularizations, our family of estimators reach a bound scaling as $(k^* / n) \log(p/k^*)$. This bound strictly improves over existing results for L1-regularized versions of all three losses \citep{L1-SVM,Wainwright-logreg,wang2013l1,quantile-reg} and it matches the best rate known for the least-squares case \citep{bellec2018slope}. 
In addition, the L1 regularization parameter $\lambda$ uses the sparsity $k^*$. In contrast, similar to least-squares \citep{bellec2018slope}, Slope presents the statistical advantage of adapting to unknown sparsity: the sparsity degree $k^*$ does not have to be specified.

\medskip
\noindent
\textbf{Discussion for Group L1-L2: } 
For Group L1-L2, our family of estimators reach a bound scaling as $(s^* / n) \log\left( G/s^* \right) + m^* / n$. This bound improves over the best rate known for the least-squares case \citep{huang2010benefit}, which scales as $(s^* /n) \log\left( G \right) + m^* / n$. This is due to the stronger cone condition derived in Theorem \ref{cone-condition}.


\medskip
\noindent
\textbf{Comparison of both bounds for group-sparse signals: }
We compare the upper bounds of Group L1-L2 regularization to L1 and Slope regularizations when sparsity is structured.
Let us first consider two edge cases. 

\textbf{(i)} If all the groups are of size $k^*$ and the optimal solution is contained in only one group---that is, $g_*=k^*$, $G = \lceil p / k^* \rceil$, $s^*=1$, $m^*=k^*$, $\gamma=1$---the  bound for Group L1-L2 is lower than the ones for L1 and Slope. Group L1-L2 is superior as it strongly exploits the problem structure. 

\textbf{(ii)} If now all the groups are of size one---that is, $g_*= 1$, $G = p$, $s^*=k^*$, $m^*=k^*$, $\gamma=1$---both bounds are similar as the group structure is not relevant. 

For the general case, when $m_* / k_* \ll \log(p/k^*)$, the signal is efficiently covered by the groups---the group structure is useful---and we say that the signal is strongly group-sparse \citep{huang2010benefit}. In this case, the upper bound for Group L1-L2 is lower than the one for L1 and Slope. That is, similar to the regression case \citep{huang2010benefit}, Group L1-L2 is superior to L1 for strongly group-sparse signals. However, when $m_* / k_*$ is larger, sparsity is not as useful.

\section{Connection with least-squares}\label{sec:regression}
As previously discussed, the rate derived for L1 and Slope regularizations in Corollary \ref{main-corollary} matches the optimal minimax rate achieved for the least-squares case \citep{bellec2018slope}, whereas the  rate for Group L1-L2 improves over the least-squares case \citep{huang2010benefit}. 

We propose herein to show the flexibility of our approach by introducing a simplified version of our framework and proof techniques which allows us to (i) simply recover the best rate known \citep{bellec2018slope} for L1-regularized least-squares estimator---aka Lasso \citep{tibshirani1996regression}---for a sub-Gaussian noise (\citet{bellec2018slope} assume a Gaussian noise) and (ii) improve the best rate known for Group L1-L2 regularized least-squares---aka Group Lasso \citep{huang2010benefit}.  

\smallskip
\noindent
We consider the usual linear regression framework, with response $\B{y} \in \mathbb{R}^n$ and model matrix $\B{X} \in \mathbb{R}^{n \times p}$:
\begin{equation}\label{linreg}
\B{y} = \B{X} \B{\beta}^* + \B{\epsilon}
\end{equation}
where the entries of $\B{\epsilon}=(\epsilon_1, \ldots, \epsilon_p)$ are independent sub-Gaussian realizations $\epsilon_j \sim \subG(\sigma^2), ~ \forall j$. The Lasso estimator is defined as a solution of the convex problem
\begin{equation} \label{lasso}
\min \limits_{ \B{\beta} \in \mathbb{R}^{p} } \;\; \frac{1}{n} \| \B{y} - \B{X} \B{\beta}  \|_2^2+ \lambda \| \B{\beta} \|_1,
\end{equation}
whereas the Group Lasso estimator solves the problem
\begin{equation} \label{group-lasso}
\min \limits_{ \B{\beta} \in \mathbb{R}^{p} } \;\; \frac{1}{n} \| \B{y} - \B{X} \B{\beta} \|_2^2 + \lambda \sum_{g=1}^G \| \B{\beta}_g \|_{2}.
\end{equation}
We use the notations previously introduced. The analysis of Lasso and Group Lasso only requires two assumptions, which simplifies the framework developped in Section \ref{sec: framework}.
 
Our first Assumption \ref{asu-col} is an adaptation of Assumption \ref{asuSG}. For Lasso, it assumes a bound on the L2 norm of the columns of the model matrix, as in the literature \citep{bellec2018slope}. For Group Lasso, it draws inspiration from Assumptions 4.1 and 4.2 in \citet{huang2010benefit} and assumes, for each group, an upper bound for the quadratic form associated with $\B{X}_g^T \B{X}_g$---where $\B{X}_g \in \mathbb{R}^{n \times n_g}$ denotes the restriction of the model matrix to the columns $\mathcal{I}_g$ of group $g$.
\begin{asu} \label{asu-col} 
    \textbf{Upper bounds: }
	\newline
	$\bullet$ For Lasso, the model matrix satisfies  $\| \B{X}_j \|_2 \le \sqrt{n}, ~ \forall j.$
	\smallskip
	\newline
	$\bullet$ For Group Lasso, let $\mu_{\max}(\B{X}_g^T \B{X}_g)$ be the highest eigenvalue of the positive semi-definite symmetric matrix $\B{X}_g^T \B{X}_g$. Then it holds a.s.:
	$$ \sup \limits_{g=1, \ldots G} \mu_{\max} (\B{X}_g^T \B{X}_g)  \le n.$$
\end{asu}

\smallskip
\noindent
\textit{\textbf{Restricted eigenvalue conditions: }} We reuse Assumptions \ref{asu4}$.2(k, \gamma)$ and \ref{asu4}$.4(s, \epsilon)$ to study Problems \eqref{lasso} and \eqref{group-lasso}. We simply replace $\nabla^2 \mathcal{L}(\B{\beta}^* )$ by the Hessian of the least-squares loss $\frac{1}{n} \B{X}^T \B{X}$.

\subsection{Cone conditions}\label{sec: cone condition-regression}

The proof of Theorem \ref{cone-condition}  can be adapted to derive two cone conditions satisfied by a Lasso and a Group Lasso estimator, respectively solutions of Problem \eqref{lasso} and \eqref{group-lasso}.

\begin{theorem} \label{cone-condition-regression}
	Let $\delta \in \left(0, 1 \right)$, $\alpha\ge 2$ and assume that Assumption \ref{asu-col} holds.
	\newline
	As previously, we denote $\lambda_j^{(r)} =  \sqrt{  \log(2re/j) }, \forall j, \forall r$ and fix the parameters $\eta = 24 \alpha \sigma \sqrt{\frac{\log(2/ \delta)}{n}}$, $\lambda = \eta \lambda_{k^*}^{(p)}$ for Lasso and $\lambda_G = \eta \lambda_{s^*}^{(G)} + \alpha \sigma\sqrt{ \frac{\gamma m^*}{s^* n} }$ for Group Lasso.
	\newline
	The following results holds with probability at least $1 - \delta$.
	
	\noindent
	$\bullet$ Let $\hat{\B{\beta}}_1$ be a solution of the Lasso Problem \eqref{lasso} with parameter $\lambda$ and let $S_0\subset \{1,\ldots,p\}$ be the subset of indexes of the $k^*$ highest coefficients of $\B{h}_1:= \hat{\B{\beta}}_1 - \B{\beta}^*$. It holds:
	$$\B{h}_1 \in \Lambda \left( S_0, \ \gamma_1^*:= \frac{\alpha}{\alpha -1},  \ \gamma_2^*:= \frac{\sqrt{k^*}}{\alpha -1} \right).$$
		
	\noindent
	$\bullet$  Let $\hat{\B{\beta}}_{L1-L2}$ be a solution of the Group Lasso Problem \eqref{group-lasso} with parameter $\lambda_G$. Let $\mathcal{J}_0 \subset \{1,\ldots,G\}$ be the subset of indexes of the $s^*$  highest groups of $\B{h}_{L1-L2} :=\hat{\B{\beta}}_{L1-L2}  - \B{\beta}^*$ for the L2 norm. We additionally denote $m_0$ the total size of these $s^*$ groups and assume  $m_0 \le \gamma m^*$ for some $\gamma \ge 1$. It then holds:
	$$\B{h}_{L1-L2} \in \Omega \left( \mathcal{J}_0, \; \epsilon^*_1:=\frac{\alpha}{\alpha - 1}, \; \epsilon^*_2:=\frac{\sqrt{s^*}}{\alpha - 1} \right).$$
\end{theorem}
The proof is presented in Appendix \ref{sec: appendix_cone-condition-regression}. Again, we use our new Lemma \ref{upper-bound-sup} (cf. Appendix \ref{sec: appendix_lemma_bound}) to control the maximum of sub-Gaussian random variables. As a consequence, the regularization parameter $\lambda^2$ for Lasso is of the order of $\log(p/k^*) /n $ and matches prior results \citep{bellec2018slope}. For Group Lasso, our parameter $\lambda_G^2$ is of the order of $\log(G/s^*) / n + m^* /  (s^* n)$ and improve over \citet{huang2010benefit}.

\subsection{Upper bounds for L2 coefficients estimation}
We present our main bounds for Lasso and Group Lasso.
\begin{theorem} \label{main-results-regression}
	Let $\delta \in \left(0, 1 \right), ~ \alpha \ge 2$. We consider the same assumptions and notations than Theorem \ref{cone-condition-regression}.  In addition, we assume that Assumption \ref{asu4}.1$(k^*, \gamma^*)$ hold for Lasso and Assumption \ref{asu4}.2$(s^*, \epsilon^*)$ holds for Group Lasso.
	
	\noindent
	The Lasso estimator satisfies with probability at least $1-\delta$:
	\begin{equation*}
		\begin{split}   
		\| \B{h}_1 \|_2 \lesssim &
		\frac{ \alpha \sigma }{\kappa^*} \sqrt{ \frac{ k^* \log\left( p/k^* \right) \log\left( 2/\delta \right) }{n}}.
		\end{split}
	\end{equation*}
	In addition, the Group Lasso estimator satisfies with same probability:
	\begin{equation*}
		\begin{split}
		&\| \B{h}_{L1-L2} \|_2 \lesssim 
		\frac{ \alpha \sigma }{\kappa^*} \sqrt{ \frac{ s^* \log\left( G/s^* \right) \log\left( 2/\delta \right) + \gamma m^* }{n}}.
		\end{split}
	\end{equation*}
	where $\kappa^* = \kappa \left(S_0, \gamma_1^*, \gamma_2^* \right)$ for Lasso and $\kappa^* = \kappa \left(\mathcal{J}_0, \epsilon_1^*, \epsilon_2^*\right)$ for Group Lasso.
\end{theorem}
The proof is presented in Appendix \ref{sec: appendix_main-results-regression}. It follows from the cone conditions and the use of the restricted eigenvalue assumptions. In addition, we obtain by integration our bounds in expectation, presented in Appendix \ref{sec: appendix_main-corollary-regression}.

\medskip
\noindent
\textbf{Discussion: } For Lasso, we match the best rate known \citep{bellec2018slope}. For Group-Lasso, our rate improves over existing results \citep{huang2010benefit}. Note that a comparative study of both bounds would give similar observations to the ones discussed in Section \ref{sec: error-bound}.

\subsection{Summary}
The following Table \ref{table:sota} summarizes the main results and novelties of this paper, for the principal estimators of interest. 
\begin{table}[h!]
	\centering
	\vspace{-.5em}
	\caption{Error bounds derived in this work for the main estimators of interest, as a function of the problem parameters. Rates with a $~\star~$ strictly improve over the best previously known results. Rates with a $~\star \star~$ are, to the best of our knowledge, new results.}
	\smallskip
	\begin{tabular}{p{0.25\textwidth}p{0.2\textwidth}p{0.3\textwidth}}
		\toprule
		Loss  & Regularization & Rate \\
		\midrule
		logistic, hinge, quantile & L1 & $(k^*/n) \log(p/k^*) \newline \star$ \\
		\midrule
		logistic, hinge, quantile & Slope & $(k^*/n) \log(p/k^*) \newline \star \star$ \\
		\midrule
		logistic, hinge, quantile & Group L1-L2 & $(s^*/n) \log\left( G / s^* \right) + m^* / n \newline \star \star$ \\
		\midrule
		least-squares & L1 & $(k^*/n) \log(p/k^*) $ \\
		\midrule
		least-squares & Group L1-L2  & $(s^*/n) \log\left( G / s^* \right) + m^* / n \newline  \star$\\
		\bottomrule
	\end{tabular}
	\label{table:sota}
	\vspace{-1em}
\end{table}

\section{Empirical analysis}\label{sec: First-Order}

\subsection{First order algorithm}

All the estimators studied in Section \ref{sec: error-bound} are convex. In particular, each one of our main estimators of interest pairs one of the three main losses that fall into our framework 
with one of the three regularizations studied---the L1 regularization, the Slope regularization or the Group L1-L2 regularization\footnote{We do not consider here the least-squares loss, as it has been widely empirically studied in the litterature \citep{slope-introduction,wang2013l1,tibshirani1996regression,huang2010benefit}.}. However, there is no existing general package that can be used for a fast empirical study of these nine estimators. Therefore, in Appendix \ref{sec:first-order}, we propose a proximal gradient algorithm which solves the tractable Problems \eqref{l1-problem}, \eqref{slope-problem} and \eqref{group-problem} when the number of variables is of the order of $100,000s$---compatible with modern applications and datasets in machine learning. Our proposed method (i) smoothes the non-convex hinge and regression losses (ii) applies a thresholding operator for all three regularizations considered and (iii) achieves a convergence rate of $O(1/\epsilon)$ to obtain an $\epsilon$-accurate solution.

Although the idea of pairing smoothing with first order methods has been proposed \citep{beck2012smoothing}, we release an effective implementation of these nine estimators, and propose an empirical evaluation of their performance. Our code is provided with the Supplementary materials.

\subsection{Empirical analysis}
Appendix \ref{sec:simu} proposes a numerical study that compares the estimators studied with standard non-sparse baselines for computational settings where the signal is either sparse or group-sparse, and the number of variables is of the order of $100,000s$. Our numerical findings (i) show that our algorithms scale to large datasets and (ii) enhance the empirical performance of our estimators for various settings. 

\newpage
\bibliography{arxiv}

\newpage
~
\newpage
 \setcounter{page}{1}

\begin{appendices}

	\section{Usefull properties of sub-Gaussian random variables}
	
	This section presents useful preliminary results satisfied by sub-Gaussian random variables. In particular, Lemma \ref{upper-bound-sup} provides a probabilistic upper-bound on the maximum of sub-Gaussian random variables.
	
	\subsection{Preliminary results}
	
	Under Assumption \ref{asuSG}, the random variables $\partial f\left( \langle \B{x}_i,  \B{\beta}^*  \rangle, y_i  \right)  x_{ij},  \ \forall i,j$ are sub-Gaussian.  They all consequently satisfy the next Lemma  \ref{lemma-lecture-notes}:
	
	\begin{lemma} \label{lemma-lecture-notes}
		Let $Z \sim \subG(\sigma^2)$ for a fixed $\sigma>0$. Then for any $t>0$ it holds 
		$$\mathbb{P}( |Z| > t) \le 2\exp \left( \frac{-t^2}{2 \sigma^2} \right). $$
		\medskip
		\noindent
		In addition, for any positive integer $\ell \ge 1$ we have:
		$$\mathbb{E}\left( | Z |^{\ell} \right) \le  (2\sigma^2)^{\ell/2} \ell \Gamma(\ell/2)$$
		where $\Gamma$ is the Gamma function defined as  $\Gamma(t) =  \int_{0}^{\infty} x^{t-1} e^{-x} dx, \ \forall t>0.$ 
		
		\medskip
		\noindent
		Finally, let $Y=Z^2 - \mathbb{E}(Z^2)$ then we have
		\begin{equation}\label{eq-lemma-preliminary}
		\mathbb{E} \left( \exp\left( \frac{1}{16 \sigma^2} Y \right) \right) \le \frac{3}{2},
		\end{equation}
		and as a consequence $\mathbb{E} \left( \exp\left( \frac{1}{16 \sigma^2} Z^2\right)  \right) \le 2.$
	\end{lemma}
	
	\begin{Proof}
		The first two results correspond to Lemmas 1.3 and 1.5 from \citet{lecture-notes}.  
		\newline
		In particular $\mathbb{E}\left( | Z |^2 \right) \le  4\sigma^2$. 
		
		\medskip
		\noindent
		In addition, using the proof of Lemma 1.12 we have:
		$$\mathbb{E} \left( \exp(tY ) \right) \le 1 + 128 t^2 \sigma^4, \ \forall |t| \le \frac{1}{16\sigma^2}.$$
		Equation \eqref{eq-lemma-preliminary} holds in the particular case where $t=1/16\sigma^2.$ 
		The last part of the lemma combines our precedent results with the observation that $\frac{3}{2} e^{1/4} \le 2$.
	\end{Proof}
	
	\medskip
	\noindent
	We will also need the following Lemma \ref{lemma-asu3}.
	\begin{lemma} \label{lemma-asu3}
		Let $\left\{ (\B{x_i},y_i) \right\}_{i=1}^n$, $( \B{x_i},y_i) \in \mathbb{R}^p\times \mathcal{Y}$ be independent samples from an unknown distribution. Let $f$ be a loss satisfying Assumption \ref{asu1} and $\B{\beta}^*$ be a theoretical minimizer of $f$. If Assumption \ref{asuSG} is satisfied, then it holds $$\frac{1}{\sqrt{n}} \sum_{i=1}^n  \partial f\left( \langle \B{x}_i,  \B{\beta}^*  \rangle; y_i  \right)  x_{ij} \sim \subG(L^2 M^2), \forall j.$$
	\end{lemma}
	
	\begin{Proof}
		We note $S_i = \partial f\left( \langle \B{x}_i,  \B{\beta}^*  \rangle, y_i  \right), \ \forall i $. 
		\noindent
		Since $\B{\beta}^*$ minimizes the theoretical loss, we have $\mathbb{E}(S_i x_{ij})=0, \ \forall i,j$. 
		\noindent
		We fix $M>0$ such that:
		$$\mathbb{E}\left( \exp(t S_i x_{ij} ) \right) \le e^{L^2 M^2 t^2 / 2}, \ \forall t>0,  \forall i,j.$$
		Since the samples are independent, it holds $\forall t>0,$
		\begin{align*}
		\begin{split}
		\mathbb{E}\left( \exp \left( \frac{t}{\sqrt{n}} \sum_{i=1}^n S_i x_{ij} \right) \right) 
		= \prod_{i=1}^n  \mathbb{E}\left( \exp \left( \frac{t}{\sqrt{n}}  S_i  x_{ij} \right) \right) 
		\le \prod_{i=1}^n e^{ L^2 M^2 t^2 /2n}  = e^{L^2 M^2 t^2 / 2},
		\end{split}
		\end{align*}
		which concludes the proof.
	\end{Proof}

	\subsection{A bound for the maximum of sub-Gaussian variables} \label{sec: appendix_lemma_bound}
	
	As a second consequence of Lemma \ref{lemma-lecture-notes}, the next two technical lemmas derive a probabilistic upper-bound for the maximum of sub-Gaussian random variables. Lemma \ref{lemma-Slope} is an extension for sub-Gaussian random variables of Proposition E.1 \citep{bellec2018slope}.
	\begin{lemma}\label{lemma-Slope}
		Let $g_1,\ldots g_r$ be sub-Gaussian random variables with variance $\sigma^2$. Denote by $(g_{(1)}, \ldots, g_{(r)})$ a non-increasing rearrangement of $(|g_1|, \ldots, |g_r|)$.  Then $\forall t>0$ and $\forall j \in \left\{1, \ldots,r \right\}$:
		$$ \mathbb{P}\left( \frac{1}{j \sigma^2} \sum_{k=1}^j g_{(k)} ^2 > t \log\left( \frac{2r}{j} \right) \right) \le \left( \frac{2r}{j} \right)^{1- \frac{t}{16}}. $$
	\end{lemma}
	
	\begin{Proof}
		We first apply a Chernoff bound: 
		\begin{align*}
		\begin{split}
		\mathbb{P}\left( \frac{1}{j\sigma^2} \sum_{k=1}^j g_{(k)} ^2 > t \log\left( \frac{2r}{j} \right) \right) 
		\le \mathbb{E} \left( \exp \left( \frac{1}{16 j \sigma^2} \sum_{k=1}^j g_{(k)} ^2 \right)  \right)   \left( \frac{2r}{j} \right)^{-\frac{t}{16}}.
		\end{split}
		\end{align*}
		We then use Jensen inequality to obtain
		\begin{align*}
		\begin{split}
		\mathbb{E} \left( \exp\left( \frac{1}{16 j \sigma^2} \sum_{k=1}^j g_{(k)} ^2 \right) \right)
		\le \frac{1}{ j} \sum_{k=1}^j \mathbb{E} \left( \exp\left( \frac{1}{16 \sigma^2} g_{(k)}^2 \right) \right) 
		& \le \frac{1}{ j} \sum_{k=1}^r \mathbb{E} \left( \exp\left( \frac{1}{16 \sigma^2}  g_{k}^2 \right) \right) \\
		& \le \frac{2r}{j} \text{  with Lemma \ref{lemma-lecture-notes}}.
		\end{split}
		\end{align*}
	\end{Proof}
	
	\noindent
	Using Lemma \ref{lemma-Slope}, we can derive the following bound holding with high probability:
	\begin{lemma}\label{upper-bound-sup}
		We consider the same assumptions and notations than Lemma \ref{lemma-Slope}. In addition, we define the coefficients $\lambda_j^{(r)} = \sqrt{ \log(2r/j) }, \ j=1,\ldots\,r$ similar to Theorem \ref{cone-condition}. Then for $\delta \in \left(0, \frac{1}{2} \right)$, it holds with probability at least $1-\delta$:
		$$ \sup \limits_{j=1,\ldots,r} \left\{ \frac{ g_{(j)} }{\sigma \lambda_j^{(r)}} \right\} \le 12 \sqrt{ \log(1 / \delta)}.  $$ 
	\end{lemma}
	
	\begin{Proof}
		We fix $\delta \in \left(0, \frac{1}{2} \right)$ and $j \in \left\{1, \ldots,r \right\}$. We upper-bound $g_{(j)}^2$ by the average of all larger variables:
		$$g_{(j)}^2 \le \frac{1}{j} \sum_{k=1}^j g_{(k)}^2. $$
		Applying Lemma \ref{lemma-Slope} gives, for $t>0$:
		\begin{align*}
		\begin{split}
		\mathbb{P}\left( \frac{ g_{(j)}^2 }{\sigma^2 \left(\lambda_j^{(r)} \right)^2}  > t  \right) &\le \mathbb{P}\left( \frac{1}{j \sigma^2} \sum_{k=1}^j g_{(k)} ^2 > t  \left(\lambda_j^{(r)} \right)^2 \right) \le \left( \frac{j}{2r} \right)^{\frac{t}{16} -1 }.
		\end{split}
		\end{align*}
		We fix $t = 144 \log(1/\delta)$ and use an union bound to get:
		\begin{align*}
		\begin{split}
		\mathbb{P}\left( \sup \limits_{j=1,\ldots,r} \frac{  g_{(j)}  }{\sigma \lambda_j^{(r)}}  > 12 \sqrt{\log(1/\delta) }  \right) 
		\le \left( \frac{1}{2r} \right)^{9 \log(1/\delta) -1 }  \sum_{j=1}^r j^{9 \log(1/\delta)  -1 }. 
		\end{split}
		\end{align*}
		Since $\delta < \frac{1}{2}$ it holds that $9 \log(1/\delta)  -1\ge 9 \log(2) -1 > 0$, then the map $t>0 \mapsto t^{9 \log(1/\delta)  -1 }$ is increasing. An integral comparison gives:
		$$\sum_{j=1}^r j^{9 \log(1/\delta)  -1 } \le \frac{1}{2} \left( r+1 \right)^{9 \log(1/\delta) } = \frac{1}{2} \delta^{-9 \log(r+1)}.$$
		In addition  $9 \log(1/\delta) -1 \ge 7 \log(1/\delta)$ and 
		$$\left( \frac{1}{2r} \right)^{9 \log(1/\delta) -1 } \le \left( \frac{1}{2r} \right)^{ -7 \log(\delta)  } = \delta^{ 7\log(2r)}.$$
		Finally, by assuming $r\ge2$, then we have $ 7\log(2r) - 9\log(r+1) > 1$ and we conclude: 
		$$ \mathbb{P}\left( \sup \limits_{j=1,\ldots,r} \frac{  g_{(j)}  }{\sigma \lambda_j^{(r)}}  > 12 \sqrt{\log(1/\delta) }  \right) \le \delta.$$
	\end{Proof}

	\section {Proof of Theorem \ref{cone-condition}}  \label{sec: appendix_cone-condition}
	We use the minimality of  $\hat{\B{\beta}}$ and Lemma \ref{lemma-Slope} to derive the cone conditions. 
	
	\begin{Proof}   
		We first consider a general solution of Problem \eqref{general} with regularization $\Omega(.)$ before studying the cases of L1, Slope and Group L1-L2 regularizations.
		
		\smallskip
		\noindent
		$\hat{\B{\beta}}$ is the solution of the learning Problem \eqref{general} hence:
		\begin{align}\label{inf-equation}
		\begin{split}
		&\frac{1}{n} \sum_{i=1}^n f \left( \langle \B{x_i},  \hat{\B{\beta}} \rangle ;  y_i \right) + \Omega( \hat{\B{\beta}} )  \le \frac{1}{n} \sum_{i=1}^n f \left( \langle \B{x_i},  \B{\beta}^* \rangle ;  y_i \right) +  \Omega( \B{\beta}^* ).
		\end{split}
		\end{align}
		Similar to Theorem \ref{hoeffding-sup}, we define $ \Delta \left(\B{\beta}^*, \B{h} \right)= \frac{1}{n} \sum_{i=1}^n f \left( \langle \B{x_i},  \hat{\B{\beta}} \rangle ;  y_i \right) -  \frac{1}{n} \sum_{i=1}^n f \left( \langle \B{x_i},  \B{\beta}^* \rangle ;  y_i \right)$.
		
		\smallskip
		\noindent
		Equation \eqref{inf-equation} can be written in a compact form as:
		$$\Delta \left(\B{\beta}^*, \B{h} \right) \le \Omega( \B{\beta}^* ) - \Omega (\hat{\B{\beta}}).$$
		We lower bound $\Delta \left(\B{\beta}^*, \B{h} \right)$ by exploiting the existence of a bounded sub-Gradient $\partial f $:
		$$\Delta \left(\B{\beta}^*, \B{h} \right)  \ge S \left(\B{\beta}^*, \B{h} \right) :=  \frac{1}{n} \sum_{i=1}^n \partial f \left( \langle \B{x_i},  \B{\beta}^*  \rangle ;  y_i \right) \langle \B{x_i},  \B{h}  \rangle.$$
		We now consider each regularization separately.

		\paragraph{L1 regularization: } 
		For L1 regularization, we have:
		\begin{align*}
		\begin{split}
		|  S \left(\B{\beta}^*, \B{h} \right) | 
		&=\left|  \frac{1}{n} \sum_{i=1}^n \sum_{j=1}^{p} \partial f \left( \langle \B{x_i},  \B{\beta}^*  \rangle ;  y_i \right)   x_{ij} h_j   \right| \\
		&\le \frac{1}{\sqrt{n}}  \sum_{j=1}^{p} \left(   \frac{1}{\sqrt{n} } \left| \sum_{i=1}^n \partial f \left( \langle \B{x_i},  \B{\beta}^*  \rangle ;  y_i \right)  x_{ij} \right|   \right) |h_j|.
		\end{split}
		\end{align*}
		Let us define the random variables $ g_j =  \frac{1}{\sqrt{n} } \sum_{i=1}^n \partial f \left( \langle \B{x_i},  \B{\beta}^*  \rangle ;  y_i \right)  x_{ij} , \ j=1,\ldots,p.$
		
		\smallskip
		\noindent
		Under Assumption \ref{asuSG}, Lemma \ref{lemma-asu3} guarantees that $g_1,\ldots,g_p$ are sub-Gaussian with variance $L^2 M^2$. A first upper-bound of the quantity $| S \left(\B{\beta}^*, \B{h} \right) | $ could be obtained by considering the maximum of the sequence $\left\{g_j\right\}$. However, Lemma \ref{upper-bound-sup} gives us a stronger result.  We note $\lambda_j = \lambda_j^{(p)}$ where we drop the dependency upon $p$. 
		
		\smallskip
		\noindent
		Since $\frac{\delta}{2} \le \frac{1}{2}$ we introduce a non-increasing rearrangement $(g_{(1)}, \ldots, g_{(p)})$  of $(|g_1|, \ldots, |g_p|)$. We recall that $S_0 = \left\{1,\ldots, k^*\right\}$ denotes the subset of indexes of the $k^*$ highest elements of $\B{h}$ and we  use Lemma \ref{upper-bound-sup} to get, with probability at least $1-\frac{\delta}{2}$:
		\begin{align}\label{upper-bound-SG}
		\begin{split}
		|  S \left(\B{\beta}^*, \B{h} \right) | &\le \frac{1}{\sqrt{n}} \sum_{j=1}^p | g_{j} | |h_{j} | = \frac{1}{\sqrt{n}} \sum_{j=1}^p g_{(j)} |h_{(j)} | = \frac{1}{\sqrt{n}}  \sum_{j=1}^p \frac{ g_{(j)} }{LM \lambda_j} LM \lambda_j | h_{(j)} | \\
		&\le \frac{1}{\sqrt{n}} \sup \limits_{j=1,\ldots,p} \left\{ \frac{ g_{(j)} }{LM \lambda_j} \right\} 
		\sum_{j=1}^p  LM \lambda_j | h_{(j)} | \\
		&\le 12 L M \sqrt{ \frac{\log(2/ \delta)}{n} } \sum_{j=1}^p \lambda_j | h_{(j)} | \textnormal{ with Lemma } \ref{upper-bound-sup}\\
		&\le 12 L M \sqrt{ \frac{\log(2/ \delta)}{n} } \sum_{j=1}^p \lambda_j | h_{j} | \textnormal{ since } \lambda_1\ge \ldots\ge \lambda_p \textnormal{ and } |h_1|\ge \ldots\ge |h_p| 
		\end{split}
		\end{align}
		To conclude, by pairing Equations  \eqref{inf-equation} and \eqref{upper-bound-SG} it holds:
		\begin{equation}\label{common}
		- 12 L M \sqrt{ \frac{\log(2/ \delta)}{n} } \sum_{j=1}^p \lambda_j | h_{j} | \le \lambda \| \B{\beta}^* \|_1 - \lambda \| \hat{\B{\beta}} \|_1.
		\end{equation}
		We refer to $A = - 12 L M \sqrt{ \frac{\log(2/ \delta)}{n} } \sum_{j=1}^p \lambda_j | h_{j} | $ and $ B = \lambda \| \B{\beta}^* \|_1 - \lambda \| \hat{\B{\beta}} \|_1$ as the respective left-hand and right-hand sides of Equation  \eqref{common}.
		
		\smallskip
		\noindent
		We assume without loss of generality that $|h_1| \ge \ldots \ge |h_p|$. We define $S_0 = \left\{1,\ldots,k^*\right\}$ as the set of the $k^*$ highest coefficients of $\B{h} =  \hat{\B{\beta}} - \B{\beta}^*$. Let  $S^*$ be the support of $\B{\beta}^*$. By definition of $S_0$ it holds:
		\begin{align}\label{inf-equation-sup}
		\begin{split}
		B
		\le  \lambda \| \B{\beta}^*_{S^*} \|_1 - \lambda \| \hat{\B{\beta}}_{S^*} \|_1 -\lambda \| \hat{\B{\beta}}_{(S^*)^c} \|_1
		&\le  \lambda \| \B{h}_{S^*} \|_1 -\lambda \| \B{h}_{(S^*)^c}  \|_1\\
		&\le  \lambda \| \B{h}_{S_0} \|_1 -\lambda \| \B{h}_{(S_0)^c}  \|_1.
		\end{split}
		\end{align}
		In addition, we lower bound the left-hand side of Equation  \eqref{common} by:
		\begin{align}\label{upper-bound-SG-slope}
		\begin{split}
		- A
		\le 12 L M \sqrt{ \frac{\log(2/ \delta)}{n} } \left( \sum_{j=1}^{k^*} \lambda_j | h_j |  +  \lambda_{k^*} \| \B{h}_{(S_0)^c}  \|_1  \right).
		\end{split}
		\end{align}
		Cauchy-Schwartz inequality leads to:
		\begin{align*}
		\sum_{j=1}^{k^*} \lambda_j | h_j |  &\le \sqrt{\sum_{j=1}^{k^*} \lambda_j ^2 } \| \B{h}_{S_0}  \|_2 \le \sqrt{k^*\log(2pe /k^*)}  \| \B{h}_{S_0}  \|_2,
		\end{align*}
		where we have used the Stirling formula to obtain
		\begin{align*}
		\sum_{j=1}^{k^*} \lambda_j ^2 = \sum_{j=1}^{k^*}  \log(2p/j) &= k^*  \log(2p) - \log(k^* !) \\
		&\le k^* \log(2p) - k^*\log(k^*/ e) = k^* \log(2pe /k^*).
		\end{align*}
		In the statement of Theorem \ref{cone-condition} we have defined $\lambda = 12 \alpha L M \sqrt{ \frac{1}{n} \log(2pe/k^*) \log(2/ \delta)}$. 
		
		\smallskip
		\noindent
		Because $\lambda_{k^*} \le  \sqrt{\log(2pe/k^*)}$, Equation \eqref{upper-bound-SG-slope} leads to:
		$$- A \le \frac{1}{\alpha}\lambda \left( \sqrt{k^*}  \| \B{h}_{S_0}  \|_2  +  \| \B{h}_{(S_0)^c}  \|_1 \right)$$
		Combined with Equation \eqref{inf-equation-sup}, it holds with probability at least $1-\frac{\delta}{2}:$
		$$ - \frac{\lambda}{\alpha} \left( \sqrt{k^*}  \| \B{h}_{S_0}  \|_2  +  \| \B{h}_{(S_0)^c}  \|_1 \right) \le  \lambda \| \B{h}_{S_0} \|_1 -\lambda \| \B{h}_{(S_0)^c}  \|_1,$$
		which immediately leads to:
		$$ \| \B{h}_{(S_0)^c}  \|_1 \le \frac{\alpha}{\alpha -1}  \| \B{h}_{S_0}  \|_1 + \frac{ \sqrt{k^*}}{\alpha -1}  \| \B{h}_{S_0}  \|_2.$$
		We conclude that $\B{h} \in \Lambda \left(S_0, \ \frac{\alpha}{\alpha -1}, \  \frac{\sqrt{k^*}}{\alpha -1} \right)$ with probability at least $1-\frac{\delta}{2}$.

		\paragraph{Slope regularization: }  For the Slope regularization, Equation  \eqref{common} still holds and the quantity $A$ is still defined. We define $B$ by replacing the L1 regularization with Slope. We still assume $|h_1| \ge \ldots \ge |h_p|$. To upper-bound $B$, we define a permutation $\phi \in \mathcal{S}_p$ such that $\| \B{\beta}^* \|_S = \sum_{j=1}^{k^*} \lambda_j | \beta^*_{\phi(j)} | $ and $| \hat{\beta}_{\phi(k^*+1)} | \ge \ldots \ge | \hat{\beta}_{\phi(p)} |$. It holds:
		\begin{align}\label{inequality-slope}
		\begin{split}
		\frac{1}{\eta}B = \frac{1}{\eta} \| \B{\beta}^* \|_{\mathcal{S}} - \frac{1}{\eta} \| \hat{\B{\beta}} \|_{\mathcal{S}}
		&= \sum_{j=1}^{k^*} \lambda_j | \beta^*_{\phi(j)} | - \max \limits_{\psi \in \mathcal{S}_p} \sum_{j=1}^p \lambda_j |  \hat{\beta}_{\psi(j)} |\\
		&\le \sum_{j=1}^{k^*} \lambda_j | \beta^*_{\phi(j)} | -  \sum_{j=1}^p \lambda_j |  \hat{\beta}_{\phi(j)} |\\
		&= \sum_{j=1}^{k^*} \lambda_j \left( | \beta^*_{\phi(j)} | -  | \hat{\beta}_{\phi(j)} | \right) -
		\sum_{j=k^*+1}^p \lambda_j  | \hat{\beta}_{\phi(j)} | \\
		&\le \sum_{j=1}^{k^*} \lambda_j | h_{\phi(j)} | - \sum_{j=k^*+1}^p \lambda_j  | h_{\phi(j)} | .
		\end{split}
		\end{align}
		Since $\left\{ \lambda_j \right\}$ is monotonically non decreasing: $\sum_{j=1}^{k^*} \lambda_j | h_{\phi(j)} | \le \sum_{j=1}^{k^*} \lambda_j | h_j |$.
		\smallskip 
		Because $| h_{\phi(k^*+1)} | \ge \ldots \ge | h_{\phi(p)} |$: $\sum_{j=k^*+1}^p \lambda_j  | h_j | \le \sum_{j=k^*+1}^p \lambda_j  | h_{\phi(j)} |$. 
		\smallskip 
		It consequently holds:
		\begin{equation}\label{last-inequality-slope}
		\frac{1}{\eta}B \le \sum_{j=1}^{k^*} \lambda_j | h_{j} | - \sum_{j=k^*+1}^p \lambda_j  | h_{j} |
		\end{equation}
		In addition, since $\eta =  12 \alpha L M \sqrt{ \frac{\log(2/ \delta)}{n} }$, we obtain with probability at least $1 - \frac{\delta}{2}$:
		$$A = - 12 L M \sqrt{ \frac{\log(2/ \delta)}{n} } \sum_{j=1}^p \lambda_j | h_j | = - \frac{\eta}{\alpha}  \| \B{h} \|_S.$$
		Thus, combining this last equation with Equation \eqref{last-inequality-slope}, it holds with probability at least $1 - \frac{\delta}{2}$:
		$$- \frac{1}{\alpha}  \| \B{h} \|_S \le  \sum_{j=1}^{k^*} \lambda_j | h_{j} | - \sum_{j=k^*+1}^p \lambda_j  | h_{j} |,$$
		which is equivalent to saying that with probability at least $1- \frac{\delta}{2}$:
		\begin{equation} 
		\sum_{j=k^*+1}^p \lambda_j | h_j |   \le \frac{\alpha+1}{\alpha-1} \sum_{j=1}^{k^*} \lambda_j | h_j |,
		\end{equation}
		that is $\B{h} \in \Gamma\left(k^*,  \frac{\alpha+1}{\alpha-1} \right)$.

		\paragraph{Group L1-L2 regularization: } 
		For Group L1-L2 regularization, we also introduce the vector of sub-Gaussian random variables $\B{g} = (g_1, \ldots, g_p)$ with $g_j =  \frac{1}{\sqrt{n} } \sum_{i=1}^n \partial f \left( \langle \B{x_i},  \B{\beta}^*  \rangle ;  y_i \right)  x_{ij}, \forall j$. We then have:
		\begin{align}\label{equations-group}
		|  S \left(\B{\beta}^*, \B{h} \right) | 
		=\frac{1}{\sqrt{n}} \left|  \langle  \B{g},   \B{h}  \rangle  \right|
		\le \frac{1}{\sqrt{n}}  \sum_{g=1}^G \left| \langle  \B{g}_g, \B{h}_g  \rangle  \right|
		\le \frac{1}{\sqrt{n}}  \sum_{g=1}^G \| \B{g}_g \|_2 \|  \B{h}_g  \|_2,
		\end{align}
		where we have used Cauchy-Schwartz inequality on each group.
		
		\medskip
		\noindent
		We have denoted $n_g$ the cardinality of the set of indexes $\mathcal{I}_g$ of group $g$ and $\B{n}=(n_1, \ldots, n_G)$. 
		Let us fix $g \le G, \B{u}_g \in \mathbb{R}^{n_g}$. As the variable $\partial f\left( \langle \B{x}_i,  \B{\beta}^*  \rangle, y_i  \right) (\B{x}_i)_g$ is sub-Gaussian with variance $\sigma^2$, it holds:
		$$\mathbb{E}\left( \exp \left(t ~ \partial f\left( \langle \B{x}_i,  \B{\beta}^*  \rangle, y_i  \right) \right) (\B{x}_i)_g^T \B{u}_g \right) \le \exp \left( \frac{L^2 M^2 t^2 \| \B{u}_g \|_2^2}{2} \right) \ \forall t>0,  \forall i.$$
		As a consequence, since the rows of the design matrix are independent, it holds:
		\begin{align}\label{required-hsu}
		\begin{split}
		\mathbb{E}\left( \exp \left( \B{g}_g^T \B{u}_g \right) \right) &=\mathbb{E}\left( \exp \left( \frac{1}{\sqrt{n}} \sum_{i=1}^n \partial f\left( \langle \B{x}_i,  \B{\beta}^*  \rangle, y_i  \right)  (\B{x}_{i})_g^T \B{u}_g \right) \right) \\
		&= \prod_{i=1}^n  \mathbb{E}\left( \exp \left( \frac{1}{\sqrt{n}}  \partial f\left( \langle \B{x}_i,  \B{\beta}^*  \rangle, y_i  \right) (\B{x}_i)_g^T \B{u}_g \right) \right) \\
		&\le \prod_{i=1}^n  \exp \left( \frac{L^2 M^2  \| \B{u}_g \|_2^2}{ 2n } \right)\\
		&=  \exp \left( \frac{L^2 M^2 \| \B{u}_g \|_2^2}{2} \right).
		\end{split}
		\end{align}
		We can then use Theorem 2.1 from \citet{hsu2012tail}. By denoting $\B{I}_g$ the identity matrix of size $n_g$ it holds:
		$$\mathbb{P}\left( \| \B{I}_g \B{g}_g  \|_2^2 \ge L^2 M^2 \left( \tr(\B{I_g}) + 2 \sqrt{\tr(\B{I_g}^2)  t} + 2 ||| \B{I_g} ||| t \right) \right) \le e^{-t},$$
		which gives
		$$\mathbb{P}\left( \| \B{g}_g \|_2 - LM \sqrt{n_g}  \ge L M \sqrt{2t} \right) = \mathbb{P}\left( \| \B{g}_g \|_2^2 \ge L^2 M^2 \left( \sqrt{n_g} + \sqrt{2 t} \right)^2 \right) \le e^{-t},$$
		which is equivalent from saying that:
		\begin{equation}\label{subGauss-group}
		\mathbb{P}\left( \| \B{g}_g \|_2^2 -  LM \sqrt{n_g}  \ge t \right)  \le \exp\left( \frac{-t^2}{2 L^2 M^2} \right).
		\end{equation}
		Let us define the random variables
		$f_g = \max \left(0,  \| \B{g}_g \|_2 -  LM \sqrt{n_g} \right), \ g=1,\ldots,G$. Equation \eqref{subGauss-group} shows that $f_g$ satisfies the same tail condition than a sub-Gaussian random variable with variance $L^2 M^2$ and we can apply Lemma \ref{upper-bound-sup}. In addition, following Equation \eqref{equations-group} it holds:
		$$| S \left(\B{\beta}^*, \B{h} \right) | \le \frac{1}{\sqrt{n}}  \sum_{g=1}^G \left( \| \B{g}_g \|_2 -  LM  \sqrt{n_g}  \right) \| \B{h}_g  \|_2 + \frac{1}{\sqrt{n}}  \sum_{g=1}^G  LM  \sqrt{n_g} \| \B{h}_g  \|_2.$$
		We introduce a non-increasing rearrangement $(f_{(1)}, \ldots, f_{(G)})$  of $(|f_1|, \ldots, |f_G|)$. 
		In addition, we assume without loss of generality that $\| \B{ h }_1 \|_{2} \ge \ldots \| \B{ h }_G \|_{2}$. 
		We have defined $\mathcal{J}_0 = \left\{1,\ldots, s^*\right\}$ as the subset of indexes of the $s^*$ groups of $\B{h}$  with highest L2 norm.
		We define a permutation $\psi$ such that $n_{\psi(1)} \ge \ldots \ge n_{\psi(G)}$. Similar to the above, Lemma \ref{upper-bound-sup} gives with probability at least $1-\frac{\delta}{2}$---we use the coefficients $\lambda_g^{(G)} = \sqrt{\log(2Ge / g)}  $:
		
		\begin{align}\label{upper-bound-SG-group}
		\begin{split}
		| S \left(\B{\beta}^*, \B{h} \right) | 
		&\le \frac{1}{\sqrt{n}}  \sum_{g=1}^G \left( \| \B{g}_g \|_2 -  LM \sqrt{n_g}  \right) \|  \B{h}_g  \|_2 + \frac{ LM }{\sqrt{n}}  \sum_{g=1}^G \sqrt{n_g} \|  \B{h}_g  \|_2 \\
		&\le \frac{1}{\sqrt{n}} \sum_{g=1}^G |f_{g} |  \|  \B{h}_g ||_2 + \frac{ LM }{\sqrt{n}}  \sum_{g=1}^G \sqrt{n_g} \|  \B{h}_g  \|_2\\
		&=\frac{1}{\sqrt{n}} \sum_{g=1}^G \frac{f_{(g)}}{LM \lambda_g^{(G)} }   LM \lambda_g^{(G)} \|  \B{h}_{(g)} ||_2 + \frac{ LM }{\sqrt{n}}  \sum_{g=1}^G \sqrt{n_g} \|  \B{h}_g  \|_2\\
		&\le 12 LM \sqrt{ \frac{\log(2/ \delta)}{n} } \sum_{g=1}^G \lambda_g^{(G)} \| \B{h}_{(g)} \|_{2} + \frac{ LM }{\sqrt{n}}  \sum_{g=1}^G \sqrt{n_{g}} \|  \B{h}_{g}  \|_2\\
		&\le 12 LM  \sqrt{ \frac{\log(2/ \delta)}{n} } \sum_{g=1}^G \lambda_g^{(G)} \|  \B{h}_{g} \|_{2} + \frac{LM}{\sqrt{n}}  \sum_{g=1}^G \sqrt{n_{\psi(g)}} \| \B{h}_{g} \|_{2} \\
		&~~~\text{ since } \lambda_1^{(G)} \ge \ldots \ge \lambda_G^{(G)},  ~ \| \B{ h }_1 \|_{2} \ge \ldots \| \B{ h }_G \|_{2} \text{ and } n_{\psi(1)} \ge \ldots \ge n_{\psi(G)} \\
		&\le 12LM \sqrt{ \frac{\log(2/ \delta)}{n} } \left(  \sqrt{s^*\log(2Ge /s^*)}   \left( \sum_{g \in \mathcal{J}_0} \| \B{h}_{g} \|_{2}^2 \right)^{1/2} + \lambda_{s^*}^{(G)} \sum_{g \notin \mathcal{J}_0} \| \B{h}_{g} \|_{2} \right)\\
		&~~~ + \frac{LM}{\sqrt{n}}   \left( \left( \sum_{g \in \mathcal{J}_0} n_{\psi(g)} \right)^{1/2} \left( \sum_{g \in \mathcal{J}_0} \| \B{h}_{g} \|_{2}^2 \right)^{1/2}  + \max_{g= s^* + 1, \ldots, G} \sqrt{n_{\psi(g)}} \sum_{g \notin \mathcal{J}_0 } \| \B{h}_{g} \|_{2} \right)\\ 
		&\le 12 L M \sqrt{ \frac{\log(2Ge /s^*)}{n} \log(2/ \delta) } \left(  \sqrt{s^*}   \| \B{h}_{\mathcal{T}_0} \|_{2} + \sum_{g \notin \mathcal{J}_0} \| \B{h}_{g} \|_{2} \right)\\
		&~~~ + \frac{LM}{\sqrt{n}}  \left( \sqrt{m_0} \| \B{h}_{\mathcal{T}_0} \|_{2}  + \sqrt{\frac{m_0}{s^*}} \sum_{g \notin \mathcal{J}_0 } \| \B{h}_{g} \|_{2} \right) \text{ since } \frac{m_0}{s^*} \ge n_{\psi(s^* + 1)} \ge \ldots \ge n_{\psi(G)} \\ 
		&\le \left(12 L M \sqrt{ \frac{\log(2Ge /s^*)}{n} \log(2/ \delta) } + LM \sqrt{ \gamma \frac{m^* / s^*}{n} } \right) \left(  \sqrt{s^*}   \| \B{h}_{\mathcal{T}_0} \|_{2} + \sum_{g \notin \mathcal{J}_0} \| \B{h}_{g} \|_{2} \right),
		\end{split}
		\end{align}
		where $\mathcal{T}_0 = \cup_{g \in \mathcal{J}_0} \mathcal{I}_g$ has been defined as the subset  of all indexes across the $s^*$ groups in $\mathcal{J}_0$, $m_0$ is the size of the $s^*$ largest groups, and the Stirling formula gives: $\sum \limits_{g=1}^{s^*} \left(\lambda_g^{(G)} \right)^2 \le s^*\log(2Ge /s^*)$.
		
		\smallskip
		\noindent
		We have defined  $\lambda_G = 12 \alpha L M \sqrt{ \frac{1}{n} \log(2Ge/s^*) \log(2/ \delta)} + \alpha LM \sqrt{ \frac{\gamma m^*}{s^* n} }$ and $\mathcal{J}^* \subset \{1,\ldots,G\}$ as the smallest subset of group indexes such that the support of $\B{\beta}^*$ is included in the union of these groups. By pairing Equations  \eqref{inf-equation} and \eqref{upper-bound-SG-group} it holds:
		\begin{equation} 
		\begin{split}
		-\frac{\lambda_G}{\alpha}  \left(  \sqrt{s^*}   \| \B{h}_{\mathcal{T}_0} \|_{2} + \sum_{g \notin \mathcal{J}_0} \| \B{h}_{g} \|_{2} \right) 
		&\le \lambda_G \sum_{g \in \mathcal{J}^*} \| \B{h}_g \|_{2} - \lambda_G \sum_{g \notin \mathcal{J}^*} \| \B{h}_g \|_{2}\\
		&\le \lambda_G \sum_{g \in \mathcal{J}_0} \| \B{h}_g \|_{2} - \lambda_G \sum_{g \notin \mathcal{J}_0} \| \B{h}_g \|_{2},
		\end{split}
		\end{equation}
		which is equivalent to saying that with probability at least $1- \frac{\delta}{2}$:
		\begin{equation} 
		\sum_{g \notin \mathcal{J}_0} \| \B{h}_g \|_{2}   \le  \frac{\alpha}{\alpha - 1}  \sum_{g \in \mathcal{J}_0} \| \B{h}_g \|_{2} +  \frac{\sqrt{s^*} }{\alpha - 1}  \| \B{h}_{\mathcal{T}_0} \|_{2},
		\end{equation}
		that is $\B{h} \in \Omega\left(\mathcal{J}_0, \frac{\alpha}{\alpha-1}, \frac{\sqrt{s^*} }{\alpha - 1} \right)$.
	\end{Proof}

	\section {Proof of Theorem \ref{restricted-strong-convexity} }  \label{sec: appendix_restricted-strong-convexity}
	
	The restricted strong convexity conditions presented in Theorem \ref{restricted-strong-convexity} are a consequence of Theorem \ref{hoeffding-sup}, which derives a control of the supremum of the difference between an empirical random variable and its expectation. This supremum is controlled over a bounded set of sequences of length $q$ of $m$ sparse vectors with disjoint supports. Its proof is presented in Appendix \ref{sec: hoeffding-sup}. It uses Hoeffding's inequality to obtain an upper bound of the inner supremum for any sequence of $m$ sparse vectors. The result is extended to the outer supremum with an $\epsilon$-net argument. We first  prove Theorem \ref{restricted-strong-convexity} before Theorem \ref{hoeffding-sup}.
	\begin{Proof} 
		The proof of Theorem \ref{restricted-strong-convexity} is divided in two steps. First, we lower-bound the quantity $\Delta\left(\B{\beta}^*, \B{h} \right) $ using a partition of $\left\{1, \ldots, p\right\}$  and applying Theorem \ref{hoeffding-sup}. Second, we apply the cone conditions derived in Theorem \ref{cone-condition} to use the restricted eigenvalue conditions from Assumption \ref{asu4}.

		\paragraph{Step 1:} 
		First, let us fix a partition $S_1, \ldots, S_q$ of $\left\{1, \ldots, p\right\}$ such that $|S_{\ell}| \le m, \ \forall \ell$ and define the corresponding sequence $\B{h}_{S_1}, \ldots, \B{h}_{S_q}$ of $m$ sparse vectors corresponding to the decomposition of $\B{h}=\hat{\B{\beta}}-\B{\beta}^*$. We note that:
		\begin{align} \label{decomposition}
		\begin{split}
		\Delta(\B{\beta}^*, \B{h}) &=\frac{1}{n} \sum_{i=1}^n f \left( \langle \B{x_i},  \B{\beta}^* + \B{h}  \rangle ;  y_i \right)  - \frac{1}{n} \sum_{i=1}^n  f \left( \langle \B{x_i},  \B{\beta}^* \rangle ;  y_i \right) \\
		&=\frac{1}{n} \sum_{i=1}^n f \left( \langle \B{x_i},  \B{\beta}^* + \sum_{j=1}^{q} \B{h}_{S_j}   \rangle ;  y_i \right)  - \frac{1}{n} \sum_{i=1}^n  f \left( \langle \B{x_i},  \B{\beta}^* \rangle ;  y_i \right)\\
		&=  \sum_{\ell=1}^{q} \left\{ \frac{1}{n} \sum_{i=1}^n f \left( \langle \B{x_i},  \B{\beta}^* + \sum_{j=1}^{\ell} \B{h}_{S_j}   \rangle ;  y_i \right)  - \frac{1}{n} \sum_{i=1}^n  f \left( \langle \B{x_i},  \B{\beta}^*  + \sum_{j=1}^{\ell-1} \B{h}_{S_j}  \rangle ;  y_i \right)  \right\}\\
		&= \sum_{\ell=1}^{q} \Delta \left( \B{\beta}^*  + \sum_{j=1}^{\ell-1} \B{h}_{S_j} , \ \B{h}_{S_\ell} \right)\\
		&= \sum_{\ell=1}^{q} \Delta \left( \B{w}_{\ell-1} , \ \B{h}_{S_\ell} \right)
		\end{split}
		\end{align}
		where we have defined  $\B{w}_{\ell} = \B{\beta}^*  + \sum_{j=1}^{\ell} \B{h}_{S_j}, \forall \ell$ and $\B{h}_{S_0}=\B{0}$.
		
		\smallskip
		\noindent
		We now consider the trivial partition of $\left\{1, \ldots, p\right\}$ for which we apply Theorem \ref{hoeffding-sup}. We fix $k=k^*$, $m=1$, and consider the partition $S_1=\left\{1\right\}$, $S_2=\left\{2\right\}$, 
		$S_q=\left\{p\right\}$ with
		$q = p$. It holds $m\le k$ and Assumption \ref{asu5} guarantees $\log(7Rp) \le k^*$ then $m \log(7Rq) \le k$. Consequently, since $\| \B{h}_{S_\ell} \|_1 \ge 3R, \forall \ell$,  Theorem \ref{hoeffding-sup} guarantees that for all regularization schemes, it holds with probability at least $1 - \frac{\delta}{2}$:
		$$\left| \Delta \left( \B{w}_{\ell-1} ,  \B{h}_{S_\ell} \right) - \mathbb{E} \left( \B{w}_{\ell-1} ,  \B{h}_{S_\ell}  \right) \right|
		\ge \tau^* \| \B{h}_{S_\ell}  \|_1 - 2 \phi , ~ \forall \ell.$$
		
		\smallskip
		\noindent
		As a result, following Equation \eqref{decomposition}, we have:
		\begin{align} \label{lower-bound-expectation-1}
		\begin{split}
		\Delta(\B{\beta}^* , \B{h})
		\ge   \sum_{\ell=1}^{q}  \left\{ \mathbb{E} \left( \B{w}_{\ell-1}  ,  \B{h}_{S_\ell}  \right) -  \tau^* \| \B{h}_{S_\ell}  \|_1 - 2 \phi \right\}
		&= \mathbb{E} \left( \sum_{\ell=1}^{q} \Delta \left( \B{w}_{\ell-1} , \ \B{h}_{S_\ell} \right) \right) - \sum_{\ell=1}^q \tau^* \| \B{h}_{S_\ell}  \|_1 - 2 q \phi\\
		&= \mathbb{E} \left( \Delta(\B{\beta}^* , \B{h}) \right) -  \tau^* \| \B{h}  \|_1 - 2 q \phi.
		\end{split}
		\end{align}
		In addition, we have:
		$$
		\mathbb{E} \left( \Delta(\B{\beta}^* , \B{h}) \right)  = 
		\frac{1}{n} \sum_{i=1}^n \mathbb{E} \left\{  f \left( \langle \B{x_i},  \B{\beta}^* + \B{h}  \rangle ;  y_i \right)  -  f \left( \langle \B{x_i},  \B{\beta}^*  \rangle ;  y_i \right)  \right\} =\mathcal{L}(\B{\beta}^* + \B{h}) - \mathcal{L}(\B{\beta}^*).
		$$
		Consequently, we conclude that with probability at least $1 - \frac{\delta}{2}$:
		\begin{equation} \label{lower-bound-expectation}
		\Delta(\B{\beta}^* , \B{h})\ge \mathcal{L}(\B{\beta}^* + \B{h}) - \mathcal{L}(\B{\beta}^*) -   \tau^* \| \B{h}  \|_1 - 2 q \phi.
		\end{equation}
		
		\paragraph{Step 2:} 
		We now lower-bound the right-hand side of Equation  \eqref{lower-bound-expectation}. Since $\mathcal{L}$ is twice differentiable, a Taylor development around $\B{\beta}^*$ gives:
		$$\mathcal{L}(\B{\beta}^* + \B{h}) - \mathcal{L}(\B{\beta}^*) 
		= \nabla \mathcal{L}(\B{\beta}^*)^T\B{h}  + \frac{1}{2} \B{h}^T \nabla^2 \mathcal{L}(\B{\beta}^*)^T\B{h} + o \left( \| \B{h}\|_2^2 \right).$$
		The optimality of $\B{\beta}^*$ implies $\nabla \mathcal{L}(\B{\beta}^*)=0$. In addition, by using Theorem \ref{cone-condition}, we obtain with probability at least $1- \frac{\delta}{2}$ that $\B{h} \in \Lambda \left(S_0, \gamma_1^*, \gamma_2^* \right)$ for L1 regularization,  $\B{h} \in \Gamma \left(k^*, \omega^*\right)$ for Slope regularization and $\B{h} \in \Omega \left(\mathcal{J}_0, \epsilon_1^*, \epsilon_2^*\right)$ for Group L1-L2 regularization. 
		Consequently, for each regularization, we can use the restricted eigenvalue conditions defined in Assumption \ref{asu4}. However we do not want to keep the term $o \left( \| \B{h} \|_2^2 \right)$ as it can hide non trivial dependencies.

		\smallskip
		\noindent
		We use the shorthand $\kappa^*$ and $r^*$ for the restricted eigenvalue constant and maximum radius introduced in the growth conditions in Assumption \ref{asu5}:  $\kappa^* = \kappa \left(k^*, \gamma_1^*, \gamma_2^* \right)$ and $r^* = r \left(k^*, \gamma_1^*, \gamma_2^* \right)$ for L1 regularization,  $\kappa^* = \kappa \left(k^*, \omega^*\right)$, $r^* = r \left(k^*, \omega^*\right)$ for Slope regularization, $\kappa^* = \kappa \left(s^*, \epsilon_1^*, \epsilon_2^*\right)$ and $r^* = r \left(s^*, \epsilon_1^*, \epsilon_2^*\right)$ for Group L1-L2 regularization. We consider the two mutually exclusive existing cases separately.
		
		\medskip
		\noindent
		\textbf{Case 1:} 
		If $\| \B{h} \|_2 \le r^*$, then using Theorem \ref{cone-condition} and Assumption  \ref{asu4}, it holds with probability at least  $1- \frac{\delta}{2}$:
		\begin{equation}\label{case1-LB}
		\mathcal{L}(\B{\beta}^* + \B{h}) - \mathcal{L}(\B{\beta}^*) \ge \frac{1}{4}  \kappa^* \|\B{h} \|_2^2.
		\end{equation}
		
		\medskip
		\noindent
		\textbf{Case 2:} If now $\| \B{h} \|_2 \ge r^*$, then using the convexity of $\mathcal{L}$ thus of $t \to \mathcal{L}\left( \B{\beta}^* + t \B{h} \right)$, we similarly obtain with the same probability:
		\begin{align} \label{trick}
		\begin{split}
		\mathcal{L}(\B{\beta}^* + \B{h}) - \mathcal{L}(\B{\beta}^*) 
		& \ge  \frac{\|\B{h} \|_2}{r^* } \left\{ \mathcal{L} \left(\B{\beta}^* + \frac{r^* }{\|\B{h} \|_2} \B{h} \right)   - \mathcal{L}   (\B{\beta}^* )  \right\} \textnormal{by convexity} \\
		& \ge  \frac{\|\B{h} \|_2}{r^* } \inf \limits_{ \substack{\B{z}: \  \B{z} \in \Lambda(S_0, \gamma_1^*, \gamma_2^*) \\ \| \B{z} \|_2 = r^* }  }  \left\{ \mathcal{L}(\B{\beta}^* + \B{z} )   - \mathcal{L}(\B{\beta}^*)  \right\} \\
		&\ge    \frac{\|\B{h} \|_2}{r^* } \ \frac{1}{4} \kappa^*  (r^*)^2 = \frac{1}{4} \kappa^* r^*  \|\B{h} \|_2,
		\end{split}
		\end{align}
		where the cone used is for L1 regularization. The same equation holds for Slope and Group L1-L2 regularizations by respectively replacing $\Lambda \left(S_0, \gamma_1^*, \gamma_2^* \right)$ with $\Gamma \left(k^*, \omega^*\right)$ and $\Omega \left(\mathcal{J}_0, \epsilon_1^*, \epsilon_2^*\right)$ 
		
		\medskip
		\medskip
		\noindent
		Combining Equations \eqref{lower-bound-expectation}, \eqref{case1-LB} and \eqref{trick}, we conclude that with probability at least $1-\frac{\delta}{2}$, the following restricted strong convexity condition holds:
		\begin{equation}\label{proof-rsc}
		\Delta(\B{h}) \ge \frac{1}{4}  \kappa^* \|\B{h} \|_2^2 \wedge   \frac{1}{4} \kappa^* r^*  \|\B{h} \|_2 -   \tau^*  \| \B{h}\|_1 - 2 p \phi.
		\end{equation}
		We now prove Theorem \ref{hoeffding-sup}.
	\end{Proof}

	\subsection {Proof of Theorem \ref{hoeffding-sup} }  \label{sec: hoeffding-sup}
	To prove Theorem \ref{hoeffding-sup}, we first use Hoeffding's inequality to obtain an upper bound of the inner supremum for any sequence of $m$ sparse vectors. The result is extended to the outer supremum with an $\epsilon$-net argument. 
	\begin{Proof}
		Let $k, m, q \in \left\{1, \ldots, p\right\}$ be such that $m \le k$, $ n \le q$, $m \log(7Rq) \le k$ and  $S_1, \ldots S_q$ be a partition of $\left\{1, \ldots, p\right\}$ of size $q$ such that $|S_{\ell}| \le m, \; \forall \ell \le q$. We divide the proof in 3 steps. We first upper-bound the inner supremum for any sequence of $m$ sparse vectors $\B{z}_{S_1}, \ldots, \B{z}_{S_q}$. We then extend this bound for the supremum over a compact set of sequences through an $\epsilon$-net argument.
		
		\paragraph{Step 1: } Let us fix a sequence  $\B{z}_{S_1}, \ldots, \B{z}_{S_q} \in \mathbb{R}^{p}: \ \Supp(\B{z}_{S_{\ell} })  \subset S_{\ell}, \forall \ell$ and $\| \B{z}_{S_{\ell} }  \|_1 \le 3R,  \forall \ell$. 
		\smallskip
		\noindent
		In particular, $\| \B{z}_{S_{\ell} } \|_0 \le m \le k, \forall \ell$. In the rest of the proof, we define  $\B{z}_{S_0} = \B{0}$ and
		\begin{equation}  \label{w_l_def}
		\B{w}_{\ell} = \B{\beta}^*  + \sum_{j=1}^{\ell} \B{z}_{S_j}, \forall \ell.
		\end{equation}
		In addition, we  introduce $Z_{i \ell}, \ \forall i,  \ell$ as follows
		$$
		Z_{i \ell} = f \left( \langle \B{x}_i, \B{w}_{\ell}   \rangle ;  y_i \right) -  f \left( \langle \B{x_i},  \B{w}_{\ell-1}  \rangle ;  y_i \right) 
		= f \left( \langle \B{x_i},  \B{w}_{\ell-1} + \B{z}_{S_\ell}  \rangle ;  y_i \right) - f \left( \langle \B{x}_i, \B{w}_{\ell-1}   \rangle ;  y_i \right).  
		$$
		In particular, let us note that:
		\begin{align}
		\begin{split}
		\Delta \left( \B{w}_{\ell-1} , \ \B{z}_{S_\ell} \right) 
		&= \frac{1}{n} \sum_{i=1}^n   f \left( \langle \B{x}_i,  \B{w}_{\ell-1} + \B{z}_{S_{\ell}} \rangle ;  y_i \right)  - \frac{1}{n} \sum_{i=1}^n   f \left( \langle \B{x}_i,   \B{w}_{\ell-1}\rangle ;  y_i \right)  \\
		&= \frac{1}{n} \sum_{i=1}^n \left\{  f \left( \langle \B{x}_i,  \B{w}_{\ell-1} + \B{z}_{S_{\ell}} \rangle ;  y_i \right)  -  f \left( \langle \B{x}_i,   \B{w}_{\ell-1}\rangle ;  y_i \right) \right\}  \\
		&= \frac{1}{n} \sum_{i=1}^n  Z_{i\ell}.
		\end{split}
		\end{align}
		Assumption \ref{asu1} guarantees that $f(.,y)$ is $L$-Lipschitz $\forall y$ then:
		$$|Z_{i\ell} | \le L \left| \langle \B{x}_i, \B{z}_{S_{\ell}}  \rangle \right|, \forall i, \ell.$$
		Hence, with Hoeffding's lemma, the centered bounded random variable $Z_{i\ell} - \mathbb{E}(Z_{i\ell})$ is sub-Gaussian with variance $L^2 \left| \langle \B{x}_i, \B{z}_{S_{\ell}}  \rangle \right|^2$.
		Thus, the centered random variable
		$\Delta \left( \B{w}_{\ell-1} , \ \B{z}_{S_\ell} \right) - \mathbb{E}\left( \Delta\left( \B{w}_{\ell-1}  , \ \B{z}_{S_\ell} \right)  \right) $ is sub-Gaussian with variance $\frac{L^2}{n} \| \B{X}  \B{z}_{S_{\ell}} \|_2^2$. 
		Using  Assumption \ref{asu4}$.1(k)$, it is then sub-Gaussian with variance 
		$\frac{L^2 \mu(k)^2}{nk} \| \B{z}_{S_{\ell}} \|_1^2$. It then holds, $\forall t>0$,
		\begin{align} \label{upper-bound-2k-sparse}
		\begin{split}
		\mathbb{P}\left( \left| \Delta\left(  \B{w}_{\ell-1} , \ \B{z}_{S_\ell} \right)  - \mathbb{E}\left( \Delta\left( \B{w}_{\ell-1}  , \ \B{z}_{S_\ell} \right)  \right) \right|  > t  \| \B{z}_{S_\ell} \|_1  \right) 
		&\le 2\exp \left(- \frac{k n t^2 }{ 2 L^2 \mu(k)^2 }  \right), \forall \ell.
		\end{split}
		\end{align}
		Equation \eqref{upper-bound-2k-sparse} holds for all values of $\ell$. Thus, an union bound immediately gives:
		\begin{equation} \label{union-upper-bound-2k-sparse}
		\mathbb{P}\left( \sup \limits_{\ell=1,\ldots,q} \left\{   \left| \Delta\left( \B{w}_{\ell-1} , \ \B{z}_{S_\ell} \right)  - \mathbb{E}\left( \Delta\left(  \B{w}_{\ell-1}  , \ \B{z}_{S_\ell} \right)  \right) \right|  - t   \| \B{z}_{S_\ell} \|_1 \right\} > 0 \right) \le 2 q \exp \left(- \frac{k n t^2 }{ 2 L^2 \mu(k)^2 }  \right).
		\end{equation}

		\paragraph{Step 2: } We  extend the result to any sequence of vectors $\B{z}_{S_1}, \ldots, \B{z}_{S_q} \in \mathbb{R}^{p}: \ \Supp(\B{z}_{S_{\ell}})  \subset S_{\ell}, \forall \ell$ and $\| \B{z}_{S_{\ell}}  \|_1 \le 3R,  \forall \ell$ with an $\epsilon$-net argument.
		
		\smallskip
		\noindent
		We recall that an $\epsilon$-net of a set $\mathcal{I}$ is a subset $\mathcal{N}$ of $\mathcal{I}$ such that each element of $I$ is at a distance at most  $\epsilon$ of $\mathcal{N}$. We know from Lemma 1.18 from \citet{lecture-notes}, that for any $\epsilon \in (0,1)$, the ball $\left\{ \B{z} \in \mathbb{R}^d: \ \|  \B{z} \|_1 \le R  \right\}$ has an $\epsilon$-net of cardinality $| \mathcal{N} | \le \left(\frac{2R+1}{\epsilon} \right)^d$ -- the $\epsilon$-net is defined in term of L1 norm. In addition, we can create this set such that it contains $\B{0}$.
		
		\smallskip
		\noindent
		Consequently, we use Equation \eqref{union-upper-bound-2k-sparse} on a product of $\epsilon$-nets $\mathcal{N}_{m,R} = \prod \limits_{\ell=1}^q \mathcal{N}_{m,R}^{\ell}$.  Each $\mathcal{N}_{m,R}^{\ell}$ is an $\epsilon$-net of the bounded sets of $m$ sparse vectors $ \mathcal{I}_{m,R}^{\ell} = \left\{ \B{z}_{S_{\ell}} \in \mathbb{R}^{p}: \ \Supp(\B{z}_{S_{\ell}})  \subset S_{\ell} \ ; \ \| \B{z}_{S_{\ell}}  \|_1 \le 3R\right\}$ which contains $\B{0}_{S_{\ell}}$. We note $\mathcal{I}_{m,R} = \prod \limits_{\ell=1}^q \mathcal{I}_{m,R}^{\ell}$. Since $|S_{\ell}| \le m, \forall \ell \le q$, it then holds:
		\begin{align} \label{sup-epsilon-net}
		\begin{split}
		&\mathbb{P}\left( \sup \limits_{ \left( \B{z}_{S_1}, \ldots, \B{z}_{S_q}\right) \in \mathcal{N}_{m,R}  } \ \ \left\{ \sup \limits_{\ell=1,\ldots,q} \left\{   \left| \Delta\left( \B{w}_{\ell-1} , \ \B{z}_{S_\ell} \right)  - \mathbb{E}\left( \Delta\left(  \B{w}_{\ell-1}  , \ \B{z}_{S_\ell} \right)  \right) \right|  - t  \| \B{z}_{S_\ell} \|_1 \right\} > 0 \right\} \right)\\
		&\le 2 q \left(\frac{6R+1}{\epsilon} \right)^{m} q  \exp \left(- \frac{k n t^2 }{ 2 L^2 \mu(k)^2 } \right) \\
		&= 2 q^2 \left(\frac{6R+1}{\epsilon} \right)^{m} \exp \left(- \frac{k n t^2 }{ 2 L^2 \mu(k)^2 } \right).
		\end{split}
		\end{align}
		
		\paragraph{Step 3: } We now extend Equation \eqref{sup-epsilon-net} to control any vector in $\mathcal{I}_{m,R}$.  For $\B{z}_{S_1}, \ldots, \B{z}_{S_q} \in \mathcal{I}_{m,R}$, there exists $\tilde{\B{z}}_{S_1}, \ldots, \tilde{\B{z}}_{S_q}\in \mathcal{N}_{m,R}$ such that $\| \B{z}_{S_\ell} - \tilde{\B{z}}_{S_\ell} \|_1 \le \epsilon, \forall \ell.$ Similar to Equation \eqref{w_l_def}, we define:
		$$ \tilde{\B{w}}_{\ell} = \B{\beta}^*  + \sum_{j=1}^{\ell} \tilde{\B{z}}_{S_j}, \forall \ell.$$ 
		For a given $t$, let us define 
		$$f_t \left( \B{w}_{\ell-1} , \ \B{z}_{S_{\ell}} \right) = \left| \Delta \left( \B{w}_{\ell-1} , \ \B{z}_{S_{\ell}} \right) - \mathbb{E} \left( \B{w}_{\ell-1} , \ \B{z}_{S_{\ell}} \right) \right| - t  \| \B{z}_{S_{\ell}} \|_1, \forall \ell.$$
		\smallskip
		\noindent
		We fix $\ell_0(t)$ such that $\ell_0 \in \argmax \limits_{\ell=1,\ldots,q}  \left\{   f_{2t} \left( \B{w}_{\ell-1} , \ \B{z}_{S_{\ell}} \right)  \right\}$. The choice of $2t$ will be justified later. We fix $t$ and will just note $\ell_0=\ell_0(t)$ when no confusion can be made. 
		
		\smallskip
		\noindent
		With Assumption \ref{asu1} we obtain:
		\begin{align} \label{help-step3}
		\begin{split}
		&\left| \Delta \left( \B{w}_{\ell_0-1} , \ \B{z}_{S_{\ell_0}} \right)
		- \Delta\left( \tilde{\B{w}}_{\ell_0-1} , \ \tilde{\B{z}}_{S_{\ell_0}} \right) \right| \\
		&= \frac{1}{n} \left| \sum_{i=1}^n f \left( \langle \B{x}_i,  \B{w}_{\ell_0}   \rangle ;  y_i \right) -\sum_{i=1}^n f \left( \langle \B{x}_i,  \tilde{\B{w}}_{\ell_0}  \rangle ;  y_i \right) +  \sum_{i=1}^n  f \left( \langle \B{x}_i,  \tilde{\B{w}}_{\ell_0-1}  \rangle ;  y_i \right)    - \sum_{i=1}^n  f \left( \langle \B{x}_i,  \B{w}_{\ell_0-1}  \rangle ;  y_i \right) \right| \\
		&\le \frac{1}{n} \sum_{i=1}^n L \left| \langle \B{x}_i, \B{w}_{\ell_0} - \tilde{\B{w}}_{\ell_0} \rangle  \right| + \frac{1}{n} \sum_{i=1}^n L \left| \langle \B{x}_i, \B{w}_{\ell_0-1} - \tilde{\B{w}}_{\ell_0-1} \rangle  \right|  \\
		&= \frac{1}{n} \sum_{i=1}^n L \left|  \sum_{\ell=1}^{\ell_0}  \langle \B{x}_i, \B{z}_{S_\ell} - \tilde{\B{z}}_{S_\ell} \rangle  \right| + \frac{1}{n} \sum_{i=1}^n L \left| \sum_{\ell=1}^{\ell_0-1}  \langle \B{x}_i,\B{z}_{S_\ell} - \tilde{\B{z}}_{S_\ell} \rangle  \right|  \\
		&\le \frac{2}{n} \sum_{i=1}^n \sum_{\ell=1}^{q} L \left| \langle \B{x}_i, \B{z}_{S_\ell} - \tilde{\B{z}}_{S_\ell} \rangle  \right| \\
		&= 2 \sum_{\ell=1}^{q}   \frac{L}{n}  \left\| \B{X} \right( \B{z}_{S_\ell} - \tilde{\B{z}}_{S_\ell} \left) \right\|_1 \\
		&\le 2 \sum_{\ell=1}^{q}   \frac{L}{\sqrt{n}}  \left\| \B{X} \right( \B{z}_{S_\ell} - \tilde{\B{z}}_{S_\ell} \left) \right\|_2 \text{with Cauchy-Schwartz inequality} \\
		&\le  2 \sum_{\ell=1}^{q}  \frac{L}{\sqrt{k} }  \mu(k) \ \left\| \B{z}_{S_\ell} - \tilde{\B{z}}_{S_\ell} \right\|_1 \text{with Assumption \ref{asu4}$.1(k)$}\\
		&\le  \frac{2q}{\sqrt{k}}  L \mu(k) \epsilon =  \frac{2}{q^2}  L \mu(k) \le  \frac{2}{n q}  L \mu(k) :=\phi.
		\end{split}
		\end{align}
		where we have fixed $\epsilon=\frac{\sqrt{k}}{{q^3}}$ and used that $n \le q$ and $\phi = \frac{2}{nq}  L \mu(k)$.
		It then holds
		\begin{align*} 
		\begin{split}
		f_t \left( \tilde{\B{w}}_{\ell_0-1} , \ \tilde{\B{z}}_{S_{\ell_0}} \right)  
		&\ge f_t \left( \B{w}_{\ell_0-1} , \ \B{z}_{S_{\ell_0}} \right) -  \left| \Delta \left( \B{w}_{\ell_0-1} , \ \B{z}_{S_{\ell_0}} \right) -  \Delta\left( \tilde{\B{w}}_{\ell_0-1} , \ \tilde{\B{z}}_{S_{\ell_0}} \right)  \right|  \\
		&\ \ \ - \left| \mathbb{E}\left(  \Delta \left( \B{w}_{\ell_0-1} , \ \B{z}_{S_{\ell_0}}\right) -  \Delta\left( \tilde{\B{w}}_{\ell_0-1} , \ \tilde{\B{z}}_{S_{\ell_0}} \right) \right)\right|   - t  \| \B{z}_{S_{\ell_0}} - \tilde{\B{z}}_{S_{\ell_0}}  \|_1  \\
		&\ge f_t \left( \B{w}_{\ell_0-1} , \ \B{z}_{S_{\ell_0}} \right)  -  2 \phi - t \epsilon.
		\end{split}
		\end{align*}
		\medskip
		\noindent
		\textbf{Case 1:} Let us assume that $\| \B{z}_{S_{\ell_0}} \|_1 \ge \epsilon$, then we have:
		\begin{equation}\label{step3-eps}
		f_t \left( \tilde{\B{w}}_{\ell_0-1} , \ \tilde{\B{z}}_{S_{\ell_0}} \right)  \ge f_t \left( \B{w}_{\ell_0-1} , \ \B{z}_{S_{\ell_0}} \right)  -  t \| \B{z}_{S_{\ell_0}} \|_1 - 2\phi \ge f_{2t} \left( \B{w}_{\ell_0-1} , \ \B{z}_{S_{\ell_0}} \right) - 2 \phi. 
		\end{equation}
		\medskip
		\noindent
		\textbf{Case 2:} We now assume $\| \B{z}_{S_{\ell_0}} \|_1 \le \epsilon$. Since $\B{0}_{S_{\ell_0} } \in  \mathcal{N}_{k,R}$ we derive, similar to Equation  \eqref{help-step3}:
		$$\left| \Delta \left( \B{w}_{\ell_0-1} , \ \B{z}_{S_{\ell_0}} \right)
		- \Delta\left( \B{w}_{\ell_0-1} , \ \B{0}_{S_{\ell_0}} \right) \right| 
		\le \frac{L \mu(k) }{\sqrt{k} } \left\| \B{z}_{S_{\ell_0} } \right\|_1 \le \frac{L \mu(k) }{\sqrt{k} } \epsilon  \le \frac{\phi}{q},$$
		which then implies that:
		$$f_{2t} \left( \B{w}_{\ell_0-1} , \ \B{z}_{S_{\ell_0}} \right) - 2 \frac{\phi}{q} 
		\le f_{2t} \left( \B{w}_{\ell_0-1} , \ \B{0}_{S_{\ell_0}}  \right),$$
		\smallskip
		\noindent
		In this case, we can define a new $\tilde{\ell}_0$ for the sequence $\B{z}_{S_1}, \ldots, \B{z}_{S_{\ell_0-1}}, \B{0}_{S_{\ell_0}}, \B{z}_{S_{\ell_0+1}}, \ldots, \B{z}_{S_q}$.  After at most $q$ iterations, by using the result in Equation \eqref{step3-eps} and the definition of $\ell_0$, we finally get that  $f_{2t} \left( \B{w}_{\ell_0-1} , \ \B{z}_{S_{\ell_0}} \right) - 2 \phi \le f_t \left( \tilde{\B{w}}_{\ell_0-1} , \ \tilde{\B{z}}_{S_{\ell_0}} \right)$ for some $\tilde{\B{z}}_{S_1}, \ldots, \tilde{\B{z}}_{S_q}\in \mathcal{N}_{m,R}$.
		\\
		\newline
		By combining cases 1 and 2, we obtain: $\forall t \ge \phi, \ \forall \B{z}_{S_1}, \ldots, \B{z}_{S_q} \in \mathcal{I}_{m,R}, \ \exists \tilde{\B{z}}_{S_1}, \ldots, \tilde{\B{z}}_{S_q}\in \mathcal{N}_{m,R}$:
		$$\sup \limits_{\ell=1,\ldots,q}  f_{2t} \left( \B{w}_{\ell-1} , \ \B{z}_{S_{\ell}} \right) -2 \phi = f_{2t} \left( \B{w}_{\ell_0-1} , \ \B{z}_{S_{\ell_0}} \right)  - 2 \phi \le f_{t} \left( \tilde{\B{w}}_{\ell_0-1} , \ \tilde{\B{z}}_{S_{\ell_0}} \right) \le \sup \limits_{\ell=1,\ldots,q} f_{t} \left( \tilde{\B{w}}_{\ell-1} , \ \tilde{\B{z}}_{S_{\ell}} \right).$$
		This last relation is equivalent to saying that $\forall t$:
		\begin{equation}\label{supremum-domination}
		\sup \limits_{ \B{z}_{S_1}, \ldots, \B{z}_{S_q} \in \mathcal{I}_{m,R} } \left\{    \sup \limits_{\ell=1,\ldots,q}  f_t \left( \B{w}_{\ell-1} , \ \B{z}_{S_{\ell}} \right)  \right\}  - 2 \phi
		\le 
		\sup \limits_{ \B{z}_{S_1}, \ldots, \B{z}_{S_q} \in \mathcal{N}_{m,R} } \left\{    \sup \limits_{\ell=1,\ldots,q}  f_{t/2} \left( \tilde{\B{w}}_{\ell-1} , \ \tilde{\B{z}}_{S_{\ell}}, \right) \right\}.
		\end{equation}
		As a consequence, we have $\forall t$:
		\begin{align}  \label{sup-compact-set}.
		\begin{split}
		&\mathbb{P}\left( \sup \limits_{ \B{z}_{S_1}, \ldots, \B{z}_{S_q} \in \mathcal{I}_{m,R} } \ \ \sup \limits_{\ell=1,\ldots,q} \left\{   \left| \Delta\left( \B{w}_{\ell-1} , \ \B{z}_{S_\ell} \right)  - \mathbb{E}\left( \Delta\left(  \B{w}_{\ell-1}  , \ \B{z}_{S_\ell} \right)  \right) \right|  - t  \| \B{z}_{S_\ell} \|_1 - 2 \phi \right\} > 0  \right)\\
		&\le \mathbb{P}\left( \sup \limits_{ \B{z}_{S_1}, \ldots, \B{z}_{S_q} \in \mathcal{N}_{m,R} } \ \ \sup \limits_{\ell=1,\ldots,q} \left\{   \left| \Delta\left( \B{w}_{\ell-1} , \ \B{z}_{S_\ell} \right)  - \mathbb{E}\left( \Delta\left(  \B{w}_{\ell-1}  , \ \B{z}_{S_\ell} \right)  \right) \right|  - \frac{t}{2}  \| \B{z}_{S_\ell} \|_1 \right\} > 0  \right)\\
		&\le 2  q^2 \left(\frac{6R+1}{\epsilon} \right)^{m} \exp \left(- \frac{ k n (t/2)^2 }{ 2 L^2 \mu(k)^2 } \right)\\
		& \le \left( 2q \right)^2 (7R)^{m} q^{3m}  \exp \left(- \frac{k n t^2 }{16 L^2 \mu(k)^2 } \right).
		\end{split}
		\end{align}
		We want this last term to be lower than $\frac{\delta}{2}$. 
		\newline
		We then want to select $t$ such that $t^2 \ge \frac{16 L^2 \mu(k)^2 }{kn}\left[3m \log(7Rq) + 2 \log\left( 2q \right) + \log\left( \frac{2}{\delta}\right) \right]$ holds. To this end, since $m \log(7Rq) \le k $, we define: 
		$$\tau=\tau(k, m, q) = 9 L  \mu(k) \sqrt{\frac{1}{n} + \frac{\log\left( 2/\delta\right) }{nk}  }.$$
		We conclude that with probability at least $1-\frac{\delta}{2}$:
		$$ \sup \limits_{ \B{z}_{S_1}, \ldots, \B{z}_{S_q} \in \mathcal{I}_{m,R} } \left\{\sup \limits_{\ell=1,\ldots,q} \left\{   \left| \Delta\left( \B{w}_{\ell-1} , \ \B{z}_{S_\ell} \right)  - \mathbb{E}\left( \Delta\left(  \B{w}_{\ell-1}  , \ \B{z}_{S_\ell} \right)  \right) \right|  - \tau  \| \B{z}_{S_\ell} \|_1 - 2 \phi \right\}  \right\}  \le 0.$$
	\end{Proof}

	\section {Proof of Theorem \ref{main-results} } \label{sec: appendix_main-results}

	\begin{Proof} We now prove our main Theorem  \ref{main-results} for the three regularizations considered.  
		
		\paragraph{L1 regularization: } For L1 regularization, we have proved in Theorem \ref{cone-condition} that $\B{h}= \hat{\B{\beta}} _1- \B{\beta}^* \in \Lambda \left( S_0, \ \gamma_1^*,  \ \gamma_2^* \right)$  where $S_0$ has been defined as the subset of the $k^*$ highest elements of $\B{h}$. We have defined $\kappa^* = \kappa \left(k^*, \gamma_1^*, \gamma_2^* \right)$,  $r^* = r \left(k^*, \gamma_1^*, \gamma_2^* \right)$ and $\tau^* = \tau(k^*) $.
		
		\smallskip
		\noindent
		Since $\mu(k^*) \le \alpha M$, then $9 L \mu(k^*) \sqrt{\frac{1 }{n} + \frac{\log\left( 2/\delta\right) }{nk^*} } \le 12 \alpha L M \sqrt{\frac{\log\left( 2e \right) \log(2/ \delta) }{n} }$, hence we have $\tau^* \le \eta \lambda^{(p)}_{p} \le \eta \lambda_{k^*}^{(p)} = \lambda$---where $\lambda^{(r)}_{j} = \sqrt{\log(2re / j)}$. 
		
		\smallskip
		\noindent
		By pairing Equation \eqref{inf-equation} with the restricted strong convexity derived in Theorem \ref{restricted-strong-convexity}, it holds with probability at least $1-\delta$:
		\begin{equation*}
		\begin{split}
		\frac{1}{4}  \kappa^* \left\{ \|\B{h}\|_2^2 \wedge r^* \|\B{h}\|_2 \right\} 
		&\le \tau^* \|\B{h}\|_1 + \lambda \| \B{h}_{S^*} \|_1 -\lambda \| \B{h}_{(S^*)^c}  \|_1 + 2 p\phi\\
		&= \tau^* \| \B{h}_{S_0} \|_1 + \tau^* \| \B{h}_{(S_0)^c} \|_1 + \lambda \| \B{h}_{S_0} \|_1 -\lambda \| \B{h}_{(S^*)^c}  \|_1  + 2 p\phi\\
		&\le \tau^* \| \B{h}_{S_0} \|_1 + \lambda \| \B{h}_{S_0} \|_1 \text{ since } \tau^* \le \lambda \text { and } \| \B{h}_{(S_0)^c} \|_1 \le  \| \B{h}_{(S^*)^c}  \|_1  + 2 p\phi\\
		&\le \left(\tau^* + \lambda \right) \sqrt{k^*}  \|\B{h}_{S_0} \|_2  + 2 p\phi \text{ from Cauchy-Schwartz inequality} \\
		&\le \left(\tau^* + \lambda \right) \sqrt{k^*}  \|\B{h} \|_2  + 2 p\phi.
		\end{split}
		\end{equation*}
		We have defined $\phi=\frac{2}{pn}  L \mu(k)$. Let us currently assume that $2p\phi \le \left(\tau^* + \lambda \right) \sqrt{k^*}  \|\B{h} \|_2$. It then holds with probability at least $1-\delta$:
		\begin{equation}
		\label{use-tau-lambda}
		\frac{1}{4}  \kappa^* \left\{ \|\B{h}\|_2 \wedge r^*  \right\} \le 2\left(\tau^* + \lambda \right) \sqrt{k^*}
		\end{equation}
		Exploiting Assumption \ref{asu5}$.1(p, k^*, \alpha, \delta)$, and using the definitions of $\lambda$ and $\tau^*$ as in Theorems \ref{cone-condition} and  \ref{hoeffding-sup}, Equation \eqref{use-tau-lambda} leads to: 
		\begin{equation*}
		\begin{split}
		\frac{1}{4}  \kappa^* \|\B{h}\|_2 
		&\le 24 \alpha L M \sqrt{ \frac{ k^* \log(2pe/k^*) }{n}  \log(2/ \delta) }
		+ 18 L  \mu(k^*) \sqrt{\frac{k^* + \log\left( 2/\delta\right)}{n}  }.
		\end{split}
		\end{equation*}
		Hence we obtain with probability at least $1 - \delta$:
		\begin{equation*}
		\begin{split}
		&\|\B{h}\|_2^2 \lesssim 
		\left( \frac{ \alpha L M }{\kappa^*} \right)^2  \frac{ k^* \log\left( p/k^* \right) \log\left( 2/\delta \right) }{n} + \left( \frac{ L \mu(k^*)  }{ \kappa^*} \right)^2   \frac{ k^* + \log\left( 2/ \delta \right) }{n}.
		\end{split}
		\end{equation*}
		If now $\left(\tau^* + \lambda \right) \sqrt{k^*}  \|\B{h} \|_2 \le 2 p\phi$ then $\|\B{h} \|_2 \le \frac{2 L \mu(k)}{n \left(\tau^* + \lambda \right) \sqrt{k^*}}$ which is smaller than the above quantity, and concludes the proof.

		\paragraph{Slope regularization: } For Slope regularization, the cone condition derived in Theorem \ref{cone-condition} gives $\B{h} = \hat{\B{\beta}}_{\mathcal{S}}  - \B{\beta}^*\in \Gamma \left( k^*, \omega^*\right)$.  In addition, we have defined $\kappa^* = \kappa \left(k^*, \omega^*\right)$, $r^* = r \left(k^*, \omega^*\right)$ and $\tau^* = \tau(k^*) $. Similar to the above, we denote $S_0$ the subset of the $k^*$ highest elements of $\B{h}$, and note $\lambda_j =  \lambda^{(p)}_{j}$ where we drop the dependency upon $p$. 
		
		\smallskip
		\noindent
		Pairing Equation \eqref{inf-equation} and the restricted strong convexity derived in Theorem \ref{restricted-strong-convexity}, we obtain with probability at least $1-\delta$:
		\begin{equation}\label{preliminary-slope}
		\begin{split}\frac{1}{4}  \kappa^* \left\{ \|\B{h}\|_2^2 \wedge r^* \|\B{h}\|_2 \right\} 
		&\le \tau^* \|\B{h}\|_1 + \eta \sum_{j=1}^{k^*} \lambda_j | h_{j} | - \eta \sum_{j=k^*+1}^p \lambda_j  | h_{j} | + 2 p\phi\\
		&\le \tau^*  \| \B{h}_{S_0} \|_1 + \tau^* \| \B{h}_{(S_0)^c} \|_1 + \eta \sum_{j=1}^{k^*} \lambda_j | h_{j} |  - \eta \sum_{j=k^*+1}^p \lambda_j  | h_{j} | + 2 p\phi\\
		&\le \tau^*  \| \B{h}_{S_0} \|_1 + \eta \sum_{j=1}^{k^*} \lambda_j | h_{j} | | + 2 p\phi \text{ since } \tau^* \le \eta \lambda_p.
		\end{split}
		\end{equation}
		Hence by using Cauchy-Schwartz inequality we obtain:
		\begin{equation*}
		\begin{split}
		\frac{1}{4}  \kappa^* \left\{ \|\B{h}\|_2^2 \wedge r^* \|\B{h}\|_2 \right\} 
		&\le \tau^* \sqrt{k^*}  \| \B{h}_{S_0}  \|_2 + \eta \sqrt{ k^* \log(2pe /k^*) } \| \B{h}_{S_0}  \|_2 + 2 p\phi\\
		&= \tau^* \sqrt{k^*} \| \B{h}_{S_0}  \|_2  + \lambda \sqrt{k^*} \| \B{h}_{S_0}  \|_2  + 2 p\phi\\
		&\le \left(\tau^*  + \lambda \right) \sqrt{k^*}   \| \B{h}  \|_2  + 2 p\phi,\\
		\end{split}
		\end{equation*}
		which is equivalent to Equation \eqref{use-tau-lambda}. We conclude the proof as above by exploiting Assumption \ref{asu5}$.2(p, k^*,  \alpha, \delta)$.

		\paragraph{Group L1-L2 regularization: } For Group L1-L2 regularization, the cone condition proved in Theorem \ref{cone-condition} gives $\B{h} = \hat{\B{\beta}}_{L1-L2}  - \B{\beta}^*\in \Omega \left( \mathcal{J}_0,  \epsilon_1^*=\frac{\alpha}{\alpha-1},  \epsilon_2^* = \frac{\sqrt{s^*} }{\alpha - 1}  \right)$, where $\mathcal{J}_0$ has been defined as the subset of $s^*$ groups with highest L2 norm. We have defined $\kappa^* = \kappa \left(s^*, \epsilon_1^*, \epsilon_2^*\right)$, $r^* = r \left(s^*, \epsilon_1^*, \epsilon_2^*\right)$ and $\tau^* = \tau(k^*) = 9 L  \mu(k^*) \sqrt{\frac{1}{n} + \frac{\log\left( 2/\delta\right)}{n k^*} }$
		\newline 
		\noindent
		In particular, since we have defined $\lambda_G  = \eta \lambda^{(G)}_{s^*}  + \alpha LM \sqrt{\frac{\gamma m_*}{s^* n} } = 12 \alpha LM \sqrt{  \frac{\log\left( 2Ge/s^* \right)  \log\left( 2/\delta \right)}{n}   } + \alpha LM \sqrt{ \frac{\gamma m_*}{s^* n} }  $ and we have assumed $\mu(k^*) \le \alpha M$, it then holds $\tau^* \le \eta \lambda^{(p)}_{p} \le \lambda_G$. 
		
		\smallskip
		\noindent
		Pairing Equation \eqref{inf-equation} and the restricted strong convexity derived in Theorem \ref{restricted-strong-convexity}, we obtain with probability at least $1-\delta$:
		\begin{equation}
		\begin{split}
		\frac{1}{4}  \kappa^* \left\{ \|\B{h}\|_2^2 \wedge r^* \|\B{h}\|_2 \right\} 
		& \le \tau^* \| \B{h} \|_1 + \lambda_G \sum_{g \in \mathcal{J}^*} \| \B{\beta}^*_g \|_{2} - \lambda_G \sum_{g \in \mathcal{J}^*} \| \hat{\B{\beta}}_g \|_{2} - \lambda_G \sum_{g \notin \mathcal{J}^*} \| \hat{\B{\beta}}_g \|_{2} + 2 p \phi\\
		& \le \tau^* \sum_{g=1}^G \| \B{h}_g \|_1 + \lambda_G \sum_{g \in \mathcal{J}^*} \| \B{h}_g \|_{2} - \lambda_G \sum_{g \notin \mathcal{J}^*} \| \B{h}_g \|_{2} + 2 p \phi\\
		& \le \tau^* \sum_{g \in \mathcal{J}^*} \| \B{h}_g \|_1 + \tau^* \sum_{g \notin \mathcal{J}^*} \| \B{h}_g \|_1 + \lambda_G \sum_{g \in \mathcal{J}^*} \| \B{h}_g \|_{2}  - \lambda_G \sum_{g \notin \mathcal{J}^*} \| \B{h}_g \|_{1} + 2 p \phi
		\end{split}
		\end{equation}
		Since $\tau^* \le \lambda_G$, we then have with probability at least $1 - \delta$:	
		\begin{equation}
		\begin{split}\frac{1}{4}  \kappa^* \left\{ \|\B{h}\|_2^2 \wedge r^* \|\B{h}\|_2 \right\} 
		&\le \tau^* \sum_{g \in \mathcal{J}^*} \| \B{h}_g \|_1 + \lambda_G \sum_{g \in \mathcal{J}^*} \| \B{h}_g \|_{2} + 2 p \phi\\
		&\le \tau^* \sqrt{m^*} \| \B{h}_{\mathcal{T}^*}  \|_2 + \lambda_G \sum_{g \in \mathcal{J}^*} \| \B{h}_g \|_{2} + 2 p \phi,
		\end{split}
		\end{equation}
		where we have used Cauchy-Schwartz inequality and denoted $\mathcal{T}^* = \cup_{g \in \mathcal{J}^*} \mathcal{I}_g$ the subset of size $m^*$ of all indexes across all the $s^*$ groups in $\mathcal{J}^*$. In addition, Cauchy-Schwartz inequality also leads to: $\sum_{g \in \mathcal{J}^*} \| \B{h}_g \|_2 \le \sqrt{s^*} \| \B{h}_{\mathcal{T}_0} \|_{2}$ since $\mathcal{J}^*$ is of size $s^*$. 
		Hence it holds with probability at least $1 - \delta$:	
		\begin{equation}
		\begin{split}
		\frac{1}{4}  \kappa^* \left\{ \|\B{h}\|_2^2 \wedge r^* \|\B{h}\|_2 \right\} 
		& \le \left( \tau^* \sqrt{m^*} + \lambda_G \sqrt{s^*} \right)  \| \B{h}_{\mathcal{T}^*} \|_{2} + 2 p \phi
		\le \left( \tau^* \sqrt{m^*} + \lambda_G \sqrt{s^*} \right)  \| \B{h} \|_{2} + 2 p \phi.
		\end{split}
		\end{equation}
		As before, if $2 p \phi \le \left( \tau^* \sqrt{m^*} + \lambda_G \sqrt{s^*} \right)  \| \B{h} \|_{2}$,
		then, by using Assumption \ref{asu5}$.3(G, s^*, m^*, \alpha, \delta)$, we obtain with probability at least $1 - \delta$:
		\begin{equation*}
		\begin{split}
		&\|\B{h}\|_2^2 \lesssim 
		\left( \frac{ \alpha L M }{\kappa^*} \right)^2 \frac{ s^* \log\left( G/s^* \right) \log\left( 2/\delta \right) + \gamma m^* }{n} + \left( \frac{ L \mu(k^*)  }{ \kappa^*} \right)^2  \frac{ m^* + \log\left( 2/ \delta \right) m^* / k^* }{n}.
		\end{split}
		\end{equation*}.
		If now $\left( \tau^* \sqrt{m^*} + \lambda_G \sqrt{s^*} \right)  \| \B{h} \|_{2} \le 2p \phi$, as before, we obtain a similar result.
	\end{Proof}

	\section {Proof of Corollary \ref{main-corollary} } \label{sec: appendix_main-corollary}
	
	\begin{Proof}
		In order to derive the bounds in expectation, we define the bounded random variable: 
		$$ Z =  \frac{{\kappa^*}^{2} }{L^2} \| \hat{\B{\beta}}  - \B{\beta}^*\|_2^2,$$
		where $\kappa^*$ depends upon  the regularization used. We assume that Assumptions \ref{asu5}$.1(p, k^*, \alpha, \delta)$,  \ref{asu5}$.2(p, k^*,  \alpha, \delta)$ and \ref{asu5}$.3(G, s^*, m^*,  \alpha, \delta)$ are satisfied for a small enough $\delta_0$  in the respective cases of the L1, Slope and Group L1-L2 regularizations. Hence can fix $C_0 > 0$ such that $\forall \delta \in \left(0, 1 \right)$, it holds with probability at least $1-\delta$:
		\begin{align*} 
		\begin{split}
		Z &\le C_0  H_1  \log(2/\delta)  +  C_0 H_2,
		\end{split}
		\end{align*} 
		where $H_1 = \frac{1}{n} \left(\alpha^2 M^2 k^* \log\left( p/k^* \right) + \mu(k^*)^2 \right)$ and $H_2 = \frac{1}{n} \mu(k^*)^2 k^* $ for L1 and Slope regularizations. 
		
		\smallskip
		\noindent
		Similarly $H_1 = \frac{1}{n} (\alpha^2 M^2 s^* \log\left( G/s^* \right) + \mu(k^*)^2 m^*)$ and $H_2 =  \frac{1}{n} (\alpha^2 \gamma m^* + \mu(k^*)^2  m^*)$ for Group L1-L2 regularization.
		
		\smallskip
		\noindent
		Then it holds $\forall t \ge t_0 = \log(2):$
		$$\mathbb{P}\left( Z/C_0 \ge H_1 t + H_2 \right) \le 2e^{-t}.$$
		Let $q_0 = H_1 t_0$, then $\forall q \ge q_0$
		\begin{equation}\label{eq-integration}
		\mathbb{P}\left( Z/C_0 \ge q + H_2\right) \le 2\exp\left( - \frac{q}{H_1}  \right).
		\end{equation}
		Consequently, by integration we have:
		\begin{align} 
		\begin{split}
		\mathbb{E}(Z) &= \displaystyle \int_0^{+ \infty}  C_0\mathbb{P}\left( |Z| /C_0 \ge q \right)dq\\
		&\le \displaystyle \int_{H_2 + q_0}^{+ \infty}  C_0 \mathbb{P}\left( |Z| /C_0 \ge  q\right) dq + C_0 (H_2 + q_0) \\
		&= \displaystyle \int_{q_0}^{+ \infty}  C_0 \mathbb{P}\left( |Z| /C_0 \ge  q + H_2\right) dq + C_0 (H_2 + q_0) \\
		&\le \displaystyle \int_{q_0}^{+ \infty}  2C_0 \exp \left(-\frac{q}{H_1}   \right)dq +  C_0 H_2 + C_0 H_1 t_0 \text{ using Equation \eqref{eq-integration}  }\\
		&= 2C_0 H_1 \exp \left( -\frac{q_0}{H_1}  \right)  +  C_0 H_2 + C_0 H_1 \log(2)  \\
		&\le C_1 \left( H_1 + H_2 \right)
		\end{split}
		\end{align}
		for $C_1 = (2 + \log(2)) C_0$. Hence we derive
		$$\mathbb{E} \| \hat{\B{\beta}} - \B{\beta}^*  \|_2^2  \lesssim   \left( \frac{L}{\kappa^*} \right)^2 \left( H_1 +  H_2\right),$$
		which, for L1 and Slope regularizations, is equivalent to:
		\begin{equation*}
		\mathbb{E} \| \hat{\B{\beta}}_{1, \mathcal{S}}  - \B{\beta}^*\|_2^2 \lesssim \left( \frac{ L}{\kappa^*} \right)^2 \left( \alpha^2 M^2  \frac{k^* \log\left( p /k^* \right)}{n} + \mu(k^*)^2  \frac{k^*}{n} \right),
		\end{equation*}
		and in the case of Group L1-L2 regularization, can be equivalently expressed as:
		\begin{equation*}
		\mathbb{E} \| \hat{\B{\beta}}_{L1-L2}  - \B{\beta}^*\|_2^2 \lesssim \left( \frac{L}{\kappa^*} \right)^2 \left( \alpha^2 M^2  \frac{s^* \log\left( G / s^* \right) + \gamma m^*}{n}  + \mu(k^*)^2  \frac{m^*}{n}  \right).
		\end{equation*}
	\end{Proof}

	\section{Proof of Theorem \ref{cone-condition-regression}} \label{sec: appendix_cone-condition-regression}
	
	We use the minimality of  $\hat{\B{\beta}}$ and Lemma \ref{upper-bound-sup} to derive the cone conditions for Lasso and Group Lasso. Our proofs follow the ones for Theorem \ref{cone-condition}.
	
	\begin{Proof}  We first present the proof for the Lasso estimator before adapting it to Group Lasso.
		\medskip
		
		\paragraph{Proof for Lasso: } $\hat{\B{\beta}}$ denotes herein a Lasso estimator, defined as a solution of the Lasso Problem \eqref{group-problem} hence:
		\begin{align*}
		\begin{split}
		\frac{1}{n}  \|  \B{y} - \B{X} \hat{\B{\beta}} \|_2^2 + \lambda \| \hat{\B{\beta}} \|_1
		&\le \frac{1}{n}  \| \B{y} - \B{X} \B{\beta}^*  \|_2^2 + \lambda \| \B{\beta}^* \|_1 = \frac{1}{n}  \| \B{\epsilon}   \|_2^2 + \lambda \| \B{\beta}^* \|_1.
		\end{split}
		\end{align*}
		We define $\B{h} =  \hat{\B{\beta}} - \B{\beta}^*$. It then holds:
		$$ \frac{1}{n} \|  \B{y} - \B{X} \hat{\B{\beta}} \|_2^2 
		=  \frac{1}{n} \|  \B{X} \B{\beta}^*  - \B{X} \hat{\B{\beta}} \|_2^2 + \frac{2}{n} \B{\epsilon}^T (\B{X} \B{\beta}^*  - \B{X} \hat{\B{\beta}}) +   \frac{1}{n} \| \B{\epsilon}   \|_2^2 =  \frac{1}{n} \|  \B{X} \B{h} \|_2^2 - \frac{2}{n} ( \B{X}^T \B{\epsilon})^T \B{h} +   \frac{1}{n} \| \B{\epsilon}   \|_2^2.$$
		Since $S^*$ is the support of $\B{\beta}^*$ and $S_0 = \left\{1,\ldots,k^*\right\}$ is the set of the $k^*$ largest coefficients of $\B{h}$, it holds:
		\begin{align}\label{rhs-lasso}
		\begin{split}
		\frac{1}{n} \| \B{X} \B{h} \|_2^2  
		&\le \frac{2}{n}  ( \B{X}^T \B{\epsilon})^T \B{h} + \lambda   \| \B{\beta}^*_{S^* } \|_1 - \lambda \| \hat{\B{\beta}}_{S^*} \|_1 - \lambda \| \hat{\B{\beta}}_{(S^*)^c } \|_1 \\
		&\le \frac{2}{n} ( \B{X}^T \B{\epsilon})^T \B{h} + \lambda \| \B{h}_{S^*} \|_{1} - \lambda \| \B{h}_{ (S^*)^c} \|_1\\
		&\le \frac{2}{n} ( \B{X}^T \B{\epsilon})^T \B{h} + \lambda \| \B{h}_{S_0 } \|_{1} - \lambda \| \B{h}_{(S_0)^c} \|_1.
		\end{split}
		\end{align}
		We now upper-bound the quantity $( \B{X}^T \B{\epsilon})^T \B{h} $. To this end, we denote $\B{g} = \B{X}^T \B{\epsilon}$.  The entries of $\B{\epsilon}$ are independent, hence Assumption \ref{asu-col} guarantees that $\forall j, g_j$ is sub-Gaussian with variance $n \sigma^2$. In addition, we introduce a non-increasing rearrangement $(g_{(1)}, \ldots, g_{(p)})$  of $(|g_1|, \ldots, |g_p|)$. We assume without loss of generality that $|h_1| \ge \ldots \ge |h_p|$. Since $\frac{\delta}{2} \le \frac{1}{2}$, Lemma \ref{upper-bound-sup} gives, with probability at least $1-\delta$:
		\begin{align}\label{upper-bound-SG-lasso}
		\begin{split}
		( \B{X}^T \B{\epsilon})^T \B{h}
		&= \sum_{j=1}^p g_j h_j \le \sum_{j=1}^p | g_j | |  h_j | 
		= \sum_{j=1}^p \frac{g_{(j)}}{\sqrt{n}\sigma \lambda_j}  \sqrt{n}\sigma \lambda_j | h_{(j)} | \\
		&\le \sqrt{n}\sigma \sup_{j=1,\ldots,p} \left\{\frac{g_{(j)}}{ \sqrt{n} \sigma \lambda_j }  \right\}   \sum_{j=1}^p \lambda_j  |h_{(j)} |\\
		&\le 12 \sqrt{n} \sigma \sqrt{\log(2/ \delta)} \sum_{j=1}^p \lambda_j  |h_{(j)} |\text{ with Lemma \ref{upper-bound-sup}}\\
		&\le 12 \sqrt{n} \sigma \sqrt{\log(2/ \delta)} \sum_{j=1}^p \lambda_j  |h_{j} | \text{ since } \lambda_1 \ge \ldots \ge \lambda_p \text { and }  | h_1 | \ge \ldots \ge |h_p| \\
		&\le 12 \sqrt{n} \sigma \sqrt{\log(2/ \delta)} \left( \sum_{j=1}^{k^*} \lambda_j  |h_{j} | + \lambda_{k^*} \| \B{h}_{(S_0)^c} \|_1  \right).
		\end{split}
		\end{align}
		As before, Cauchy-Schwartz inequality leads to:
		\begin{align*}
		\sum_{j=1}^{k^*} \lambda_j | h_j |  &\le \sqrt{\sum_{j=1}^{k^*} \lambda_j ^2 } \| \B{h}_{S_0}  \|_2 \le \sqrt{k^*\log(2pe /k^*)}  \| \B{h}_{S_0}  \|_2.
		\end{align*}
		Theorem \ref{cone-condition-regression} defines $\lambda = 24 \alpha  \sigma \sqrt{ \frac{1}{n} \log(2pe/k^*) \log(2 / \delta)}$. Because $\lambda_{k^*} \le  \sqrt{\log(2pe/k^*)}$, we can pair Equations \eqref{rhs-lasso} and \eqref{upper-bound-SG-lasso}  to obtain with probability at least $1 - \delta$:
		\begin{align}\label{fundamental-lasso}
		\begin{split}
		\frac{1}{n} \| \B{X} \B{h} \|_2^2 
		&\le \frac{2}{n} ( \B{X}^T \B{\epsilon})^T \B{h} + \lambda \| \B{h}_{S_0} \|_1 -\lambda \| \B{h}_{(S_0)^c}  \|_1 \\
		&\le 24 \frac{\sigma}{ \sqrt{n} }  \sqrt{\log(2pe/k^*) \log(1/ \delta)} \left( \sqrt{k^*}  \| \B{h}_{S_0}  \|_2  +  \| \B{h}_{(S_0)^c}  \|_1 \right)+ \lambda \| \B{h}_{S_0} \|_1 -\lambda \| \B{h}_{(S_0)^c}  \|_1\\
		&= \frac{\lambda}{\alpha} \left( \sqrt{k^*}  \| \B{h}_{S_0}  \|_2  +  \| \B{h}_{(S_0)^c}  \|_1 \right) + \lambda \| \B{h}_{S_0} \|_1 -\lambda \| \B{h}_{(S_0)^c}  \|_1. 
		\end{split}
		\end{align}
		As a first consequence, Equation \eqref{fundamental-lasso} implies that with probability at least $1 - \delta$:
		$$ \lambda \| \B{h}_{(S_0)^c}  \|_1 -  \frac{\lambda}{\alpha} \| \B{h}_{(S_0)^c}  \|_1  \le \lambda \| \B{h}_{S_0} \|_1 + \frac{\lambda}{\alpha} \sqrt{k^*}\| \B{h}_{S_0}  \|_2,$$
		which is equivalent from saying that with probability at least $1 - \delta$:
		$$ \| \B{h}_{(S_0)^c}  \|_1 \le \frac{\alpha}{\alpha -1}  \| \B{h}_{S_0}  \|_1 + \frac{ \sqrt{k^*}}{\alpha -1}  \| \B{h}_{S_0}  \|_2.$$
		We conclude that $\B{h} \in \Lambda \left(S_0, \ \frac{\alpha}{\alpha -1}, \  \frac{\sqrt{k^*}}{\alpha -1} \right)$ with probability at least $1-\delta$.
		
		\medskip
		
		\paragraph{Proof for Group Lasso: } $\hat{\B{\beta}}$ designs herein a Group Lasso estimator, defined as a solution of  the Group Lasso Problem \eqref{group-problem}. It holds:
		\begin{align*}
		\begin{split}
		\frac{1}{n}  \|  \B{y} - \B{X} \hat{\B{\beta}} \|_2^2 + \lambda_G \sum_{g=1}^G \| \hat{\B{\beta}}_g \|_2 
		&\le \frac{1}{n}  \| \B{y} - \B{X} \B{\beta}^*  \|_2^2 + \lambda_G \sum_{g=1}^G \| \B{\beta}^*_g \|_2 = \frac{1}{n}  \| \B{\epsilon}   \|_2^2 + \lambda_G  \sum_{g=1}^G \| \B{\beta}^*_g \|_2.
		\end{split}
		\end{align*}
		By definition, the support of $\B{\beta}^* $ is included in $\mathcal{J}^*$ and $\mathcal{J}_0 \subset \{1, \ldots, G\}$ is the subset of indexes of the $s^*$ highest groups of $\B{h}$ for the L2 norm. It then holds:
		\begin{align}\label{rhs-group}
		\begin{split}
		\frac{1}{n} \| \B{X} \B{h} \|_2^2  
		&\le \frac{2}{n} ( \B{X}^T \B{\epsilon})^T \B{h} + \lambda_G  \sum_{g=1}^G \| \B{\beta}^*_g \|_2 - \lambda_G \sum_{g=1}^G \| \hat{\B{\beta}}_g \|_2\\
		&= \frac{2}{n} ( \B{X}^T \B{\epsilon})^T \B{h} + \lambda_G  \sum_{g \in \mathcal{J}^*} \| \B{\beta}^*_g \|_2 - \lambda_G \sum_{g=1}^G \| \hat{\B{\beta}}_g \|_2\\
		&\le \frac{2}{n} ( \B{X}^T \B{\epsilon})^T \B{h} + \lambda_G \sum_{g \in \mathcal{J}_0} \| \B{h}_g \|_{2} - \lambda_G \sum_{g \notin \mathcal{J}_0} \| \B{h}_g \|_{2}.
		\end{split}
		\end{align}
		We now upper-bound the quantity $( \B{X}^T \B{\epsilon})^T \B{h} $. We denote $\B{g} = \B{X}^T \B{\epsilon}$ and apply Cauchy-Schwartz inequality on each group to get:
		\begin{align}\label{equations-group-regression}
		( \B{X}^T \B{\epsilon})^T \B{h}
		\le \left|  \langle  \B{g},   \B{h}  \rangle  \right|
		\le \sum_{g=1}^G \left| \langle  \B{g}_g, \B{h}_g  \rangle  \right|
		\le \sum_{g=1}^G \| \B{g}_g \|_2 \|  \B{h}_g  \|_2,
		\end{align}
		Let us fix $g \le G$. We have denoted $n_g$ the cardinality of the set of indexes $\mathcal{I}_g$ of group $g$. It then holds $\forall \B{u}_g \in \mathbb{R}^{n_g}$:
		\begin{align}
		\begin{split}
		\mathbb{E}\left( \exp \left( \B{g}_g^T \B{u}_g \right) \right) 
		&=\mathbb{E}\left( (\B{X}^T \epsilon)_g^T  \B{u}_g \right)
		=\mathbb{E}\left( (\B{X}_g^T \epsilon)^T  \B{u}_g \right) 
		=\mathbb{E}\left( \epsilon^T  \B{X}_g \B{u}_g \right)\\ 
		&=\prod_{i=1}^{n_g} \mathbb{E}\left( \epsilon_i  (\B{X}_g \B{u}_g)_i \right) \text{ by independence}\\ 
		&\le \prod_{i=1}^{n_g} \exp \left( \frac{\sigma^2 (\B{X}_g \B{u}_g)_i^2 }{2} \right) \text{ since } \forall i, \epsilon_i  (\B{X}_g \B{u}_g)_i \sim \subG( \sigma^2 (\B{X}_g \B{u}_g)_i^2 )\\
		&= \exp \left( \frac{\sigma^2 \| \B{X}_g \B{u}_g \|_2^2}{2}  \right)
		= \exp \left( \frac{\sigma^2  \B{u}_g^T \B{X}_g^T \B{X}_g  \B{u}_g}{2}   \right)\\ 
		&\le \exp \left( \frac{n \sigma^2  \| \B{u}_g  \|_2^2}{2} \right) \text{ since } \mu_{\max}(\B{X}_g^T \B{X}_g ) \le n \text{ with Assumption } \ref{asu-col}.\\  
		\end{split}
		\end{align}
		Again, we use Theorem 2.1 from \citet{hsu2012tail}. By denoting $\B{I}_g$ the identity matrix of size $n_g$ it holds:
		$$\mathbb{P}\left( \| \B{I}_g \B{g}_g  \|_2^2 \ge n \sigma^2 \left( \tr(\B{I_g}) + 2 \sqrt{\tr(\B{I_g}^2)  t} + 2 ||| \B{I_g} ||| \right) \right) \le e^{-t}, \forall t > 0$$
		which gives:
		$$\mathbb{P}\left( \| \B{g}_g \|_2^2 \ge n \sigma^2 \left( \sqrt{n_g} + \sqrt{2 t} \right)^2 \right) \le e^{-t}, \forall t > 0$$
		which is equivalent from saying that:
		\begin{equation}\label{subGauss-group-regression}
		\mathbb{P}\left( \frac{1}{ \sqrt{n}} \| \B{g}_g \|_2^2 -  \sigma\sqrt{n_g}  \ge t \right)  \le \exp\left( \frac{-t^2}{2 \sigma^2} \right), \forall t > 0.
		\end{equation}
		We define the random variables
		$f_g = \max \left(0,  \frac{1}{ \sqrt{n}}  \| \B{g}_g \|_2 -  \sigma \sqrt{n_g} \right), \ g=1,\ldots,G$. $f_g$ satisfies the same tail condition than a sub-Gaussian random variable with variance $\sigma^2$ and we can apply Lemma \ref{upper-bound-sup}. To this end, we introduce a non-increasing rearrangement $(f_{(1)}, \ldots, f_{(G)})$  of $(|f_1|, \ldots, |f_G|)$ and a permutation $\psi$ such that $n_{\psi(1)} \ge \ldots \ge n_{\psi(G)}$---where we have defined the group sizes $n_1, \ldots, n_G$. In addition, we assume without loss of generality that $\| \B{ h }_1 \|_{2} \ge \ldots \| \B{ h }_G \|_{2}$ and we note the coefficients $\lambda_g^{(G)} = \sqrt{\log(2Ge / g)}$. Following Equation \eqref{equations-group-regression}, we obtain with probability at least $1-\delta$:
		
		\begin{align}\label{upper-bound-SG-group-regression}
		\begin{split}
		\frac{1}{ \sqrt{n}} ( \B{X}^T \B{\epsilon})^T \B{h}
		&\le \sum_{g=1}^G \frac{1}{ \sqrt{n}} \| \B{g}_g \|_2 \|  \B{h}_g  \|_2=\sum_{g=1}^G \left( \frac{1}{ \sqrt{n}} \| \B{g}_g \|_2 -  \sigma  \sqrt{n_g}  \right) \| \B{h}_g  \|_2 +  \sigma \sum_{g=1}^G  \sqrt{n_g} \| \B{h}_g  \|_2\\
		&\le \sum_{g=1}^G |f_{g} |  \|  \B{h}_g ||_2 +  \sigma  \sum_{g=1}^G \sqrt{n_g} \|  \B{h}_g  \|_2\\
		&\le \sup_{g=1,\ldots,G} \left\{\frac{f_{(g)}}{\sigma \lambda_g^{(G)} }  \right\}    \sum_{g=1}^G \sigma \lambda_g^{(G)} \|  \B{h}_{(g)} ||_2 +  \sigma\sum_{g=1}^G \sqrt{n_g} \|  \B{h}_g  \|_2\\
		&\le 12 \sigma \sqrt{\log(2/ \delta)} \sum_{g=1}^G \lambda_g^{(G)} \| \B{h}_{(g)} \|_{2} +  \sigma  \sum_{g=1}^G \sqrt{n_{g}} \|  \B{h}_{g}  \|_2 \text{ with Lemma \ref{upper-bound-sup}}\\
		&\le \left(12 \sigma \sqrt{ \log(2Ge /s^*) \log(1/ \delta) } + \sigma \sqrt{ \gamma m^* / s^* } \right) \left(  \sqrt{s^*}   \| \B{h}_{\mathcal{T}_0} \|_{2} + \sum_{g \notin \mathcal{J}_0} \| \B{h}_{g} \|_{2} \right),
		\end{split}
		\end{align}
		where we have followed Equation \eqref{upper-bound-SG-group} (replacing $LM$ by $\sigma$), we have defined $\mathcal{T}_0 = \cup_{g \in \mathcal{J}_0} \mathcal{I}_g$ as the subset  of all indexes across all the $s^*$ groups in $\mathcal{J}_0$, and where $m_0$ denotes the total size of the $s^*$ largest groups. Note that we have used the Stirling formula to obtain
		$$\sum_{g=1}^{s^*} \left(\lambda_g^{(G)} \right)^2  \le s^* \log(2Ge /s^*).$$
		Theorem \ref{cone-condition} defines  $\lambda_G = 24 \alpha \sigma  \sqrt{ \frac{1}{n}  \log(2Ge/s^*) \log(2/ \delta)} + \alpha \sigma \sqrt{ \frac{\gamma m^*}{s^* n} }$. By pairing Equations  \eqref{rhs-group} and \eqref{upper-bound-SG-group-regression} it holds with probability at least $1 - \delta$:
		\begin{equation}\label{fundamental-group}
		\begin{split}
		\frac{1}{n} \| \B{X} \B{h} \|_2^2  
		&\le \frac{\lambda_G}{\alpha}\left(  \sqrt{s^*}   \| \B{h}_{\mathcal{T}_0} \|_{2} + \sum_{g \notin \mathcal{J}_0} \| \B{h}_{g} \|_{2} \right) + \lambda_G \sum_{g \in \mathcal{J}_0} \| \B{h}_g \|_{2} - \lambda_G \sum_{g \notin \mathcal{J}_0} \| \B{h}_g \|_{2},
		\end{split}
		\end{equation}
		As a first consequence, Equation \eqref{fundamental-group} implies that with probability at least $1 -  \delta$
		\begin{equation*} 
		\begin{split}
		\lambda_G \sum_{g \notin \mathcal{J}_0} \| \B{h}_g \|_{2} - \frac{\lambda_G}{\alpha} \sum_{g \notin \mathcal{J}_0} \| \B{h}_{g} \|_{2} 
		&\le \lambda_G \sum_{g \in \mathcal{J}_0} \| \B{h}_g \|_{2} + \frac{\lambda_G}{\alpha}   \sqrt{s^*}   \| \B{h}_{\mathcal{T}_0} \|_{2} 
		\end{split}
		\end{equation*}
		which is equivalent to saying that with probability at least $1- \delta$:
		\begin{equation*} 
		\sum_{g \notin \mathcal{J}_0} \| \B{h}_g \|_{2}   \le  \frac{\alpha}{\alpha - 1}  \sum_{g \in \mathcal{J}_0} \| \B{h}_g \|_{2} +  \frac{\sqrt{s^*} }{\alpha - 1}  \| \B{h}_{\mathcal{T}_0} \|_{2},
		\end{equation*}
		that is $\B{h} \in \Omega\left(\mathcal{J}_0, \frac{\alpha}{\alpha-1}, \frac{\sqrt{s^*} }{\alpha - 1} \right)$ with probability at least $1- \delta$.
	\end{Proof}

	\section{Proof of Theorem \ref{main-results-regression}}  \label{sec: appendix_main-results-regression}
	
	\begin{Proof} Our bounds respectively follow from Equations \eqref{fundamental-lasso} and \eqref{fundamental-group}.
		
		\paragraph{Proof for Lasso: }
		As a second consequence of Equation \eqref{fundamental-lasso}, because $\alpha\ge 2$, it holds with probability at least $1 - \delta$:
		\begin{equation} 
		\begin{split}
		\frac{1}{n} \| \B{X} \B{h} \|_2^2  
		&\le \frac{\lambda}{\alpha} \left( \sqrt{k^*}  \| \B{h}_{S_0}  \|_2  +  \| \B{h}_{(S_0)^c}  \|_1 \right) + \lambda \| \B{h}_{S_0} \|_1 -\lambda \| \B{h}_{(S_0)^c}  \|_1\\
		&\le \frac{\lambda}{\alpha} \sqrt{k^*}   \| \B{h}_{S_0} \|_{2} + \lambda \| \B{h}_{S_0} \|_{1} \\
		&\le 2 \lambda \sqrt{k^*}   \| \B{h}_{S_0} \|_{2}
		\le 2 \lambda \sqrt{k^*}   \| \B{h} \|_{2}, 
		\end{split}
		\end{equation}
		where we have used Cauchy-Schwartz inequality on the $k^*$ sparse vector $\B{h}_{S_0} $.
		
		\smallskip
		\noindent
		The cone condition proved in Theorem \ref{cone-condition} gives $\B{h}= \hat{\B{\beta}} _1- \B{\beta}^* \in \Lambda \left( S_0, \ \gamma_1^*=\frac{\alpha}{\alpha - 1},  \ \gamma_2^*=\frac{\sqrt{k^*}}{\alpha - 1} \right)$. We can then use the restricted eigenvalue condition defined in Assumption \ref{asu4}.1$(k^*, \gamma^*)$---where we define $\kappa^* = \kappa \left(k^*, \gamma_1^*, \gamma_2^*\right)$. It then holds with probability at least $1 - \delta$:
		$$ \kappa^* \| \B{h} \|_2^2  \le  \frac{1}{n} \| \B{X} \B{h}\|_2^2 \le 2 \lambda \sqrt{k^*}   \| \B{h} \|_{2}.$$
		By using that $\lambda=24 \alpha \sigma \sqrt{ \frac{1}{n} \log(2pe/k^*) \log(2 / \delta)}$, we conclude that it holds with probability at least $1 - \delta$:
		\begin{equation*}
		\begin{split}
		&\|\B{h}\|_2^2 \lesssim 
		\left( \frac{ \alpha \sigma }{\kappa^*} \right)^2 \frac{ k^* \log\left( p/k^* \right) \log\left( 2/\delta \right) }{n}.
		\end{split}
		\end{equation*}
		
		\paragraph{Proof for Group Lasso: }
		Similarly, as a second consequence of Equation \eqref{fundamental-group}, it holds with probability at least $1 - \delta$:
		\begin{equation} 
		\begin{split}
		\frac{1}{n} \| \B{X} \B{h} \|_2^2  
		&\le \frac{\lambda_G}{\alpha} \sqrt{s^*}   \| \B{h}_{\mathcal{T}_0} \|_{2} + \lambda_G \sum_{g \in \mathcal{J}_0} \| \B{h}_g \|_{2} 
		\le 2 \lambda_G \sqrt{s^*}   \| \B{h} \|_{2}, 
		\end{split}
		\end{equation}
		where we have used Cauchy-Schwartz inequality to obtain: $\sum_{g \in \mathcal{J}_0} \| \B{h}_g \|_{2} \le  \sqrt{s^*} \| \B{h}_{\mathcal{T}_0} \|_{2}$
		
		\smallskip
		\noindent
		The cone condition proved in Theorem \ref{cone-condition} gives $\B{h} = \hat{\B{\beta}}_{L1-L2}  - \B{\beta}^*\in \Omega \left( \mathcal{J}_0,  \epsilon_1^*=\frac{\alpha}{\alpha-1},  \epsilon_2^* = \frac{\sqrt{s^*} }{\alpha - 1}  \right)$. We can then use the restricted eigenvalue condition defined in Assumption \eqref{asu4}.2$(s^*, \epsilon^*)$---where we have defined $\kappa^* = \kappa \left(s^*, \epsilon_1^*, \epsilon_2^*\right)$. It then holds:
		$$\kappa^* \| \B{h} \|_2^2  \le 2\lambda_G \sqrt{s^*}   \| \B{h} \|_{2}.$$
		We conclude, by using the definition of $\lambda_G = 24 \alpha \sigma \sqrt{ \frac{1}{n} \log(2Ge/s^*) \log(2/ \delta)} + \alpha \sigma \sqrt{ \frac{\gamma m^*}{ns^*}  }$, that it holds with probability at least $1 - \delta$:
		\begin{equation*}
		\begin{split}
		&\|\B{h}\|_2^2 \lesssim 
		\left( \frac{ \alpha \sigma }{\kappa^*} \right)^2 \frac{ s^* \log\left( G/s^* \right) \log\left( 2/\delta \right) + \gamma m^* }{n}.
		\end{split}
		\end{equation*}
	\end{Proof}

	\section {Bounds in expectation for Lasso and Group Lasso} \label{sec: appendix_main-corollary-regression}
	
	Theorem \ref{main-results-regression} holds for any $\delta \le 1$. Thus, we obtain by integration the following bounds in expectation. The proof is presented in Appendix \ref{sec: appendix_main-corollary}.
	\begin{cor} 
		The bounds presented in Theorem \ref{main-results} additionally holds in expectation, that is:
		\begin{equation*}
		\mathbb{E} \| \hat{\B{\beta}}_{1}  - \B{\beta}^*\|_2 \lesssim \frac{\alpha \sigma}{\kappa^*}  \sqrt{ \frac{k^* \log\left( p /k^* \right)}{n} },
		\end{equation*}
		\begin{equation*}
		\mathbb{E} \| \hat{\B{\beta}}_{L1-L2}  - \B{\beta}^*\|_2 \lesssim \frac{\alpha \sigma}{\kappa^*} \sqrt{ \frac{s^* \log\left( G / s^* \right) + \gamma m^*}{n} }.
		\end{equation*}
	\end{cor}
	The proof is identical to the one presented in Section \ref{sec: appendix_main-corollary}.

	\section{First order algorithm} \label{sec:first-order} 
	
	We propose herein a first-order algorithm to solves the tractable Problems \eqref{l1-problem}, \eqref{slope-problem} and \eqref{group-problem} when the number of variables is of the order of $100,000s$.
	
	\subsection{Smoothing the loss}\label{sec:smoothing}
	We note $g(\B{\beta}) = \frac{1}{n} \sum_{i=1}^n  f \left( \langle \B{x_i},  \B{\beta} \rangle ;  y_i \right)$. Problem \eqref{general} can be formulated as: $\min \limits_{ \B{\beta} \in \mathbb{R}^{p}  } g(\B{\beta}) + \Omega( \B{\beta} )$---we drop the L1 constrain in this section. The proximal method we propose assumes $g$ to be a differentiable loss with continuous $C$-Lipschitz gradient. However, the hinge loss and the quantile regression loss are non-smooth. We propose to use herein Nesterov's smoothing method  \citep{smoothing-math} to construct a convex function with continuous Lipschitz gradient $g^{\tau}$---$g^{\tau}_{\theta}$ for quantile regression---which approximates these losses for $\tau \approx 0$. 
	
	\paragraph{Hinge loss:}  For the hinge loss, let us first note that $\max(0,t) = \frac{1}{2}(t + |t|) =  \max_{|w| \le 1} \frac{1}{2}(t + wt)$ as this maximum is achieved for $\sign(x)$. Consequently, the hinge loss can be expressed as a maximum over the $L_\infty$ unit ball:
	$$g(\B{\beta}) = \frac{1}{n}\sum \limits_{i=1}^n \max(0, z_i) =  \max \limits_{\|\B{w}\|_{\infty} \le 1}  \frac{1}{2n}  \sum \limits_{i=1}^n \left[  z_i  + w_i z_i  \right]$$ 
	where $z_i = 1 - y_i \B{x}_i^T\B{\beta}, \ \forall i$. We apply the technique suggested by  \citet{smoothing-math} and define for $ \tau >0$ the smoothed version of the loss:
	\begin{align}\label{smooth-hinge-def}
	&g^{\tau}(\B{\beta}) =  \max \limits_{\|\B{w}\|_{\infty} \le 1} \frac{1}{2n}  \sum \limits_{i=1}^n  \left[  z_i   + w_i z_i  \right] - \frac{\tau}{2n} \|\B{w}\|_2^2.
	\end{align}
	Let $\B{w}^{\tau}(\B{\beta}) \in \mathbb{R}^n:  \ w^{\tau}_i(\B{\beta}) = \min\left( 1, \frac{1}{2\tau} | z_i | \right) \sign(z_i), \ \forall i$ be the optimal solution of the right-hand side of Equation \eqref{smooth-hinge-def}. The gradient of $g^{\tau}$ is expressed as:
	\begin{equation}
	\nabla g^{\tau}( \B{\beta} ) = - \frac{1}{2n} \sum \limits_{i=1}^{n}(1+w_i^{\tau}(\B{\beta}) )y_i \B{x}_i  \in \mathbb{R}^p
	\end{equation}
	and its associated Lipschitz constant is derived from the next theorem.
	\begin{theorem} \label{lipschitz}
		Let $\mu_{\max} (\frac{1}{n} \B{X}^T \B{X})$ be the highest eigenvalue of $\frac{1}{n} \B{X}^T \B{X}$. Then $\nabla g^{\tau}$ is Lipschitz continuous with constant $ C^{\tau} = \mu_{\max} (\frac{1}{n} \B{X}^T \B{X}) / 4 \tau$.
	\end{theorem}
	The proof is presented in Appendix  \ref{sec: appendix_lipschitz}. It follows \citet{smoothing-math} and uses first order necessary conditions for optimality. 
	
	\paragraph{Quantile regression:} The same method applies to the non smooth quantile regression loss. We first note that $\max\left( (\theta-1)t, \ \theta t \right)  = \frac{1}{2}((2\theta-1)t + |t|) =  \max_{|w| \le 1} \frac{1}{2}( (2\theta-1)t + wt)$. 
	Hence the smooth quantile regression loss is defined as $g_{\theta}^{\tau}( \B{\beta} ) =  \max \limits_{\|\B{w}\|_{\infty} \le 1}\frac{1}{2n} \sum \limits_{i=1}^{n}((2 \theta -1)\tilde{z}_i+w_i \tilde{z}_i) - \frac{\tau}{2n} \|\B{w}\|_2^2$ and its gradient is:
	$$\nabla g^{\tau}_{\theta}( \B{\beta} ) = - \frac{1}{2n} \sum \limits_{i=1}^{n}(2 \theta -1+\tilde{w}_i^{\tau}(\B{\beta}) ) \B{x}_i \in \mathbb{R}^{p}$$
	where we now have $\tilde{w}^{\tau}_i = \min\left( 1, \frac{1}{2\tau} | \tilde{z}_i | \right) \sign(\tilde{z}_i)$ with $\tilde{z}_i = y_i - \B{x}_i^T\B{\beta}, \ \forall i$. The Lipschitz constant of $\nabla g^{\tau}_{\theta}$ is still given by Theorem \ref{lipschitz}. 
	
	\subsection{Thresholding operators}\label{sec:thresholding}
	Following~\citet{nesterov2004introductorynew,FISTA}, for $D\ge C$, we upper-bound the smooth $g$ (or $g^{\tau}$) around any $\B{\alpha} \in \mathbb{R}^{p}$ with the quadratic form $Q_D(\B{\alpha},.)$ defined $\forall \B{\beta} \in \mathbb{R}^{p}$ as:
	\begin{equation*}\label{convex-continuous-gradient}
	g( \B{\beta} ) \le  Q_D(\B{\alpha}, \B{\beta}) = g(\B{\alpha}) + \nabla g(\B{\alpha})^T(\B{\beta}-\B{\alpha}) + \frac{D}{2} \| \B{\beta} - \B{\alpha} \|_2^2.
	\end{equation*}
	The proximal gradient method approximates the solution of Problem $\eqref{general}$  by solving the problem:
	\begin{align*} 
	\begin{split}   
	\hat{\B\beta} &\in \argmin_{ \B{\beta}  } \left\{ Q_D(\B{\alpha},\B{\beta}) + \Omega( \B{\beta} ) \right\}  ~ \iff ~ \hat{\B\beta} \in \argmin_{ \B{\beta}  }  \frac{1}{2} \left\| \B{\beta} - \left(\B{\alpha} - \tfrac{1}{D} \nabla Q_D(\B{\alpha})  \right) \right\|_2^2 + \tfrac{1}{D} \Omega( \B{\beta} ),
	\end{split} 
	\end{align*}
	which can be solved via the the following proximal operator (evaluated at $\mu=\frac{1}{D}$):
	\begin{equation}\label{thresh-op1}
	\mathcal{S}_{\mu \Omega }(\B{\eta}) := \argmin_{ \B{\beta} \in \mathbb{R}^p  }\frac{1}{2} \left\| \B{\beta} - \B{\eta}  \right\|_2^2 + \mu \Omega( \B{\beta} ).
	\end{equation}
	We discuss computation of  \eqref{thresh-op1} for the specific choices of $\Omega$ considered.
	
	\paragraph{L1 regularization: } When $\Omega( \B{\beta} )= \lambda \| \B{\beta} \|_1$, $\mathcal{S}_{\mu \Omega }(\B{\eta})$ is available via componentwise softhresholding, where the soft-thresholding operator is: $\argmin_{ u \in {\mathbb R}} \frac{1}{2}(u-c)^2 + \mu \lambda  | u| = \sign(c) (|c| - \mu \lambda)_+$.
	
	\paragraph{Slope regularization: }
	When $\Omega( \B{\beta} )=  \sum_{j=1}^p \tilde{\lambda}_j | \beta_{(j)} | $---where $\tilde{\lambda_j}=\eta \lambda_j$---we note that, at an optimal solution to Problem~\eqref{thresh-op1}, the signs of $\beta_{j}$ and $\eta_j$ are the same \citep{slope-proximal}. Consequently, we solve the following close relative to the isotonic regression problem~\citep{robertson1988order}:
	\begin{equation}\label{slope-thresholded}
	\min \limits_{\M{u}} ~~~~  \frac{1}{2}  \left\| \M{u} - \tilde{\B{\eta}} \right\|_2^2 +  \sum \limits_{j=1}^p \mu \tilde{\lambda}_j u_j,  
	~\sbt~ u_1 \ge \ldots \ge u_p \ge 0.
	\end{equation}
	where, $\tilde{\B{\eta}}$ is a decreasing rearrangement of the absolute values of  $\B{\eta}$. A solution $\hat{u}_{j}$ of Problem~\eqref{slope-thresholded} corresponds to $|\hat{\beta}_{(j)}|$, where $\hat{\B\beta}$ is a solution of Problem~\eqref{thresh-op1}. We use the software provided by~\citet{slope-proximal} in our experiments.
	
	\paragraph{Group L1-L2: }
	For $\Omega( \B{\beta} )= \lambda \sum_{g=1}^G \| \B{\beta}_g \|_{2}$, we consider the projection operator onto an L2-ball with radius $\mu\lambda$:
	\begin{align*} \label{key-algortihm1-group-bis}
	\tilde{\mathcal{S}}_{ \tfrac1{\mu\lambda} \| \cdot \|_{2}  }  (\B{\eta})  \in \argmin_{ \B{\beta}  } \frac{1}{2}\left\| \B{\beta} - \B{\eta}  \right\|_{2}^2  \ \ \sbt \;\ \ \frac1{\mu \lambda} \| \B{\beta} \|_{2} \le 1.
	\end{align*}
	From standard results pertaining to Moreau decomposition \citep{moreau1962fonctions, bach2011convex} we have:
	\begin{equation*} \label{l_inf_prox}
	\mathcal{S}_{\mu \lambda\| . \|_{2}} (\B\eta) = \B\eta -  \tilde{\mathcal{S}}_{\tfrac1{\mu \lambda} \| \cdot \|_{2}}(\B\eta) = \left( 1 - \frac{\mu\lambda}{\| \B{\eta} \|_2} \right)_+ \B{\eta}.
	\end{equation*}
	We solve Problem \eqref{thresh-op1} with Group L1-L2 regularization by noticing the separability of the problem across the different groups, and computing $\mathcal{S}_{\mu \lambda \| . \|_{2}} (\B\eta_g)$  for every $g = 1, \ldots, G$.
	
	\subsection{First order algorithm}\label{sec:FO}
	
	Let us denote the proximal gradient mapping $\B\alpha \mapsto \hat{\B\beta}$ by the operator: 
	$ \hat{\B\beta} :=  \Theta(\B\alpha).$
	The standard version of the proximal gradient descent algorithm performs the updates: $\B\beta_{t+1} = \Theta(\B\beta_{t})$ for $T \geq 1$. The accelerated gradient descent algorithm~\citep{FISTA}, which enjoys a faster convergence rate, performs updates with a minor modification.
	It starts with ${\B\beta}_{1} = \tilde{\B\beta}_{0}$, $q_{1} =1$ and then performs the updates:
	$ \tilde{\B\beta}_{t+1} = \Theta({\B\beta}_{t})$ where, 
	${\B\beta}_{t+1} = \tilde{\B\beta}_{T} + \frac{q_{t} -1}{q_{t+1}} (\tilde{\B\beta}_{t} - \tilde{\B\beta}_{t-1} )$
	and $q_{t+1} = (1 + \sqrt{1 + 4 q_t ^2})/{2}$. We perform these updates till some tolerance criterion is satisfied, or a maximum number of iterations is reached.
	

	\subsection {Proof of Theorem \ref{lipschitz} } \label{sec: appendix_lipschitz}
	
	\begin{Proof} 
		We fix $ \tau >0$ and denote $\B{X} = (\B{X}_1, \ldots, \B{X}_p) \in \mathbb{R}^{n \times p}$ the design matrix.
		
		\smallskip
		\noindent
		For $\B{\beta} \in \mathbb{R}^{p}$, we define $\B{w}^{\tau}(\B{\beta}) \in \mathbb{R}^n$ by:
		$$w^{\tau}_i(\B{\beta}) = \min\left( 1, \frac{1}{2\tau} | z_i | \right) \sign(z_i ), \ \forall i$$
		where $z_i = 1 - y_i \B{x}_i^T\B{\beta}, \ \forall i$. We easily check that
		$$\B{w}^{\tau}(\B{\beta}) =   \argmax \limits_{\|w\|_{\infty} \le 1} \frac{1}{2n} \sum \limits_{i=1}^n \left( z_i  + w_i z_i \right) - \frac{\tau}{2n} \|w\|_2^2. $$
		Then the gradient of the smooth hinge loss is
		\begin{equation*} \label{smoothed-gradient}
		\nabla g^{\tau}( \B{\beta} ) = - \frac{1}{2n} \sum \limits_{i=1}^{n}(1+w_i^{\tau}(\B{\beta})) y_i \B{x}_i \in \mathbb{R}^{p}.
		\end{equation*}
		For every couple $\B{\beta}, \B{\gamma}  \in \mathbb{R}^{p}$ we have:
		\begin{equation} \label{diff-gradient}
		\nabla g^{\tau}(\B{\beta}) - \nabla g^{\tau}(\B{\gamma})
		= \frac{1}{2n} \sum \limits_{i=1}^{n}( w_i^{\tau}(\B{\gamma})- w_i^{\tau}(\B{\beta}) )y_i \B{x}_i.
		\end{equation}
		For $\B{a}, \B{b}\in \mathbb{R}^{n}$ we define the vector $\B{a}*\B{b} = (a_i b_i)_{i=1}^n$. Then we can rewrite Equation \eqref{diff-gradient} as:
		\begin{equation} \label{diff-gradient-bis}
		\nabla g^{\tau}(\B{\beta}) - \nabla g^{\tau}(\B{\gamma})
		= \frac{1}{2n} \B{X}^T\left[  \B{y}* \left( \B{w}^{\tau}(\B{\gamma})- \B{w}^{\tau}(\B{\beta}) \right) \right]. 
		\end{equation}
		The operator norm associated to the Euclidean norm of the matrix $\B{X}$ is $\| \B{X} \| = \max_{ \| \B{z} \|_2 = 1}  \| \B{X}\B{z} \|_2$.
		
		\smallskip
		\noindent
		Let us recall that  $\| \B{X} \|^2 = \| \B{X}^T \|^2 =  \| \B{X}^T \B{X} \| = \mu_{\max}(\B{X}^T \B{X} )$ corresponds to the highest eigenvalue of the matrix $\B{X}^T \B{X}$. Consequently, Equation \eqref{diff-gradient-bis} leads to: 
		\begin{equation} \label{first-part}
		\| \nabla L^{\tau}(\B{\beta}) - \nabla L^{\tau}(\B{\gamma}) \|_2 \le
		\frac{1}{2n} \| \B{X} \|  \left\|   \B{w}^{\tau}(\B{\gamma})- \B{w}^{\tau}(\B{\beta}) \right\|_2.
		\end{equation}
		In addition, the first order necessary conditions for optimality applied to $ \B{w}^{\tau}(\B{\beta})$ and $\B{w}^{\tau}(\B{\gamma})$ give:
		\begin{equation} \label{first-order-1}
		\sum_{i=1}^n \left\{ \frac{1}{2n}(1-y_i \B{x}_i^T\B{\beta} )- \frac{\tau}{n} w_i^{\tau}(\B{\beta} ) \right\} \left\{  w_i^{\tau}(\B{\gamma}) - w_i^{\tau}(\B{\beta})  \right\}\le 0,
		\end{equation}
		\begin{equation} \label{first-order-2}
		\sum_{i=1}^n \left\{ \frac{1}{2n}(1-y_i \B{x}_i^T\B{\gamma}) - \frac{\tau}{n} w_i^{\tau}(\B{\gamma}) \right\} \left\{  w_i^{\tau}(\B{\beta}) - w_i^{\tau}(\B{\gamma})  \right\}\le 0.
		\end{equation}
		Then by adding Equations \eqref{first-order-1} and  \eqref{first-order-2} and rearranging the terms we have:
		\begin{align*}
		\tau \|  \B{w}^{\tau}(\B{\gamma})- \B{w}^{\tau}(\B{\beta})   \|_2^2 
		&\le \frac{1}{2} \sum_{i=1}^n  y_i \B{x}_i^T ( \B{\beta} - \B{\gamma} ) \left(  w_i^{\tau}(\B{\gamma}) - w_i^{\tau}(\B{\beta})  \right) \\
		&\le  \frac{1}{2} \| \B{X}  \left( \B{\beta} - \B{\gamma}   \right) \|_2   \|  \B{w}^{\tau}(\B{\gamma}) - \B{w}^{\tau}(\B{\beta})   \|_2 \\
		&\le \frac{1}{2} \| \B{X} \|   \|  \B{\beta} - \B{\gamma}  \|_2     \|  \B{w}^{\tau}(\B{\gamma}) - \B{w}^{\tau}(\B{\beta})   \|_2,
		\end{align*}
		where we have used Cauchy-Schwartz inequality. We then have:
		\begin{equation}\label{second-part}
		\|  \B{w}^{\tau}(\B{\gamma})- \B{w}^{\tau}(\B{\beta})   \|_2
		\le \frac{1}{2\tau} \| \B{X} \|   \|  \B{\beta} - \B{\gamma}  \|_2.
		\end{equation}
		We conclude the proof by combining Equations \eqref{first-part} and \eqref{second-part}:
		\begin{align*}
		\| \nabla L^{\tau}(\B{\beta}) - \nabla L^{\tau}(\B{\gamma}) \|_2
		&\le \frac{1}{4n\tau} \| \B{X} \|^2   \|  \B{\beta}- \B{\gamma}  \|_2\\
		&= \frac{ \mu_{\max}(\frac{1}{n} \B{X}^T \B{X} )  }{4 \tau} \|  \B{\beta} - \B{\gamma}  \|_2.
		\end{align*}
	\end{Proof}

	\section{Simulations} \label{sec:simu}
	
	We compare the sparse estimators studied herein with standard baselines when the signal is sparse or group-sparse. We consider the 3 examples below with an increasing number of variables up to $100,000s$.  The computational tests were performed on a computer with Xeon $2.3$GhZ processors, $1$ CPUs, $16$GB RAM per CPU.

	\subsection{Example 1: sparse binary classification with hinge and logistic losses} Our first experiments compare L1 and Slope estimators with an L2 baseline for sparse binary classification problems. We use both the logistic and hinge losses. Our hypothesis is that (i) the estimators performance will only be affected by the statistical difficulty of the problem, not by the choice of the loss function and (ii) sparse regularizations will outperform their non-sparse opponents. 
	
	\paragraph{Data Generation:} We consider $n$ samples from a multivariate Gaussian distribution with 
	covariance matrix $\B\Sigma =((\sigma_{ij}))$ with $\sigma_{ii}=1$, $\sigma_{ij} = \rho$ if $1\le i \ne j \le k^*$ and $\sigma_{ij}=0$ otherwise. Half of the samples are from the $+1$ class and have mean $\B{\mu}_+ = (\B{\delta}_{k^*}, \ \mathbf{0}_{p-k^*} )$ where $\delta>0$. A smaller $\delta$ makes the statistical setting more difficult since the two classes get closer. The other half are from the $-1$ class and have mean $\B\mu_- = - \B\mu_+$.   We standardize the columns of the input matrix $\M{X}$ to have unit L2-norm.
	
	\medskip
	\noindent
	Following our high-dimensional study, we set $p \gg n$ and consider a sequence of increasing values of $p$. We study the effect of making the problem statistically harder by considering two settings, with a small and a large $\delta$.
	
	\paragraph{Competing methods:} We compare 3 approaches:
	
	\noindent
	$\bullet$ Method \textbf{(a)} computes a family of L1 regularized estimators for a decreasing geometric sequence of regularization parameters $\eta_0> \ldots > \eta_M$. We start from $\eta_{0} = \max_{j \in [p]} \sum_{i\in[n]}  |x_{ij}|$  so that the solution of Problem \eqref{l1-problem} is $\B{0}$ and we fix $\eta_M < 10^{-4} \eta_0$. When $f$ is the logistic loss, we use the first order algorithm presented in Section \ref{sec:FO}. When $f$ is the hinge loss, we directly solve the Linear Programming (LP) L1-SVM problem with the commercial LP solver \textsc{Gurobi} version $6.5$ with Python interface. We present an LP reformulation of the L1-SVM problem in Appendix \ref{sec:L1-ref}.
	
	\medskip
	
	\noindent
	$\bullet$ Method \textbf{(b)} computes a family of Slope regularized estimators, using the first order algorithm presented in Section \ref{sec:FO}. The Slope coefficients $\left\{ \lambda_j \right\}$ are defined in Theorem \ref{main-results}; the sequence of parameters $\left\{\eta_i\right\}$ is identical to method \textbf{(a)}. When $f$ is the hinge-loss, we consider the smoothing method defined in Section \ref{sec:smoothing} with a coefficient $\tau= 0.2$.
	
	\medskip
	
	\noindent
	$\bullet$ Method \textbf{(c)} returns a family of L2 regularized estimators with \textsc{scikit-learn} package: we start from $\eta_0 = \max_i \left\{ \| \B{x_i} \|_2^2 \right\}$ as suggested in \citet{warm-start}---and $\eta_M < 10^{-4} \eta_0$.

	\subsection{Example 2: group-sparse binary classification with hinge loss}
	Our second example considers classification problems where sparsity is structured. We compare the performance of two coefficient-based regularizations with two group regularizations. Our hypothesis is that (i) group regularizations outperform their opponents (ii) the gap in performance increases with the statistical difficulty of the problem. 
	
	\paragraph{Data Generation:}
	The $p$ covariates are drawn from a multivariate Gaussian and divided into $G$ groups of the same size $g_*$. Covariates have pairwise correlation of $\rho$ within each group, and are uncorrelated across groups. Half of the samples are from the $+1$ class with mean $\B\mu_+ = \left( \B{\delta}_{g_*}, \ldots, \B{\delta}_{g_*},  \mathbf{0}_{g_*}, \ldots, \mathbf{0}_{g_*}\right)$ where $s^*$ groups are relevant for classification; the remaining samples from class $-1$ have mean $\B\mu_- = -\B\mu_+$. The columns of the input matrix are standardized to have unit L2-norm. Similar to Example 1, we consider a sequence of increasing values of $p$ and study the effect of making the problem statistically harder by considering a small and a large $\delta$.
	
	\paragraph{Competing methods:} We compare the L1 and Slope regularized methods \textbf{(a)} and \textbf{(b)} described above with the two following group regularizations:
	\medskip
	
	\noindent
	$\bullet$ Method \textbf{(d)} computes a family of Group L1-L2 estimators with the first order algorithm presented in Section \ref{sec:FO}. We use the same sequence of regularization parameters as method \textbf{(a)}.
	\medskip
	
	\noindent
	$\bullet$ Method \textbf{(e)} considers an alternative Group L1-$L_\infty$ regularization \citep{bach2011convex}---discussed in Appendix \ref{sec:L1-Linf-ref}. We start from $\eta_0 =\max_{g \in [G]} \sum_{j \in \mathcal{I}_g} \sum_{i=1}^n |x_{ij}|$ and solve the LP formulation presented with the \textsc{Gurobi} solver.

	\begin{figure*}[h!]
		\centering
		{\sf Example 1 with hinge loss for $n=100$, $k^*=10$, $\delta=0.5$, $\rho=0.1$, $p \gg n$}
		\begin{tabular}{c c}
			\includegraphics[width=0.48\textwidth,height=0.18\textheight,  trim = .2cm 0cm .3cm .3cm,  clip = true ]{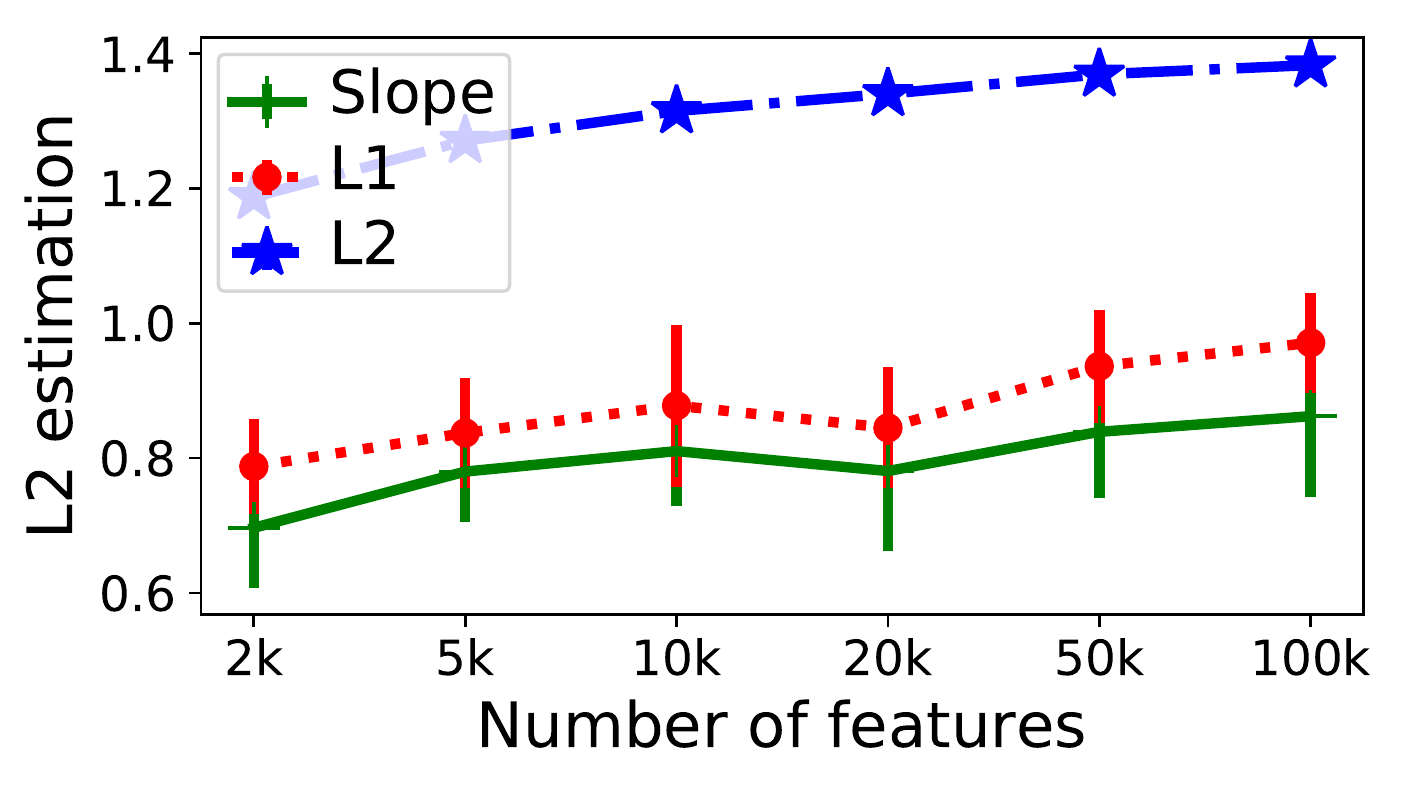}\label{fig:figc}&
			\includegraphics[width=0.48\textwidth,height=0.18\textheight,  trim = 0cm 0cm .3cm .3cm, clip = true ]{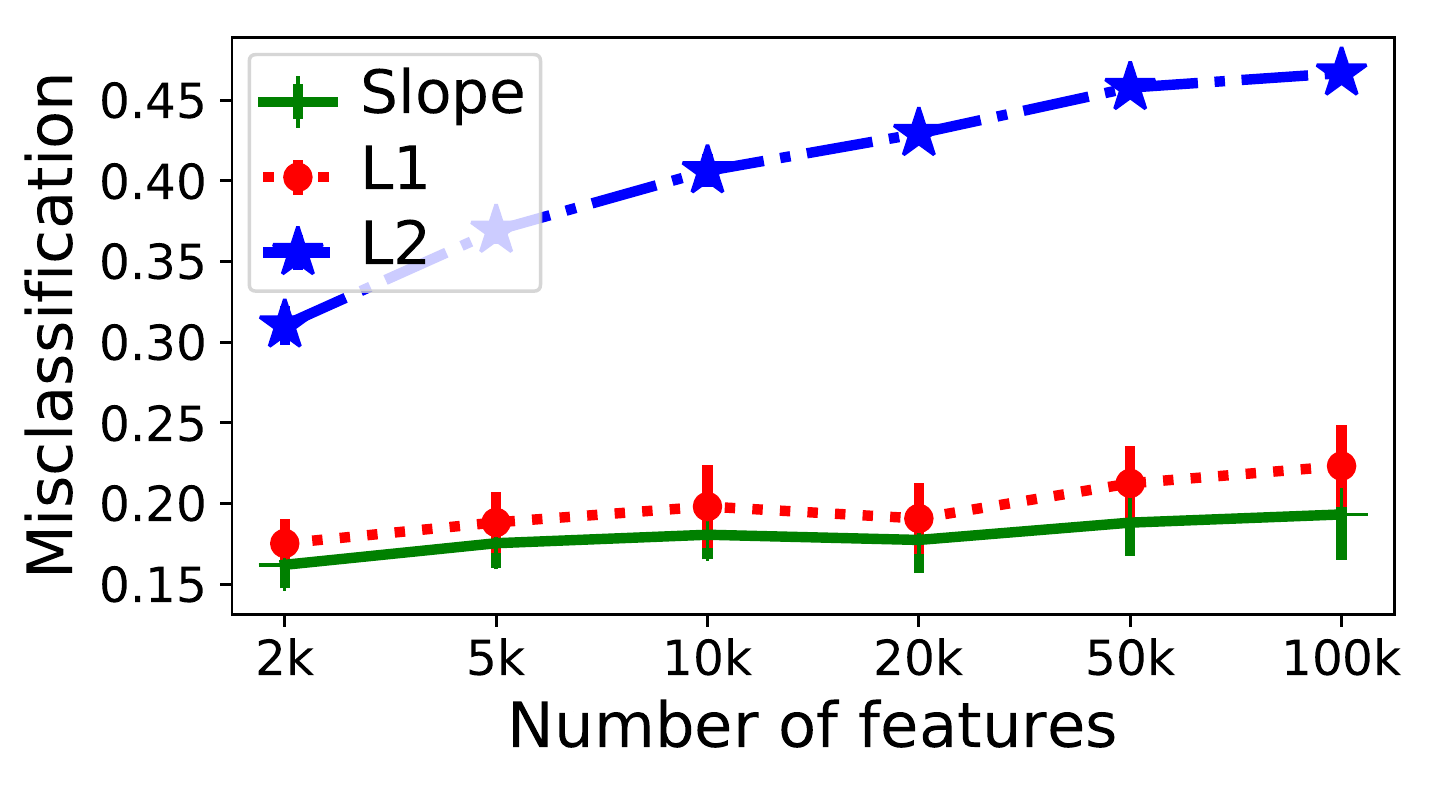}\label{fig:figd} \\
		\end{tabular}
		\smallskip\\
		{\sf Example 2 with hinge loss for $n=100$, $s^*=10$, $g_*=20$, $\delta=0.2$, $\rho=0.1$, $p \gg n$}
		\begin{tabular}{c c}
			\includegraphics[width=0.48\textwidth,height=0.18\textheight,  trim = .2cm 0cm .3cm .3cm,  clip = true ]{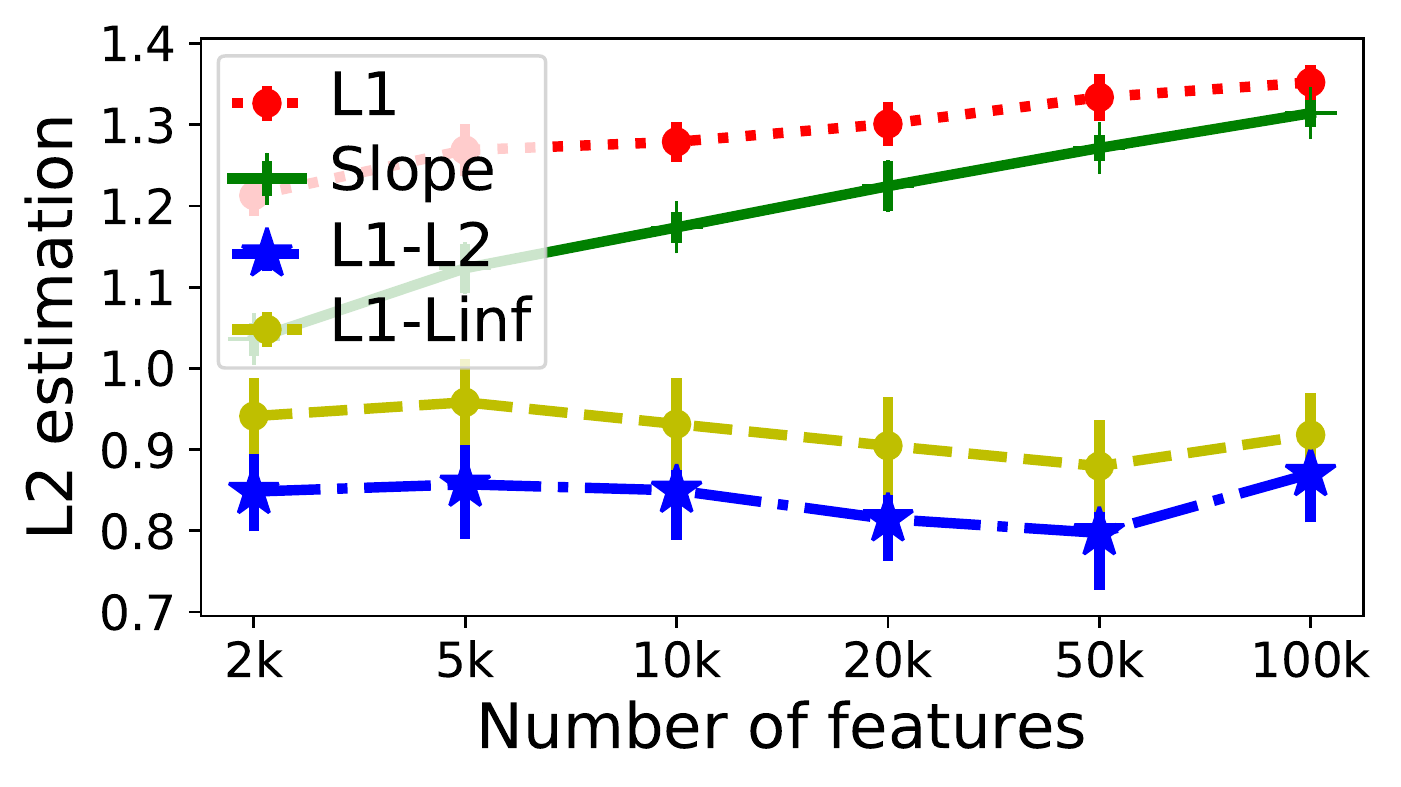}\label{fig:fige}&
			\includegraphics[width=0.48\textwidth,height=0.18\textheight,  trim = 0cm 0cm .3cm .3cm, clip = true ]{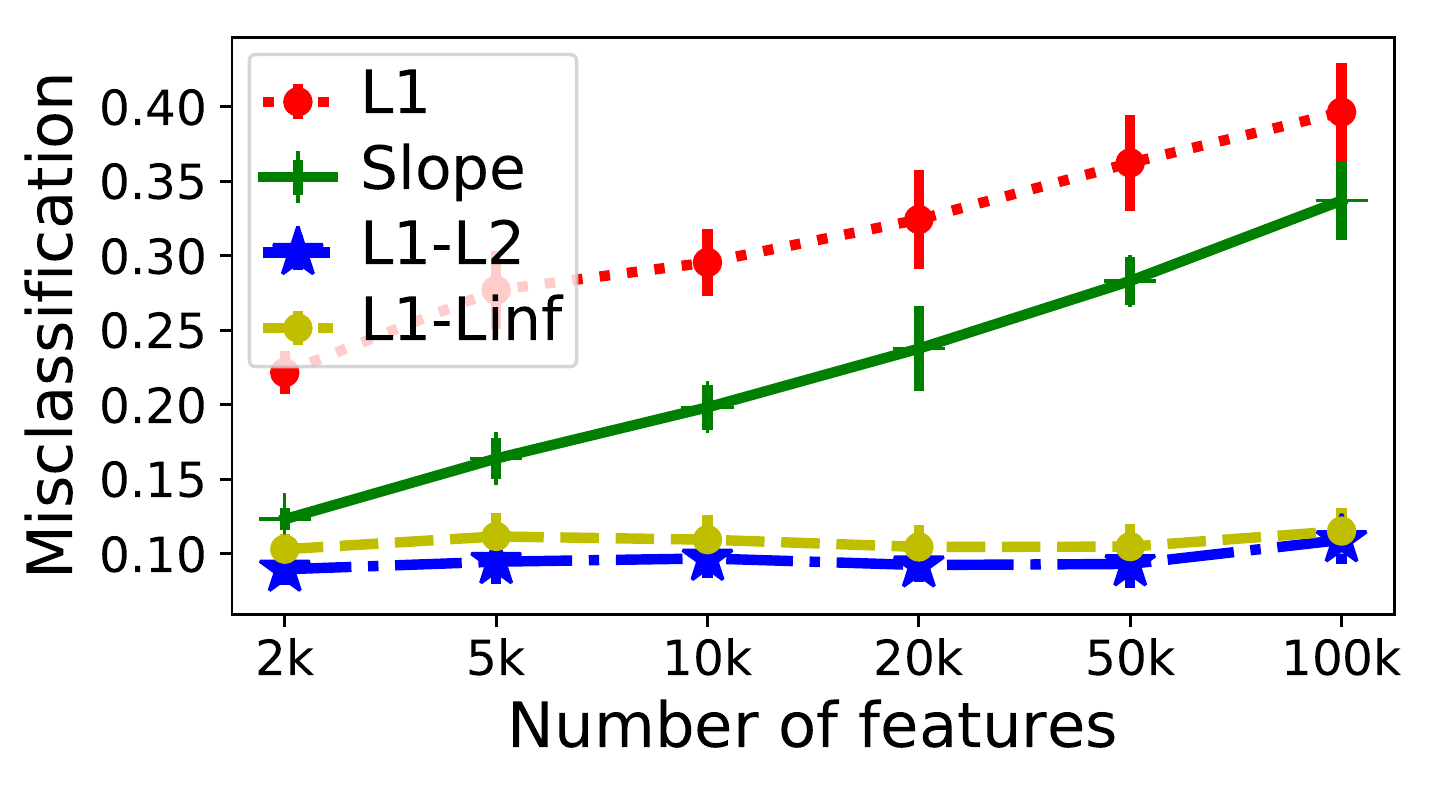}\label{fig:figf} \\
		\end{tabular}
		\smallskip\\
		{\sf Example 3 with quantile loss for $n=100$, $k^*=10$, $SNR=1$, $\rho=0.1$, $p \gg n$}
		\begin{tabular}{c c}
			\includegraphics[width=0.48\textwidth,height=0.18\textheight,  trim = .2cm 0cm .3cm .3cm,  clip = true ]{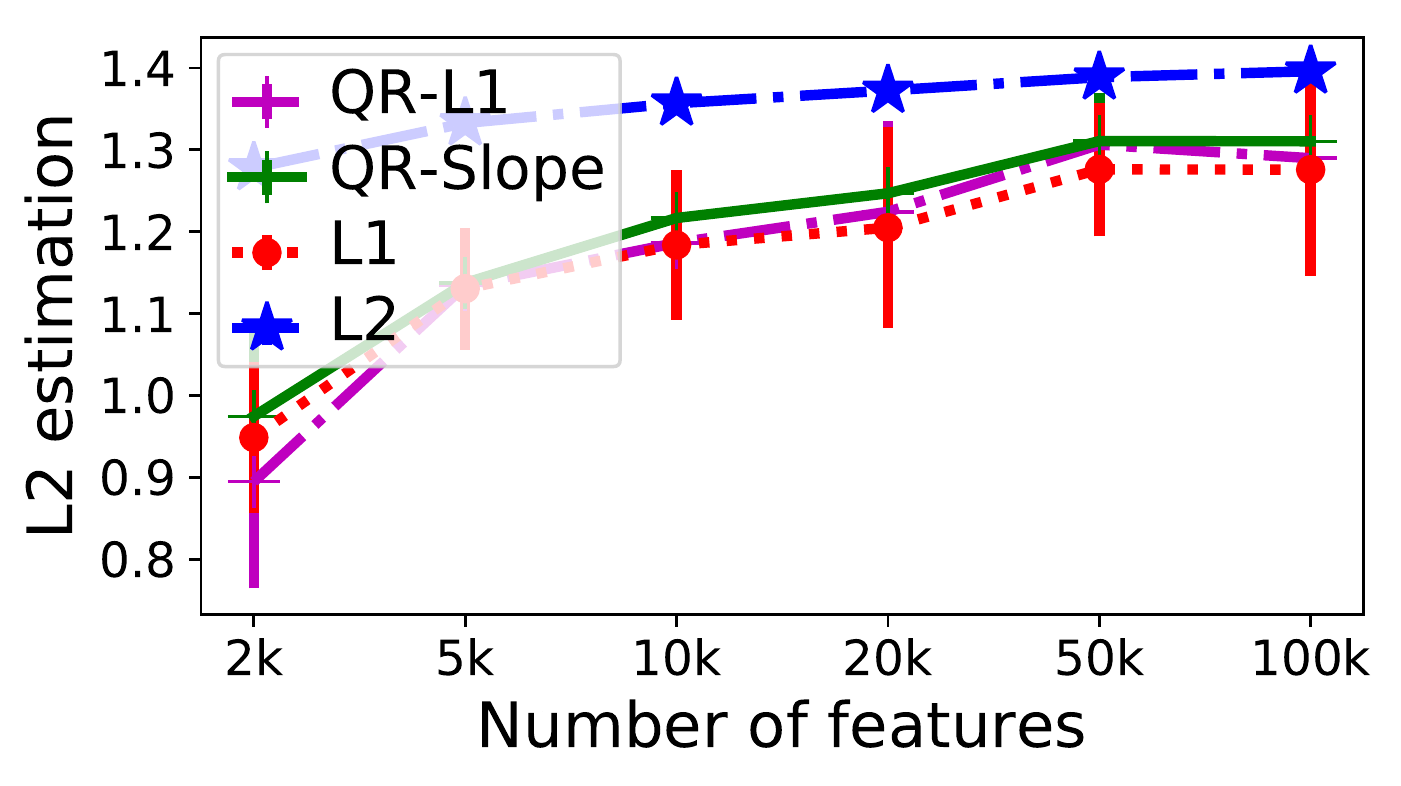}\label{fig:figg}&
			\includegraphics[width=0.48\textwidth,height=0.18\textheight,  trim = 0cm 0cm .3cm .3cm, clip = true ]{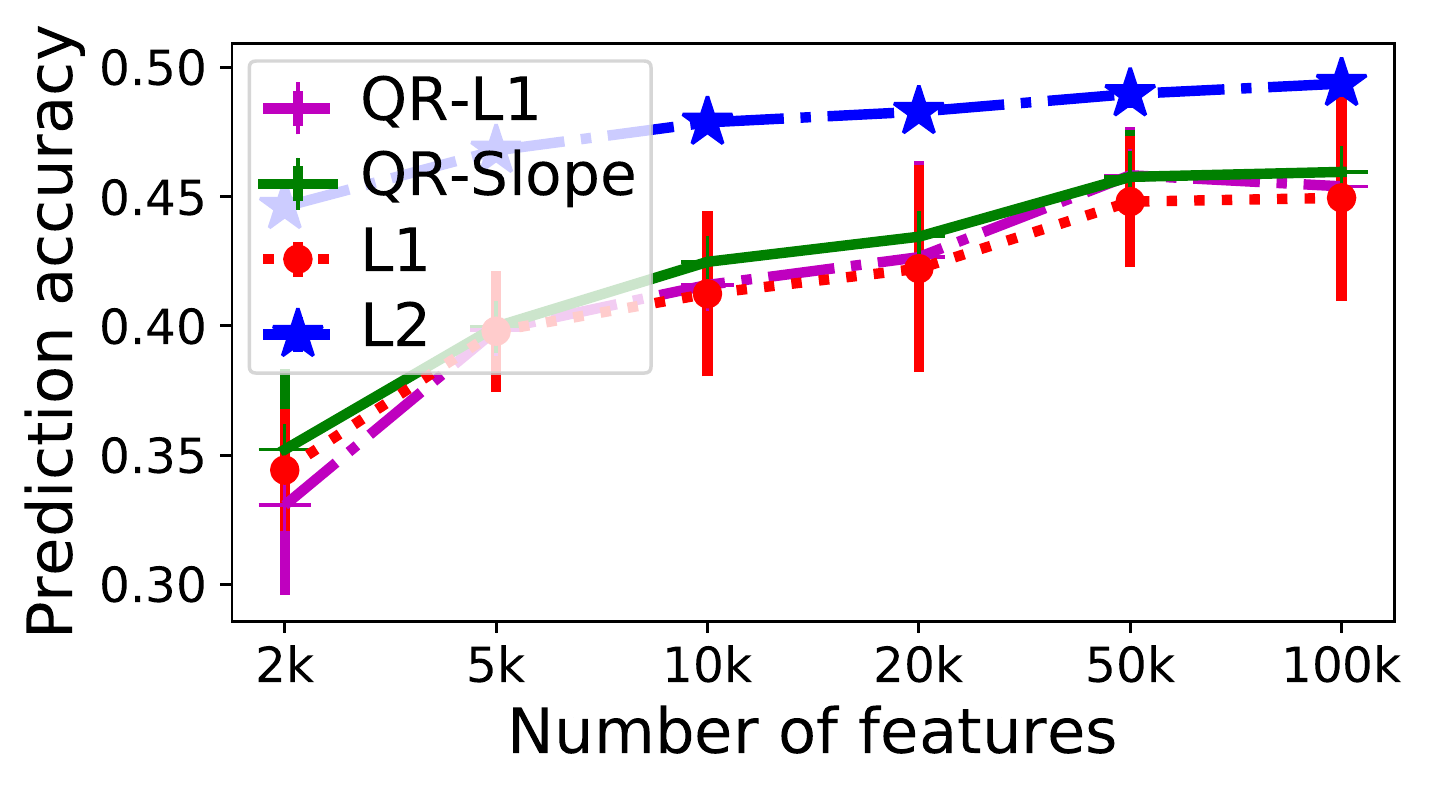}\label{fig:figh} \\
		\end{tabular}
		
		\caption{\small{ {[Top panel] L1 and Slope outperform L2 and show impressive gains for estimating the theoretical minimizer $\B{\beta}^* $ while achieving lower misclassification errors. Slope slightly performs better than L1 for the statistically hard settings.
					[Bottom middle panel] For small values of $p$, L1 and Slope compete with group regularizations. As $p$ increases, group regularizations exhibit their statistical superiority and Group L1-L2 appears as the overall winner. [Bottom panel] L1 and Slope regularized quantile regression compete with Lasso in the heteroscedastic regression case, while outperforming Ridge for both L2 estimation and prediction accuracy.
		}	}	}
		\label{fig:results}
	\end{figure*}
	
	\subsection{Example 3: sparse linear regression with heteroscedastic noise and quantile loss}
	Our last experiments compare L1 and Slope regularizations with quantile regression loss with Lasso and Ridge for regression settings. Our experiments draw inspiration from \citet{wang2013l1}: the authors showed the computational advantages of L1 regularized least-angle deviation (the quantile regression loss evaluated at $\theta= 1/2$) over Lasso for noiseless and Cauchy noise regimes. They additionally reported that the former is outperformed by Lasso for standard Gaussian linear regression. We consider herein a more challenging heteroscedastic regime---i.e. the noise is not identically distributed. Our hypothesis is that (i) L1 and Slope regularized quantile regression estimators perform similar to Lasso (ii) Ridge is outperformed by all its sparse opponents. 
	
	\paragraph{Data Generation: } We consider $n$ samples from a multivariate Gaussian distribution with 
	covariance matrix $\B\Sigma =((\sigma_{ij}))$ with $\sigma_{ij} = \rho^{|i-j|}$ if $i \ne j$ and $\sigma_{ij}=1$ otherwise. The columns of $\M{X}$are standardized to have unit L2-norm. Half of the noise observations are Gaussian and the rest is set to $0$. That is, we generate $\B{y} = \mathbf{X}\B{\beta}^* +\B{\epsilon}$  where ${\epsilon}_{i} \stackrel{\text{iid}}{\sim} N(0,\sigma^2)$ for $N/2$ randomly drawn indexes and ${\epsilon}_{i}=0$ otherwise. We set $\B{\beta^*} = (\B{\delta}_{k^*}, \ \mathbf{0}_{p-k^*} )$ and define the signal-to-noise (SNR) ratio of the problem as $\text{SNR} = \|\mathbf{X} \B{\beta}^{*}\|_2^2 / \sigma^2$. A low SNR makes the problem statistically harder. Similar to Examples 1 and 2, we consider two settings with a low and a large SNR.
	
	\paragraph{Competing methods: } We compare 4 approaches. We first consider L1 and Slope methods \textbf{(a)} and \textbf{(b)}---where we replace the hinge loss with the least-angle deviation loss. Note that in the case of L1 regularization, we directly solve the LP formulation presented in Appendix \ref{sec:L1-LAD}. We additionally introduce methods \textbf{(e)} and \textbf{(f)}, which run Lasso and Ridge using the  \textsc{scikit-learn} package: we set $\eta_0 = \| \B{X}^T \B{y}\|_{\infty}$ for Lasso so that the Lasso estimator is $\B{0}$; $\eta_0$ is set to be the highest eigenvalue of $\B{X}^T\B{X}$ for Ridge.

	
	\subsection{Metrics} Our theoretical results suggest to compare the estimators for the L2 estimation error $\left\| \frac{\hat{\B{\beta}}  }{\| \hat{\B{\beta}} \|_2 } - \frac{\B{\beta}^*  }{\| \B{\beta}^*  \|_2 }   \right\|_2,$ where $\B{\beta}^* $ is the theoretical minimizer. When it is not known in closed-form (e.g. for Examples 1 and 2), $\B{\beta}^* $ is computed on a large test set with $10,000$ samples restricted to the $k^*$ columns relevant for classification: we use the loss considered and a very small regularization coefficient  for computational stability. We also report an additional metric, namely the test misclassification performance for classification experiments (Examples 1 and 2) and the prediction accuracy $\frac{1}{n}\| \B{X} \hat{\B{\beta}} - \B{X} \B{\beta}^*  \|_2$ for regression experiments (Example 3). For a given method, we compute both test metrics for the estimator which achieves the lowest score for this additional metric on an independent validation set of size $10,000$.  Our findings are presented in Figure \ref{fig:results}. We report the mean and standard deviations values of each test metrics averaged over $10$ iterations.
	
	\begin{figure}[h!]
		\centering
		{\sf Example 1 with logistic loss for $n=100$, $k^*=10$, $\delta=1$, $\rho=0.1$, $p \gg n$}
		\begin{tabular}{c c}
			\includegraphics[width=0.48\textwidth,height=0.18\textheight,  trim = .2cm 0cm .3cm .3cm,  clip = true ]{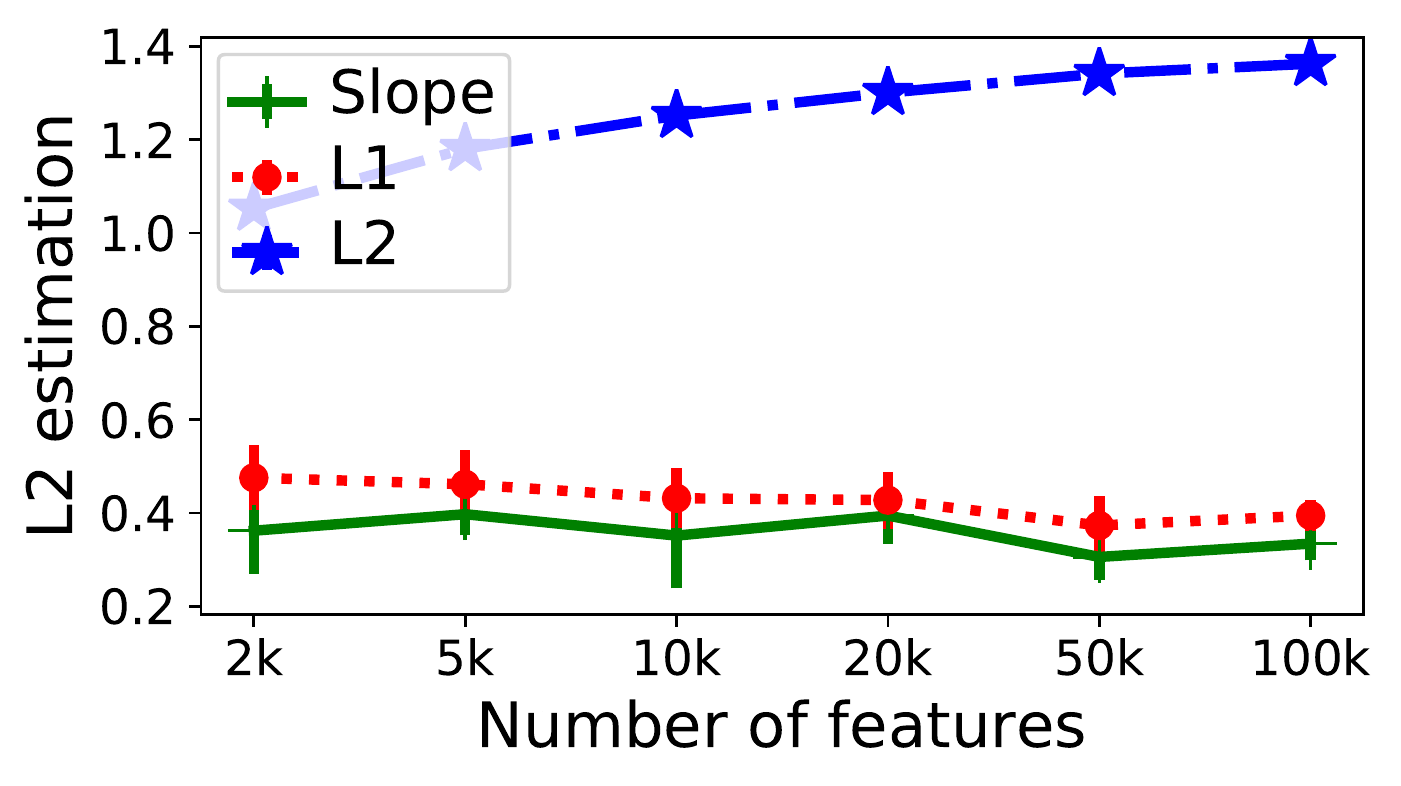}\label{fig:figa}&
			\includegraphics[width=0.48\textwidth,height=0.18\textheight,  trim = 0cm 0cm .3cm .3cm, clip = true ]{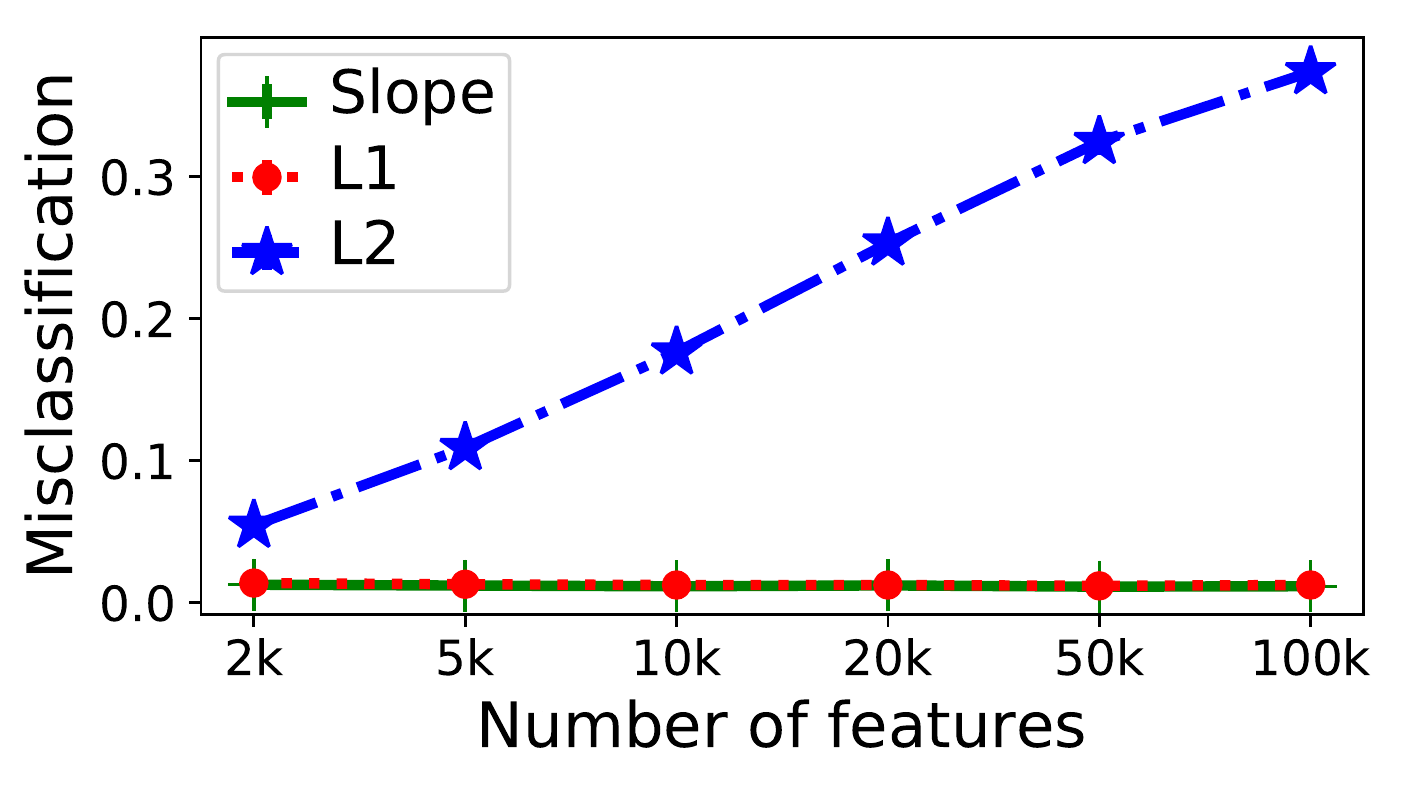}\label{fig:figb} \\
		\end{tabular}
		{\sf Example 2 with hinge loss for $n=100$, $s^*=10$, $g_*=20$, $\delta=0.4$, $\rho=0.1$, $p \gg n$}
		\begin{tabular}{c c}
			\includegraphics[width=0.48\textwidth,height=0.18\textheight,  trim = .2cm 0cm .3cm .3cm,  clip = true ]{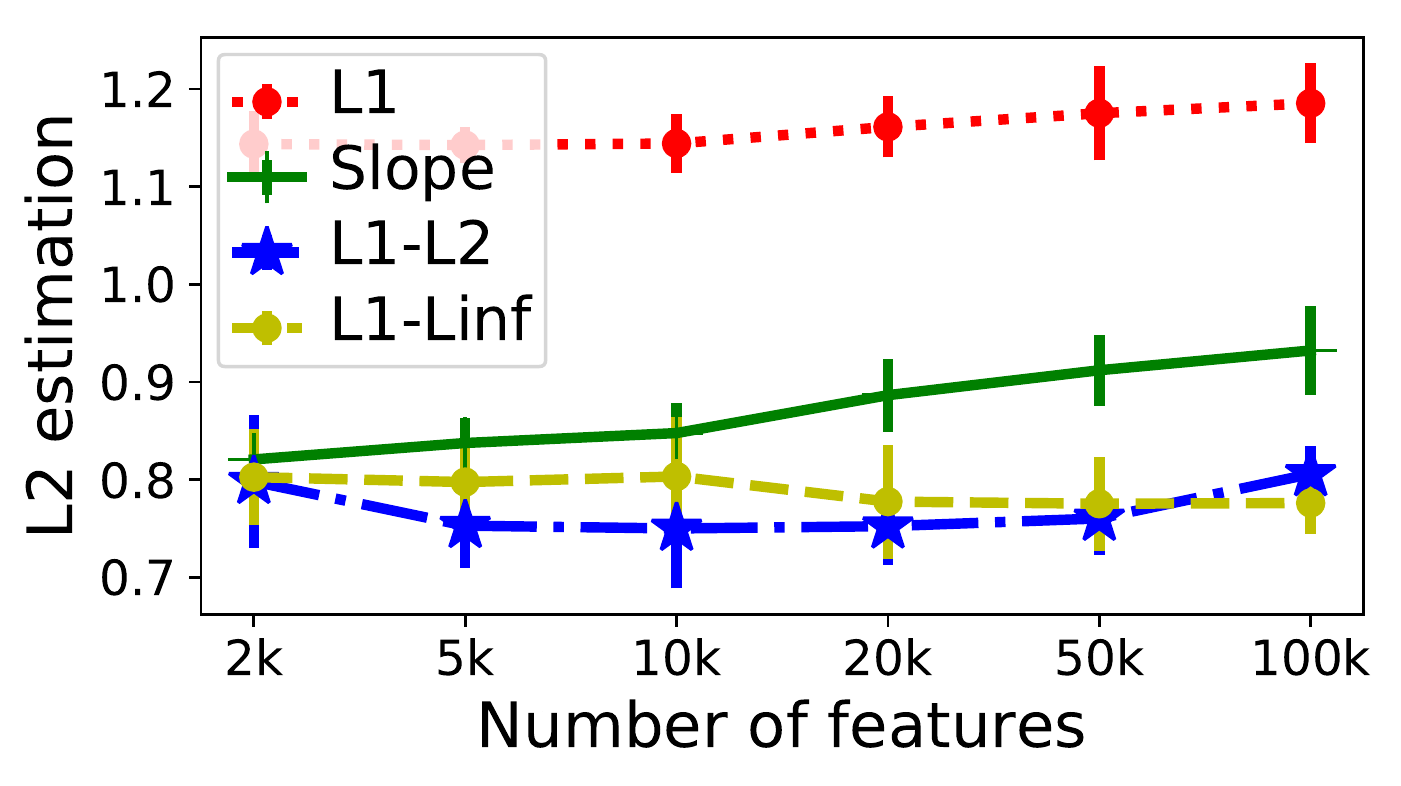}\label{fig:figi}&
			\includegraphics[width=0.48\textwidth,height=0.18\textheight,  trim = 0cm 0cm .3cm .3cm, clip = true ]{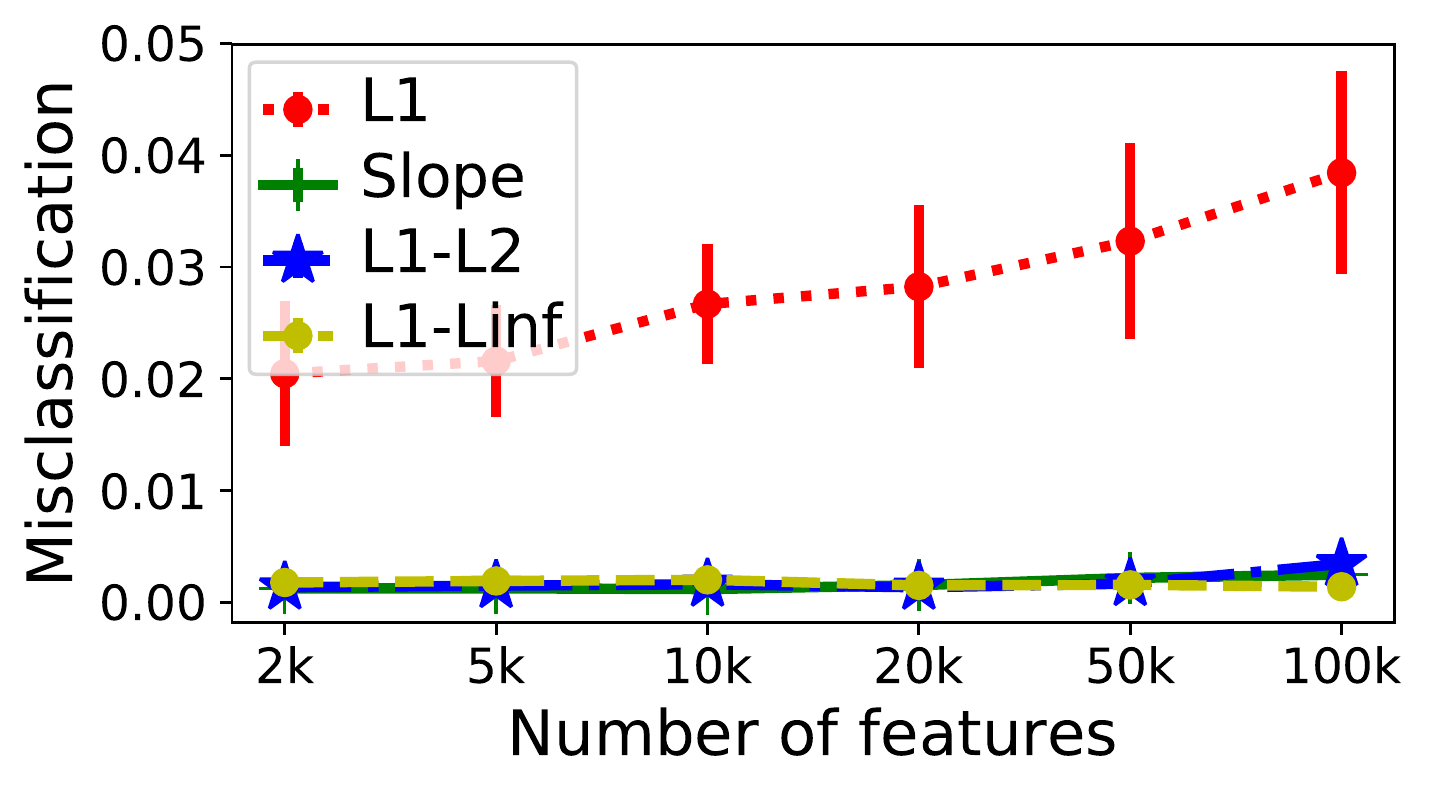}\label{fig:figj} \\
		\end{tabular}
		\smallskip\\
		{\sf Example 3 with quantile loss for $n=100$, $k^*=10$, $SNR=2$, $\rho=0.1$, $p \gg n$}
		\begin{tabular}{c c}
			\includegraphics[width=0.48\textwidth,height=0.18\textheight,  trim = .2cm 0cm .3cm .3cm,  clip = true ]{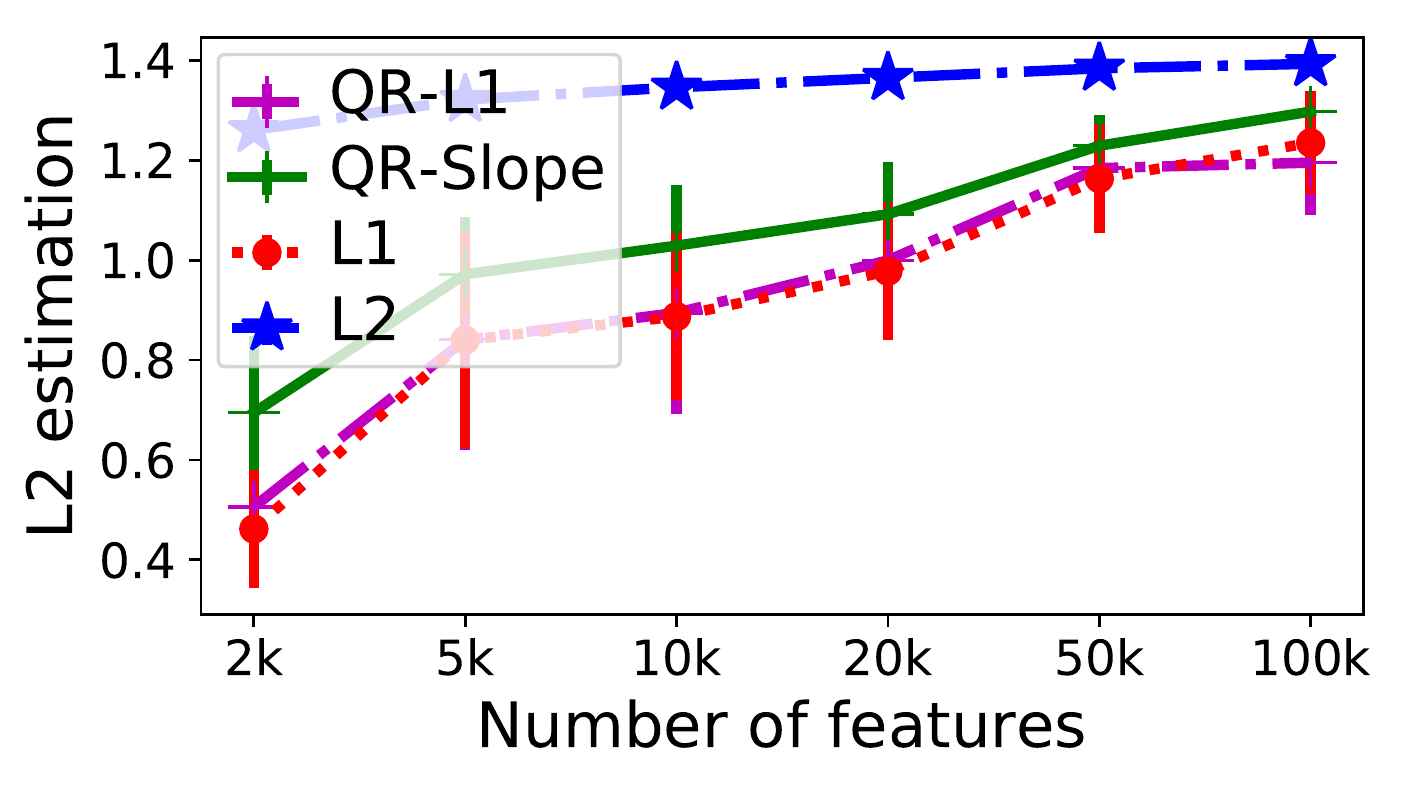}\label{fig:figk}&
			\includegraphics[width=0.48\textwidth,height=0.18\textheight,  trim = 0cm 0cm .3cm .3cm, clip = true ]{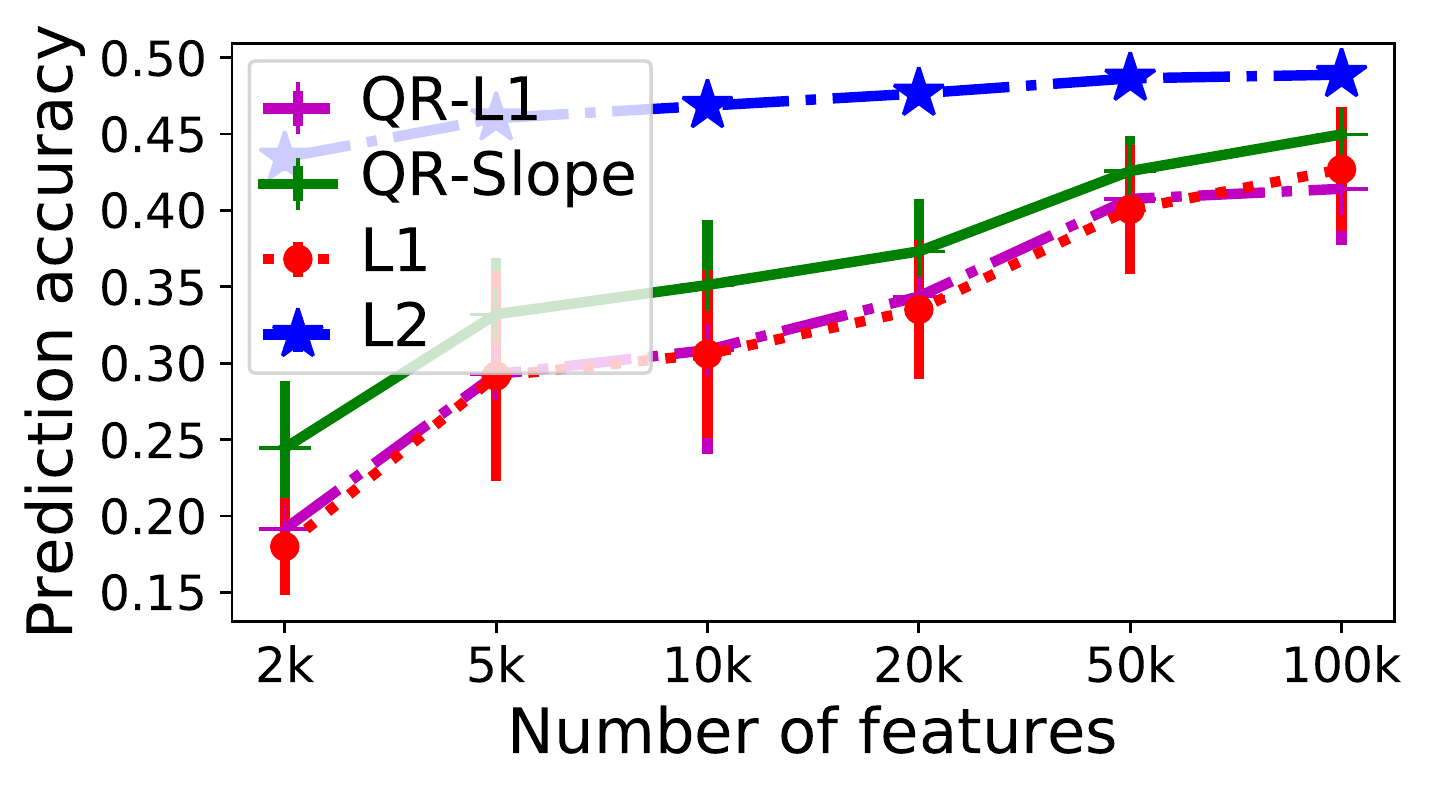}\label{fig:figl} \\
		\end{tabular}
		
		\caption{\small{ {[Top panel] All three estimators perform better when the statistical settings are simpler. The use of the logistic loss does not affect the relative performance of the estimators. [Middle panel] Slope can compete with group regularizations when the distance between the two classes increases. However the gap in performance increases for large values of $p$. [Bottom panel] When the $SNR$ increases,  Slope performance slightly decreases while L1 regularized quantile regression and Lasso exhibit very similar behaviors.
		}	}	}
		\label{fig:results-appendix}
	\end{figure}

	\subsubsection{Results} 
	
	We derive three main learning from our experiments, which complement our theoretical findings. Note that some additional experiments are presented in Appendix \ref{sec:simu-app}.
	
	\medskip
	
	\noindent
	$\bullet$ First, for sparse binary classification Example 1, our experiments show that L2 is outperformed by both L1 and Slope. In particular, L2 performs close to random guess for $\delta=0.5$ and $p > 20k$. Slope seems to achieve slightly better performance than L1 for both L2 estimation and misclassification for the statistical hard problems considered.  In addition, the simpler statistical regime $\delta=1$ presented with a logistic loss in Figure \ref{fig:results-appendix} (Appendix \ref{sec:simu-app}) reveals that the gap in performance does not depend upon the loss, and that all three estimators are affected by the statistical difficulty of the problem. 
	\medskip
	
	\noindent
	$\bullet$ Second, for group-sparse binary classification Example 2, our analysis reveals the computational advantage of group regularizations over L1 and Slope. Interestingly, Slope competes with its group opponents for the simpler statistical regime $\delta=0.4$ case presented in Figure \ref{fig:results-appendix}, Appendix \ref{sec:simu-app}---and for the hard regime when $p<5k$. However, it is significantly outperformed for hard problems with $10,000s$ of variable. In addition, Group L1-L2 regularization appears better than its L1-$L_\infty$ opponent, which additionally cannot reach the bounds presented in this paper. 
	\medskip
	
	\noindent
	$\bullet$ Finally, for sparse linear regression with heteroscedastic noise Example 3, our findings show the good performance of L1 and Slope regularized quantile regression when the SNR is low. Both methods reach a similar L2 estimation error and prediction accuracy than Lasso and appear as a solid alternative for this heteroscedastic noise regime. Note that all threee estimators reach the optimal minimax rate presented above. When the signal increases, Figure \ref{fig:results-appendix} (Appendix \ref{sec:simu-app}) suggests that L1 quantile regression and Lasso still compete with each other, while Slope performance slightly decreases. For both small and large SNR, all sparse estimators significantly outperform Ridge for both L2 estimation and prediction accuracy.

	\subsection{Additional experiments} \label{sec:simu-app}
	
	Figure \ref{fig:results-appendix} presents the three additional experiments described in Section \ref{sec:simu-app}. It considers Examples 1, 2 and 3 when the statistical settings are simpler than the ones in Figure  \ref{fig:results}---we respectively use a higher $\delta$ for Examples 1 and 2, and a higher $SNR$ for Example 3. In addition, we use the logistic loss for Example 1.
	
	\section{LP formulations for Section  \ref{sec:simu}}
	
	We present below LP formulations for the LP problems studied  in the computational experiments presented in Section \ref{sec:simu}. These formulations allows us to leverage the efficiency of modern commercial LP solvers as we solve these problems using \textsc{Gurobi} version $6.5$ with Python interface.
	
	\subsection{LP formulation for L1-SVM}\label{sec:L1-ref}
	We first consider L1 regularized SVM Problem \eqref{l1-problem} when $f$ is the hinge loss. This problem can be expressed as the following LP:
	\begin{equation} \label{L1-SVM-primal}
	\begin{myarray}[1.1]{l c c l}
	& \min \limits_{ \substack{\B{\xi} \in \mathbb{R}^n, \; \B{\beta}^+,\  \B{\beta}^- \in \mathbb{R}^p }   } &  \sum \limits_{i=1}^n \xi_i  + \lambda \sum \limits_{j=1}^p \beta_j^+  + \lambda \sum \limits_{j=1}^p \beta_j^- &\\
	&\sbt & \;\; \ \xi_i + y_i \B{x}_i^T \B{\beta}^+ - y_i \B{x}_i^T \B{\beta}^-  \ge 1 & \;\;\; i \in [n] \\
	&& \;\;\;\; \ \B{\xi} \ge 0, \ \B{\beta}^+ \ge 0, \ \B{\beta}^- \ge 0. &\\
	\end{myarray}
	\end{equation}

	\subsection{LP formulation for Group L1-$L_\infty$ SVM}\label{sec:L1-Linf-ref}
	The Group L1-L2 regularization considered in Problem \eqref{group-problem} has a popular alternative, namely the Group L1-$L_\infty$ penalty \citep{bach2011convex}, which considers the $L_\infty$ norm over the groups. Using this regularization, Problem \eqref{general} becomes
	\begin{equation} \label{L1_Linf-group}
	\min \limits_{ \B{\beta} \in \mathbb{R}^{p}:\  \| \B{\beta}  \|_1 \le 2R } \;\; \frac{1}{n}  \sum_{i=1}^n f \left( \langle \B{x_i},  \B{\beta} \rangle ;  y_i \right)  + \lambda \sum_{g=1}^G \| \B{\beta}_g \|_{\infty}.
	\end{equation}
	When $f$ is the hinge-loss, Problem \eqref{L1_Linf-group} can be expressed as an LP. To this end, we introduce the variables $\B{v}=(v_{g})_{g \in [G]}$ such that $v_g$ refers to the $L_{\infty}$ norm of the coefficients $\B\beta_{g}$. Problem \eqref{L1_Linf-group} can be reformulated as:
	\begin{equation} \label{group-SVM-primal}
	\begin{myarray}[1.1]{l c c l}
	& \min \limits_{ \substack{\B{\xi} \in \mathbb{R}^n, \; \B{\beta}^+,\  \B{\beta}^- \in \mathbb{R}^p, \; \B{v} \in \mathbb{R}^G }   } &  \sum \limits_{i=1}^n \xi_i  + \lambda \sum \limits_{g=1}^G v_g &\\
	&\sbt & \;\; \ \xi_i + y_i \B{x}_i^T \B{\beta}^+ - y_i \B{x}_i^T \B{\beta}^-  \ge 1 & \;\;\; i \in [n] \\
	&& \;\; \ v_g - \beta_j^+ - \beta_j^- \ge 0& j \in \mathcal{I}_g, \ g \in [G]   \\
	&& \;\;\;\; \ \B{\xi} \ge 0, \ \B{\beta}^+ \ge 0, \ \B{\beta}^- \ge 0, \  \B{v}  \ge 0.  &\\
	\end{myarray}
	\end{equation}
	We solve Problem \eqref{group-SVM-primal} with Gurobi in our experiments. When $f$ is the logistic loss, a proximal operator for Group L1-$L_\infty$ can be derived \citep{bach2011convex} using the Moreau decomposition presented in Section \ref{sec:thresholding}.
	
	\subsection{LP formulation for L1 regularized least-angle deviation loss}\label{sec:L1-LAD}
	Finally, when $f$ is the least-angle deviation loss \citep{wang2013l1} and $\Omega(.)$ is the L1 regularization, Problem \eqref{general} is expressed as:
	\begin{equation} \label{L1-LAD}
	\min \limits_{ \B{\beta} \in \mathbb{R}^p } 
	\sum \limits_{i=1}^n | y_i - \mathbf{x}_i^T \B{\beta}  |  + \lambda \| \B{\beta} \|_1,
	\end{equation}
	An LP formulation for Problem \eqref{L1-LAD} is:
	\begin{equation} \label{L1-LAD-LP}
	\begin{myarray}[1.1]{l c c l}
	& \min \limits_{ \substack{\B{\xi} \in \mathbb{R}^n, \; \B{\beta}^+,\  \B{\beta}^- \in \mathbb{R}^p, \; \B{v} \in \mathbb{R}^G }   } &  \sum \limits_{i=1}^n \xi_i  + \lambda \sum \limits_{j=1}^p \beta^+_j + \lambda \sum \limits_{j=1}^p \beta^-_j \\
	&\sbt & \;\; \ \xi_i \ge y_i  - \B{x}_i^T \B{\beta}^+ + \B{x}_i^T \B{\beta}^- & \;\;\; i \in [n] \\
	&& \;\; \ \xi_i \ge  \B{x}_i^T \B{\beta}^+ - \B{x}_i^T \B{\beta}^-  - y_i & \;\;\; i \in [n] \\
	&& \;\;\;\; \ \B{\xi} \ge 0, \ \B{\beta}^+ \ge 0, \ \B{\beta}^- \ge 0. &\\
	\end{myarray}
	\end{equation}
	Specific linear optimization techniques could be used for efficiently solving all three LP Problems \eqref{L1-SVM-primal}, \eqref{group-SVM-primal} and \eqref{L1-LAD-LP}. For instance, \citet{dedieu2019solving} recently combined first order methods with column-and-constraint generation algorithms to solve Problem \eqref{general} when $f$ is the hinge-loss and $\Omega(.)$ is the L1, Slope or Group L1-$L_\infty$ regularization.

\end{appendices}